\DocumentMetadata 
{
	lang		= en-US, %
	pdfversion  = 1.7,
	pdfstandard = a-2b,
}

\documentclass[twoside]{mitthesis}

\usepackage{amsmath}
\usepackage{amssymb}

\usepackage{todonotes}

\usepackage{tikz}
\usetikzlibrary{arrows.meta, positioning, calc, fit, backgrounds, matrix, decorations.pathreplacing}
\usepackage{pgffor}

\usepackage{booktabs}%

\usepackage{array}%

\usepackage{dcolumn}%
  \newcolumntype{d}[1]{D{.}{.}{#1}}%

\usepackage{longtable}%

\usepackage[nopatch=footnote]{microtype}%

\usepackage{multicol}

\usepackage{float}
\usepackage[section]{placeins}%
\usepackage{subcaption}
\usepackage{listings}
\usepackage{enumitem}
\usepackage{pifont}
\usepackage{circledsteps}
\usepackage{wrapfig}
\usepackage[most]{tcolorbox}
\tcbuselibrary{listings,skins,breakable}
\usepackage[capitalize,noabbrev]{cleveref}

\usepackage{algorithm}
\usepackage{algpseudocode}
\usepackage{bbm}
\usepackage{adforn}
\usepackage{fvextra}
\usepackage{inconsolata}

\definecolor{codegreen}{rgb}{0,0.6,0}
\definecolor{codegray}{rgb}{0.5,0.5,0.5}
\definecolor{codepurple}{rgb}{0.58,0,0.82}
\definecolor{backcolour}{rgb}{0.95,0.95,0.92}

\lstdefinestyle{mystyle}{
    commentstyle=\color{codegreen},
    keywordstyle=\color{magenta},
    numberstyle=\tiny\color{codegray},
    stringstyle=\color{codepurple},
    basicstyle=\ttfamily\footnotesize,
    breakatwhitespace=false,
    breaklines=true,
    keepspaces=true,
    showspaces=false,
    breakindent=0pt,
}
\newtcolorbox{pillbox}[2][]{colback=blue!10!white, colframe=blue!50!black,
    fonttitle=\bfseries, title=#2, sharp corners=south,
    rounded corners=north, #1}

\newcommand{\lang}{\textsc{Pasta-Lang}}
\newcommand{\pasta}{\textsc{Pasta}}

\newcommand{\async}{\texttt{\small <async>}}

\newcommand{\casync}{\texttt{\small </async>}}
\newcommand{\sync}{\texttt{\small <sync/>}}
\newcommand{\promise}{\texttt{\small <promise/>}}

\newcommand{\topic}{\texttt{\small topic}}
\newcommand{\tokens}{\texttt{\small tokens}}

\newcommand{\cmark}{\ding{51}}
\newcommand{\xmark}{\ding{55}}

\newcommand{\pastasft}{\emph{Pasta-SFT}}
\newcommand{\pastabon}[2]{\emph{Pasta-BoN-{#2}}}
\newcommand{\baseline}{\emph{Baseline-SFT}}

\newcommand{\pd}{\textsc{Planned Diffusion}}
\newcommand{\pdtag}[1]{\texttt{\small <#1>}}
\newcommand{\pdctag}[1]{\texttt{\small </#1>}}
\newcommand{\pdasync}{\texttt{\small <async>}}
\newcommand{\pdcasync}{\texttt{\small </async>}}
\newcommand{\pdsync}{\texttt{\small <sync/>}}
\newcommand{\pdeos}{\texttt{\small <eos/>}}

\newcommand{\pdsft}{\emph{PD}}
\newcommand{\pdsa}{\emph{PD-SA}}
\newcommand{\fastdllm}{\emph{Fast-dLLM}}

\newcommand{\func}[1]{\textnormal{\textsc{#1}}}

\definecolor{tagblue}{RGB}{20, 80, 200}
\definecolor{padgreen}{RGB}{36, 159, 37}
\definecolor{bg-gray}{RGB}{245, 245, 245}
\definecolor{asyncblue}{RGB}{86, 136, 185}

\lstdefinestyle{failstyle}{
    basicstyle=\ttfamily\footnotesize,
    breaklines=true,
    breakatwhitespace=false,
    columns=fullflexible,
    keepspaces=true,
    breakindent=0pt,
    breakautoindent=false,
    literate={[PAD]}{{\textcolor{padgreen}{[PAD]}}}5,
    moredelim=*[s][\color{red}]{<}{>},
    moredelim=*[s][\color{asyncblue}]{<async}{>},
    moredelim=*[s][\color{asyncblue}]{</async}{>},
}

\newtcolorbox{instructionbox}[1]{
  colback=white, colframe=black!70,
  fonttitle=\bfseries\sffamily, title=#1,
  sharp corners, boxrule=0.8pt,
  left=2mm, right=2mm, top=2mm, bottom=2mm,
}

\newtcblisting{rawoutputbox}[1]{
  colback=bg-gray, colframe=red!50!black,
  fonttitle=\bfseries\sffamily, title=#1,
  sharp corners, boxrule=0.8pt,
  listing only,
  listing options={style=failstyle},
  left=0mm, right=0mm, top=0mm, bottom=0mm,
}

\DeclareMathOperator*{\operatorE}{\mathbb{E}}

\usepackage[style=ext-numeric-comp,giveninits=true,maxbibnames=10,sorting=none,language=american]{biblatex}

    \AtEveryBibitem{%
      \ifentrytype{article}{%
        \renewbibmacro{in:}{}%
		\renewbibmacro*{issue+date}{%
		\iffieldundef{issue}%
        	{}%
        	{\printfield{issue}%
         	\setunit*{\addcomma\space}
        	}%
			\printdate
		}
      }{}
    }
	\renewbibmacro*{volume+number+eid}{%
        \printtext[bold]{\printfield{volume}}%
        \iffieldundef{number}
          {} %
          {\printtext[parens]{\printfield{number}}} %
        \setunit{\bibeidpunct}%
        \printfield{eid}
    }

\hypersetup{%
	pdfsubject={Self-Orchestrating Language Models: Leveraging Semantic Dependence for Efficient Inference},
	pdfkeywords={language models, semantic dependence, inference efficiency, parallel decoding, diffusion models, KV cache, Massachusetts Institute of Technology},
	pdfcontactemail={tianjin@mit.edu},
}

\begin{document}

\title{Self-Orchestrating Language Models: Leveraging Semantic Dependence for Efficient Inference}

\Author{Tian Jin}{Department of Electrical Engineering and Computer Science}[B.S., Haverford College, 2017][S.M., Massachusetts Institute of Technology, 2022]

\Degree{Doctor of Philosophy}{Department of Electrical Engineering and Computer Science}

\Supervisor{Michael Carbin}{Associate Professor of Electrical Engineering and Computer Science}[Department of Electrical Engineering and Computer Science]
\Supervisor{Jonathan Ragan-Kelley}{Associate Professor of Electrical Engineering and Computer Science}[Department of Electrical Engineering and Computer Science]

\Acceptor{Leslie A. Kolodziejski}{Professor of Electrical Engineering and Computer Science}{Chair, Department Committee on Graduate Students}

\DegreeDate{May}{2026}

\ThesisDate{May 15, 2026}

\Reader{Yoon Kim}{Assistant Professor}{Department of Electrical Engineering and Computer Science}
\Reader{Armando Solar-Lezama}{Professor}{Department of Electrical Engineering and Computer Science}
\Reader{Suvinay Subramanian}{Software Engineer}{Google}

\maketitle

\begin{abstract}
	
\noindent Large language models (LLMs) demonstrate impressive capabilities, but their deployment presents significant efficiency challenges. Autoregressive decoding imposes substantial inference latency and under-utilizes hardware accelerators in low batch size regimes. Discrete diffusion models can generate in parallel but struggle to match autoregressive quality without many diffusion denoising steps. Long-context reasoning creates memory bottlenecks that strain even state-of-the-art accelerators.

My thesis is that language models can direct their own inference execution strategy by annotating \emph{semantic dependence}---which tokens depend on which others---in their generation. I call such models \emph{self-orchestrating language models}. For each system, I design a runtime that acts on these annotations to parallelize autoregressive decoding, evict intermediate context, or derive denoising orders, achieving Pareto-optimal quality-efficiency trade-offs.

I demonstrate this approach through three self-orchestrating systems. First, PASTA uses semantic dependence to parallelize autoregressive decoding, training the model to annotate which output chunks can generate independently. Second, TIP uses semantic dependence to evict intermediate reasoning steps from the KV cache, reducing memory consumption while preserving accuracy. Third, Planned Diffusion uses semantic dependence to derive a denoising order for discrete diffusion, autoregressively generating a plan that specifies which chunks to denoise in parallel.
\end{abstract}

\chapter*{Acknowledgments}
\pdfbookmark[0]{Acknowledgments}{acknowledgments}

\newcommand{\leaf}{\adforn{21}}
\newcommand{\acksection}[1]{%
  \bigskip
  \begin{center}
    {\leaf\quad\textsc{#1}\quad\leaf}
  \end{center}
  \medskip
}

I arrived at MIT in the fall of 2020 after a journey from home that was an adventure in its own right. I came with the unshakable belief that I could do anything. It took a village to turn that belief into this dissertation.

I am lucky to have had Michael Carbin and Jonathan Ragan-Kelley as my advisors; working with them has been an incredibly rewarding experience for my personal growth and intellectual curiosity. Before applying to MIT, I cold-emailed many current PhD students I could find. Yunming Zhang was the one who wrote back with a detailed, thoughtful recommendation of Mike and Jonathan. This dissertation, and by extension the past six incredible years, exist in no small part because of that single email.

\textsc{To Mike:} being your student has been the quintessential MIT experience --- drinking from a firehose, and somehow learning to love it. I deeply value your commitment to your students' growth, a commitment that can, at times, feel overwhelming. But it is precisely through that rigor that we come to discover what we are truly capable of.

\textsc{To Jonathan:} starting a PhD in the middle of the pandemic could have easily derailed everything. Your support and encouragement made sure it didn't. As a performance engineer at heart, it has been a privilege to learn from one of the best up close.

\acksection{Mentors \& Collaborators}

I am deeply grateful to Gintare Karolina Dziugaite, my advisor in all but title. Karolina worked closely with me during the initial years that shaped me most as a researcher, carrying me through the many setbacks of my first few projects with equal parts unwavering support and sharp technical insight.

Working with Suvinay Subramanian has been one of the great gifts of this PhD. Suvinay is the rare kind of mentor who goes far out of his way for the people he works with --- and it is no exaggeration to say that without the internship at Google and the countless late nights we spent together, this dissertation would not exist.

This dissertation is also the product of many close collaborations. Ellie Y. Cheng consistently contributed to PASTA (Chapter~\ref{ch:pasta}) in ideation, experimental evaluation, and writing --- this work would not exist without her. Daniel Israel and I co-led the Planned Diffusion project (Chapter~\ref{ch:pd}): I handled training implementation and experimental evaluation, and Daniel developed the inference algorithm and method exposition. Thinking in Place (Chapter~\ref{ch:tip}) is mostly my own work. I also want to thank Zachary Ankner, Nikunj Saunshi, Amir Yazdanbakhsh, Blake M. Elias, and Reagan Choi --- each of whom left a mark on this dissertation. I thank Dan Roy, Vaishnavh Nagarajan, Xin Dong, and Nolan Clement for being amazing collaborators on my research on pruning during the early years of my PhD.

I thank my thesis committee members --- Yoon Kim, Armando Solar-Lezama, and Suvinay Subramanian --- for their support and feedback.

\acksection{Community}

One of the most rewarding experiences from MIT has been starting and organizing the MIT SysML Discussion Group, generously supported by my advisor Mike. Among the early members: William Brandon, Alex Gu, Vincent Huang, Yuka Ikarashi, Aniruddha (Ani) Nrusimha, Anne Ouyang, Kevin Qian, and Zhiye (Zoey) Song. The list is necessarily incomplete, but the gratitude is not. I have been part of many reading groups, but the people in this one made it something special. The talent and drive in that room were unreal. I have never seen ideas move so fast from a whiteboard to a finished paper.

The members of the Programming Systems Group and the Visual Computing Languages and Systems Group made the lab feel less like work and more like home: David Akeley, Eric Atkinson, Manya Bansal, Gilbert Bernstein, William Brandon, Kartik Chandra, Ellie Y. Cheng, Yi Ding, Jonathan Frankle, Rahul Goel, Rohan Gumaste, Yuka Ikarashi, Tzu-Mao Li, Amanda Liu, Karima Ma, Ahmed Mahmoud, Jesse Michel, Alex Reinking, Alexander Renda, Logan Weber, Cambridge Yang, and Charles Yuan.

\acksection{Family}

I thank my wife, Yilin Li. She has a way of making every setback feel okay --- and true to the Shanghainese spirit she embodies, she never lets me settle for anything less than my best. I am grateful to my parents, who made it possible for me to study in the United States at all --- none of this would exist without their vision, hard work, and unconditional love. I also thank my grandparents, Min Cao, Jielian Cheng, Hongsheng Jin, and Fengzhou Wang. Their love and care made me who I am.

\tableofcontents
\listoffigures
\listoftables

\chapter{Introduction}

Large language model (LLM) inference dominates the cost of deploying language models, and that cost grows faster than hardware can reduce it.
Inference already accounts for roughly two-thirds of all AI compute, with total demand rising an estimated four to five times per year~\cite{deloitte2026compute}---a rate that outpaces even generational leaps in hardware efficiency~\cite{nvidia2026rubin}.
This widening gap makes algorithmic improvements to LLM inference---methods that expand the quality-efficiency Pareto frontier---among the most pressing open problems in computing.

Making computation faster often begins with the same question: which instructions must consume the output of which others?
Performance engineers call these producer--consumer constraints \emph{data dependences} and exploit every degree of freedom they leave---reordering or parallelizing instructions that need not await each other's results, and reclaiming memory the moment no future instruction will read it.

\section{Data Dependence in LLM Inference}\label{sec:dependence}

The most common type of LLM---the \emph{autoregressive language model}---generates tokens sequentially.
This sequential generation under-utilizes modern hardware
accelerators---at small batch sizes, autoregressive decoding achieves
less than 20\% of peak throughput~\cite{pope2022efficiently}.
Decoding tokens in parallel would alleviate this cost,
but the autoregressive factorization forbids it: every token depends
on all predecessors.

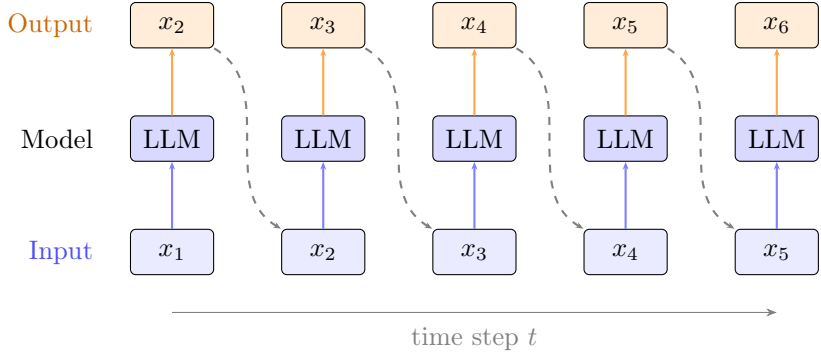
\begin{figure}[t]
\centering
\begin{tikzpicture}[
    token/.style={draw, rounded corners=2pt, minimum width=1.1cm, minimum height=0.6cm, font=\footnotesize},
    model/.style={draw, rounded corners=2pt, minimum width=1.1cm, minimum height=0.6cm, font=\footnotesize, fill=blue!15},
    input/.style={token, fill=blue!8},
    output/.style={token, fill=orange!15},
    arr/.style={-{Stealth[length=3pt]}, thick}
  ]
  \foreach \i in {1,...,5} {
    \pgfmathtruncatemacro{\inext}{\i+1}
    \pgfmathtruncatemacro{\xcoord}{(\i-1)*2}
    \node[input] (in\i) at (\xcoord, 0) {$x_{\i}$};
    \node[model] (m\i)  at (\xcoord, 1.5) {LLM};
    \node[output] (out\i) at (\xcoord, 3.0) {$x_{\inext}$};
    \draw[arr, blue!50] (in\i) -- (m\i);
    \draw[arr, orange!70] (m\i) -- (out\i);
  }
  \foreach \i in {1,...,4} {
    \pgfmathtruncatemacro{\inext}{\i+1}
    \draw[arr, dashed, gray] (out\i.south east) to[out=-30, in=150] (in\inext.north west);
  }
  \node[left, font=\footnotesize, blue!70] at (-0.9, 0) {Input};
  \node[left, font=\footnotesize] at (-0.9, 1.5) {Model};
  \node[left, font=\footnotesize, orange!80!black] at (-0.9, 3.0) {Output};
  \draw[-{Stealth[length=4pt]}, gray] (0, -0.8) -- (8.0, -0.8);
  \node[font=\footnotesize, gray] at (4.0, -1.15) {time step $t$};
\end{tikzpicture}
\caption{Autoregressive decoding generates tokens sequentially.  At each time step, the model consumes the previous token and produces the next one. Every step depends on the output of the preceding step, forming a total-order dependence chain.}
\label{fig:autoreg}
\end{figure}

Formally, an autoregressive model factorizes the joint probability of a token sequence $x_1, x_2, \ldots, x_n$ into a product of conditionals:
\begin{equation}\label{eq:autoreg}
P(x_1, x_2, \ldots, x_n) \;=\; P(x_1) \cdot P(x_2 \mid x_1) \cdot P(x_3 \mid x_1, x_2) \;\cdots\; P(x_n \mid x_1, \ldots, x_{n-1})
\end{equation}
Each token's probability distribution conditions on every preceding token.
At inference time, this factorization forces the model to generate tokens one at a time: it samples token $x_t$, feeds it back as input, and only then samples token $x_{t+1}$ (Figure~\ref{fig:autoreg}).

To understand what freedom this sequential chain leaves for parallel execution, I turn to the concept of data dependence.
A \emph{data dependence} exists between two computations when one produces a value that the other consumes~\cite{allen2002optimizing}.
The set of all data dependences in a program forms a \emph{data dependence graph}: nodes represent computations and edges represent producer-consumer relationships.
This graph reveals which computations the hardware must execute in order and which it can freely reorder or parallelize.
The fewer data dependences a program exhibits, the more freedom an optimizer has to improve its execution.

To map autoregressive decoding onto this graph, I treat each forward pass as a node: forward pass $t$ computes $p(x_t \mid x_{<t})$ and samples a token.
The conditioning on $x_{<t}$ places an edge from every earlier node to node $t$, because forward pass $t$ consumes the outputs of all preceding forward passes.
The resulting data dependence graph is a total order---no optimization that respects data dependence can reorder, parallelize, or skip any forward pass in this chain.
The factorization conditions each token on all predecessors without distinguishing which predecessors influence each token's distribution.

\section{Beyond Autoregression}\label{sec:semantic-dep}

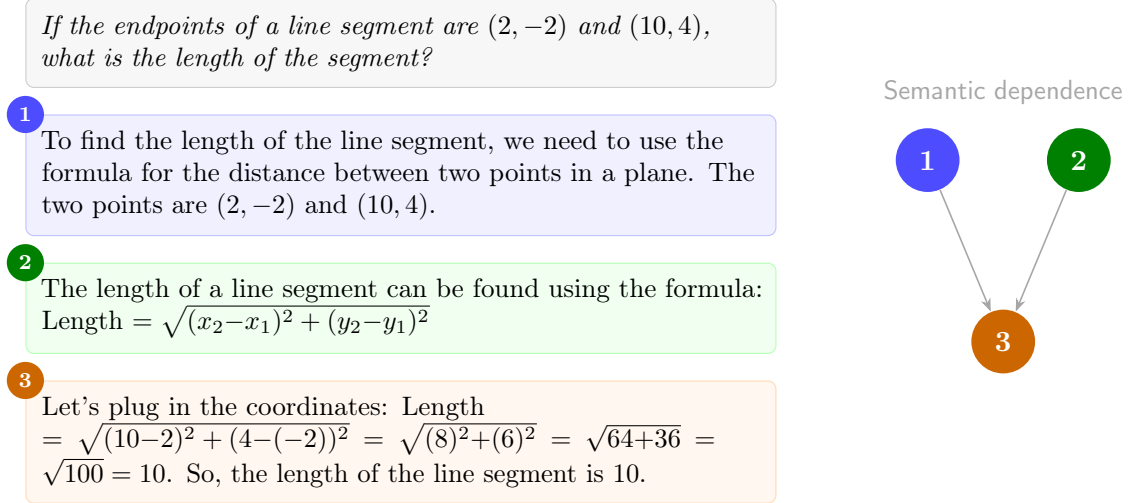
\begin{figure}[t]
\centering
\begin{tikzpicture}[
    sbox/.style={draw, rounded corners=3pt, text width=9.5cm, align=left,
                 font=\footnotesize, inner sep=6pt, line width=0.4pt},
    num/.style={circle, fill=#1, text=white, font=\scriptsize\bfseries,
                inner sep=1.5pt, minimum size=14pt},
    gnode/.style={circle, fill=#1, text=white, font=\small\bfseries,
                  minimum size=24pt},
    arr/.style={-{Stealth[length=5pt]}, semithick, gray!70},
    lbl/.style={font=\footnotesize\sffamily, gray!70},
  ]
  \node[sbox, fill=gray!6, draw=gray!40, font=\footnotesize\itshape] (prompt) at (0, 0) {%
    If the endpoints of a line segment are $(2, -2)$ and $(10, 4)$,
    what is the length of the segment?};
  \node[sbox, fill=blue!6, draw=blue!25, anchor=north] (s1)
    at ([yshift=-10pt]prompt.south) {%
    To find the length of the line segment, we need to use the formula
    for the distance between two points in a plane.
    The two points are $(2, -2)$ and $(10, 4)$.};
  \node[num=blue!70] at (s1.north west) {1};
  \node[sbox, fill=green!6, draw=green!30, anchor=north] (s2)
    at ([yshift=-10pt]s1.south) {%
    The length of a line segment can be found using the formula:
    Length $= \sqrt{(x_2{-}x_1)^2 + (y_2{-}y_1)^2}$};
  \node[num=green!50!black] at (s2.north west) {2};
  \node[sbox, fill=orange!6, draw=orange!30, anchor=north] (s3)
    at ([yshift=-10pt]s2.south) {%
    Let's plug in the coordinates:
    Length $= \sqrt{(10{-}2)^2 + (4{-}({-}2))^2}
            = \sqrt{(8)^2{+}(6)^2}
            = \sqrt{64{+}36} = \sqrt{100} = 10$.
    So, the length of the line segment is~10.};
  \node[num=orange!80!black] at (s3.north west) {3};

  \path (prompt.north east) -- (s3.south east) coordinate[midway] (mid);
  \node[lbl, anchor=south] at ([xshift=3cm, yshift=1.8cm]mid) {Semantic dependence};
  \node[gnode=blue!70]       (n1) at ([xshift=2cm, yshift=1.2cm]mid)  {1};
  \node[gnode=green!50!black] (n2) at ([xshift=4cm, yshift=1.2cm]mid)  {2};
  \node[gnode=orange!80!black](n3) at ([xshift=3cm, yshift=-1.2cm]mid) {3};
  \draw[arr] (n1) -- (n3);
  \draw[arr] (n2) -- (n3);
\end{tikzpicture}
\caption{A chain-of-thought response and its semantic dependence graph.
\textit{Left:}~the prompt (gray) and the response that the model produces sequentially, with each reasoning step colored and numbered.
\textit{Right:}~the dependence graph over the same steps.
Steps~1 and~2 do not depend on each other; step~3 depends on both.}
\label{fig:semdep-dag}
\end{figure}

Consider the chain-of-thought response in Figure~\ref{fig:semdep-dag}, where an LLM answers a geometry question in three steps.
Steps~1 and~2---identifying the coordinates and stating the distance formula---address semantically independent subproblems and could generate in either order.
Only step~3 depends on both: it plugs coordinates into the formula to compute the answer.

Steps~1 and~2 need not wait for each other, and only step~3 requires both---not because one token preceded another, but because one step's content requires another's.
I call this content-level structure \emph{semantic dependence}.%
\footnote{The idea of directed semantic relationships has historical roots: Schank's Conceptual Dependency theory represents sentence meaning as a graph of primitive concepts linked by directed dependence edges, each encoding that one concept requires another to be interpretable~\cite{schank1972conceptual}. I adopt the same organizing principle---directed dependence relationships between semantic units---but apply it to inference efficiency rather than symbolic language understanding.}
Autoregressive decoding enforces a total-order data dependence---every forward pass depends on all previous ones---but this total order over-approximates the true dependence that output generation requires.
I propose to extract semantic dependence, a different kind of data dependence, by examining the meaning of generation content.
No static analysis of the model architecture or decoding algorithm can recover it, so I propose that the model itself annotate semantic dependence during generation, and I co-design a runtime that acts on these annotations to implement inference optimizations.

\section{Thesis and Contributions}\label{sec:contributions}

My thesis is the following:

\begin{quote}\itshape
Language models can direct their own inference execution strategy by annotating \emph{semantic dependence}---which tokens depend on which others---in their generation. I call such models \emph{self-orchestrating language models}. For each system, I design a runtime that acts on these annotations to parallelize autoregressive decoding, evict intermediate context, or derive denoising orders, achieving Pareto-optimal quality-efficiency trade-offs.
\end{quote}

\noindent I demonstrate this thesis through three systems, each targeting a different inference bottleneck but sharing a uniform design: an annotation language for semantic dependence, a runtime that interprets these annotations, and a training procedure that teaches the model to produce accurate annotations.

\clearpage
\subsection{PASTA: Parallel Generation through Semantic Dependence}\label{sec:contrib-pasta}

PASTA (Parallel Structure Annotation) partitions a response into \emph{chunks}---contiguous token spans that each express a single semantic concept---and trains the model to annotate semantic dependence between them~\cite{jin2025pasta}.
Chunks without dependences on each other generate in parallel, while synchronization points enforce dependences where they exist.
I develop an annotation language that extends the model's vocabulary with special tokens for marking independent chunks and synchronization barriers.
A co-designed runtime implements \emph{asynchronous decoding}: it launches parallel generation threads for independent chunks and synchronizes them at dependence boundaries.
I train models to generate these annotations through supervised finetuning on LLM-annotated data, followed by preference optimization that jointly optimizes for quality and latency.

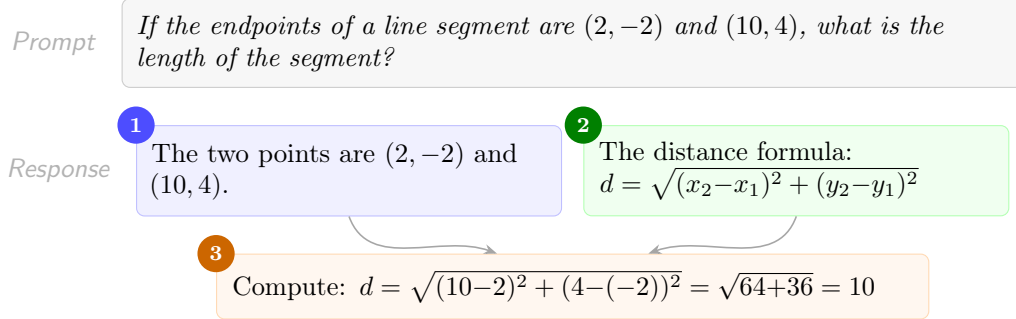
\begin{figure}[t]
\centering
\begin{tikzpicture}[
    sbox/.style={draw, rounded corners=3pt, align=left,
                 font=\footnotesize, inner sep=6pt, line width=0.4pt},
    num/.style={circle, fill=#1, text=white, font=\scriptsize\bfseries,
                inner sep=1.5pt, minimum size=14pt},
    arr/.style={-{Stealth[length=5pt]}, semithick, gray!70},
    lbl/.style={font=\footnotesize\sffamily\itshape, gray!60},
  ]
  \node[sbox, fill=gray!6, draw=gray!40, text width=11.5cm,
        font=\footnotesize\itshape] (prompt) at (0,0) {%
    If the endpoints of a line segment are $(2, -2)$ and $(10, 4)$,
    what is the length of the segment?};
  \node[lbl, anchor=east] at ([xshift=-6pt]prompt.west) {Prompt};
  \node[sbox, fill=blue!6, draw=blue!25, text width=5.2cm,
        anchor=north east] (c1)
    at ([yshift=-14pt, xshift=-4pt]prompt.south) {%
    The two points are $(2, -2)$ and $(10, 4)$.};
  \node[num=blue!70] at (c1.north west) {1};
  \node[lbl, anchor=east] at ([xshift=-6pt]c1.west) {Response};
  \node[sbox, fill=green!6, draw=green!30, text width=5.2cm,
        anchor=north west] (c2)
    at ([yshift=-14pt, xshift=4pt]prompt.south) {%
    The distance formula:\newline
    $d = \sqrt{(x_2{-}x_1)^2 + (y_2{-}y_1)^2}$};
  \node[num=green!50!black] at (c2.north west) {2};
  \node[sbox, fill=orange!6, draw=orange!30, text width=9cm,
        anchor=north] (c3)
    at ([yshift=-14pt]$(c1.south east)!0.5!(c2.south west)$) {%
    Compute: $d = \sqrt{(10{-}2)^2 + (4{-}({-}2))^2}
    = \sqrt{64{+}36} = 10$};
  \node[num=orange!80!black] at (c3.north west) {3};
  \draw[arr] (c1.south) to[out=-70, in=160] ([xshift=-1cm]c3.north);
  \draw[arr] (c2.south) to[out=-110, in=20] ([xshift=1cm]c3.north);
\end{tikzpicture}
\caption{PASTA trains the model to partition its response into chunks and generate chunks without semantic dependences in parallel.
Here, the model identifies that extracting coordinates (chunk~1) and recalling the distance formula (chunk~2) require no information from each other, so the model directs the runtime to generate them simultaneously via special-token annotations.
Chunk~3 depends on both and generates only after they complete.}
\label{fig:intro:pasta}
\end{figure}

PASTA models achieve $1.21\times$ to $1.93\times$ geometric mean speedup with quality changes from $+2.2\%$ to $-7.1\%$ on AlpacaEval, Pareto-dominating all existing asynchronous decoding methods including Skeleton-of-Thought~\cite{ning2023skeleton} and APAR~\cite{liu2024apar}.
Figure~\ref{fig:intro:pasta} shows how PASTA exploits this dependence structure.

\clearpage
\subsection{Thinking in Place: Context Management through Semantic Dependence}\label{sec:contrib-tip}

Long-context reasoning creates critical memory bottlenecks for LLM inference.
Reasoning models routinely generate tens of thousands of tokens to solve challenging problems~\cite{zhang2025deepseekr1}, and each token's key-value (KV) cache entry persists throughout generation.
Existing methods address memory bottlenecks with hand-crafted heuristics: StreamingLLM retains a fixed-size sliding window~\cite{xiao2023streamingllm}, H2O evicts tokens with low accumulated attention scores~\cite{zhang2023h2o}, and SpAtten prunes tokens based on attention-based importance~\cite{wang2021spatten}.
These heuristics require human engineering to design and do not improve as more training compute becomes available.

Thinking in Place (TIP) constrains the model to a fixed number of reasoning steps as working memory and trains it to decide which steps to evict when memory is full~\cite{jin2026tip}.
Each eviction expresses a semantic dependence judgment: the model predicts that no future step will reference the evicted content.
The runtime leverages these judgments to evict KV cache entries, reducing peak memory.
I train the model using reinforcement learning~\cite{shao2024grpo}, rewarding correct final answers so the model learns to balance retaining steps needed for future reasoning against evicting steps to free memory.

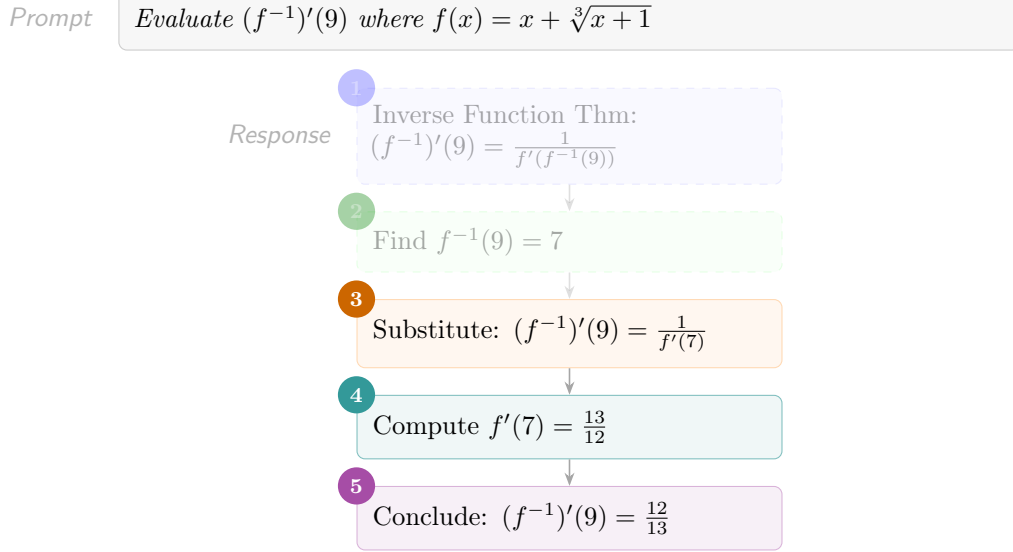
\begin{figure}[t]
\centering
\begin{tikzpicture}[
    sbox/.style={draw, rounded corners=3pt, align=left,
                 font=\footnotesize, inner sep=6pt, line width=0.4pt},
    num/.style={circle, fill=#1, text=white, font=\scriptsize\bfseries,
                inner sep=1.5pt, minimum size=14pt},
    arr/.style={-{Stealth[length=5pt]}, semithick, gray!70},
    lbl/.style={font=\footnotesize\sffamily\itshape, gray!60},
  ]
  \node[sbox, fill=gray!6, draw=gray!40, text width=11.5cm,
        font=\footnotesize\itshape] (prompt) at (0,0) {%
    Evaluate $(f^{-1})'(9)$ where $f(x) = x + \sqrt[3]{x+1}$};
  \node[lbl, anchor=east] at ([xshift=-6pt]prompt.west) {Prompt};
  \node[sbox, fill=blue!6, draw=blue!25, text width=5.2cm,
        anchor=north, dashed, opacity=0.35] (s1)
    at ([yshift=-14pt]prompt.south) {%
    Inverse Function Thm:\newline
    $(f^{-1})'(9) = \frac{1}{f'(f^{-1}(9))}$};
  \node[num=blue!70, opacity=0.35] at (s1.north west) {1};
  \node[lbl, anchor=east] at ([xshift=-6pt]s1.west) {Response};
  \node[sbox, fill=green!6, draw=green!30, text width=5.2cm,
        anchor=north, dashed, opacity=0.35] (s2)
    at ([yshift=-10pt]s1.south) {%
    Find $f^{-1}(9) = 7$};
  \node[num=green!50!black, opacity=0.35] at (s2.north west) {2};
  \node[sbox, fill=orange!6, draw=orange!30, text width=5.2cm,
        anchor=north] (s3)
    at ([yshift=-10pt]s2.south) {%
    Substitute: $(f^{-1})'(9) = \frac{1}{f'(7)}$};
  \node[num=orange!80!black] at (s3.north west) {3};
  \node[sbox, fill=teal!6, draw=teal!50, text width=5.2cm,
        anchor=north] (s4)
    at ([yshift=-10pt]s3.south) {%
    Compute $f'(7) = \frac{13}{12}$};
  \node[num=teal!80] at (s4.north west) {4};
  \node[sbox, fill=violet!6, draw=violet!30, text width=5.2cm,
        anchor=north] (s5)
    at ([yshift=-10pt]s4.south) {%
    Conclude: $(f^{-1})'(9) = \frac{12}{13}$};
  \node[num=violet!70] at (s5.north west) {5};
  \draw[arr, opacity=0.35] (s1.south) -- (s2.north);
  \draw[arr, opacity=0.35] (s2.south) -- (s3.north);
  \draw[arr] (s3.south) -- (s4.north);
  \draw[arr] (s4.south) -- (s5.north);
\end{tikzpicture}
\caption{Thinking in Place (TIP) trains the model to partition long chain-of-thought reasoning into semantically distinct steps and to keep only a bounded number of reasoning steps in memory.
Here, the model produces a five-step solution; step~3 subsumes the information of steps~1 and~2, so no future step will depend on them.
The model learns to mark these steps as dead during generation so the runtime can free their corresponding KV cache entries, reducing peak memory and speeding up subsequent token generation.}
\label{fig:intro:tip}
\end{figure}

On AIME 2024 and 2025, TIP achieves 59\% and 54\% KV cache reduction with 6.4- and 2.1-point accuracy gains respectively over dense decoding.
It Pareto-dominates heuristic-eviction baselines (StreamingLLM, Paged H2O) at comparable memory budgets, outperforming them by 3--10 accuracy points.
Figure~\ref{fig:intro:tip} shows how TIP exploits this dependence structure.

\clearpage
\subsection{Planned Diffusion: Decomposed Generation through Semantic Dependence}\label{sec:contrib-pd}

Discrete diffusion models offer an alternative to autoregressive decoding: they can generate multiple tokens in parallel through diffusion denoising~\cite{sahoo2024mdlm}.
Sampling from these models requires a \emph{denoising order}---a strategy for deciding which tokens to unmask at each step.
Existing approaches rely on heuristics: random denoising order or confidence-based thresholds~\cite{wu2025fastdllm}.
These heuristics struggle to match autoregressive quality without many denoising steps, creating a trade-off between quality and latency.

Planned Diffusion trains a single model to first autoregressively plan the response as a sequence of independent chunks, then denoise all chunks in parallel through diffusion.
The plan encodes the model's judgment that these chunks have no semantic dependence on each other.
A co-designed runtime executes planning iteratively: it denoises the independent chunks, then the model re-plans conditioned on all content generated so far.

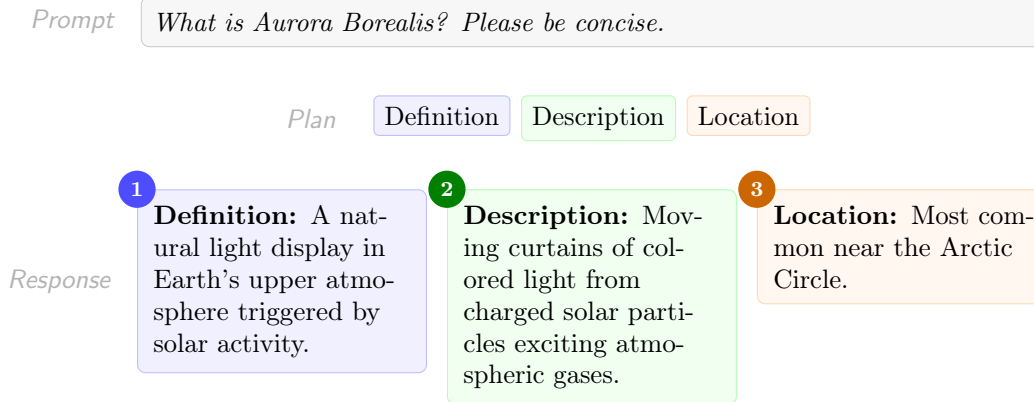
\begin{figure}[t]
\centering
\begin{tikzpicture}[
    sbox/.style={draw, rounded corners=3pt, align=left,
                 font=\footnotesize, inner sep=6pt, line width=0.4pt},
    num/.style={circle, fill=#1, text=white, font=\scriptsize\bfseries,
                inner sep=1.5pt, minimum size=14pt},
    lbl/.style={font=\footnotesize\sffamily\itshape, gray!60},
  ]
  \node[sbox, fill=gray!6, draw=gray!40, text width=11.5cm,
        font=\footnotesize\itshape] (prompt) at (0,0) {%
    What is Aurora Borealis? Please be concise.};
  \node[lbl, anchor=east] at ([xshift=-6pt]prompt.west) {Prompt};
  \node[anchor=north] (plan)
    at ([yshift=-14pt]prompt.south) {%
    \tikz[baseline=(p.base)]{
      \node[draw, rounded corners=2pt, fill=blue!6, draw=blue!25,
            font=\footnotesize, inner sep=4pt] (p) {Definition};
    }\hspace{4pt}%
    \tikz[baseline=(p.base)]{
      \node[draw, rounded corners=2pt, fill=green!6, draw=green!30,
            font=\footnotesize, inner sep=4pt] (p) {Description};
    }\hspace{4pt}%
    \tikz[baseline=(p.base)]{
      \node[draw, rounded corners=2pt, fill=orange!6, draw=orange!30,
            font=\footnotesize, inner sep=4pt] (p) {Location};
    }};
  \node[lbl, anchor=east] at ([xshift=-6pt]plan.west) {Plan};
  \node[sbox, fill=blue!6, draw=blue!25, text width=3.4cm,
        anchor=north] (c1)
    at ([yshift=-14pt, xshift=-4.1cm]plan.south) {%
    \textbf{Definition:} A natural light display in Earth's upper atmosphere triggered by solar activity.};
  \node[num=blue!70] at (c1.north west) {1};
  \node[lbl, anchor=east] at ([xshift=-6pt]c1.west) {Response};
  \node[sbox, fill=green!6, draw=green!30, text width=3.4cm,
        anchor=north] (c2)
    at ([yshift=-14pt]plan.south) {%
    \textbf{Description:} Moving curtains of colored light from charged solar particles exciting atmospheric gases.};
  \node[num=green!50!black] at (c2.north west) {2};
  \node[sbox, fill=orange!6, draw=orange!30, text width=3.4cm,
        anchor=north] (c3)
    at ([yshift=-14pt, xshift=4.1cm]plan.south) {%
    \textbf{Location:} Most common near the Arctic Circle.};
  \node[num=orange!80!black] at (c3.north west) {3};
\end{tikzpicture}
\caption{Planned Diffusion trains a single model to first plan the response as a sequence of chunks that do not depend on each other semantically, each with a distinct topic, then denoise all chunks in parallel through diffusion.
Here, the model plans three independent chunks---definition, description, and location---and the runtime denoises them simultaneously through diffusion denoising.}
\label{fig:intro:pd}
\end{figure}

Planned Diffusion achieves $1.27\times$ to $1.81\times$ speedup over autoregressive decoding with quality changes of $-0.87\%$ to $-5.4\%$ on AlpacaEval, expanding the Pareto frontier between autoregressive and diffusion baselines.
Figure~\ref{fig:intro:pd} shows how Planned Diffusion exploits this semantic dependence structure.

\subsection{Implications}\label{sec:implications}

I demonstrate across three systems that a single design methodology---an annotation language for semantic dependence, a co-designed runtime, and a training procedure---produces Pareto improvements against manually designed, domain-specific baselines across three distinct inference bottlenecks.
Across the aforementioned three systems, I train language models to manage the serving system by annotating semantic dependence and directing the runtime to implement inference optimizations.
This represents a departure from state-of-the-art serving systems, which manage model execution but never let model execution manage the serving system.
Together, these results suggest that self-orchestrating language model systems offer practitioners a versatile mechanism for improving inference efficiency beyond what contemporary serving systems can achieve.

\subsection{Scope and Limitations}\label{sec:scope}

The results are empirical in nature: I demonstrate that learned annotations Pareto-dominate existing heuristics but do not analyze what upper-bound speedup or memory reduction is achievable.
Additionally, while the three systems share a uniform design methodology, they have not converged to a single annotation language or a single runtime system that delivers all inference optimizations at once.%

\chapter{Contemporary LLM Inference Design}
\label{ch:prelim}

The previous chapter argued that autoregressive modeling imposes a total-order dependence chain on token generation, preventing standard dependence-respecting optimizations.
This chapter unpacks the two concrete bottlenecks that total-order dependence creates---sequential latency and growing memory footprint---and reviews the techniques that address them.

\section{Autoregressive Inference and Its Bottlenecks}
\label{sec:prelim:bottlenecks}

\begin{figure}[t]
\centering
\includegraphics[width=\textwidth]{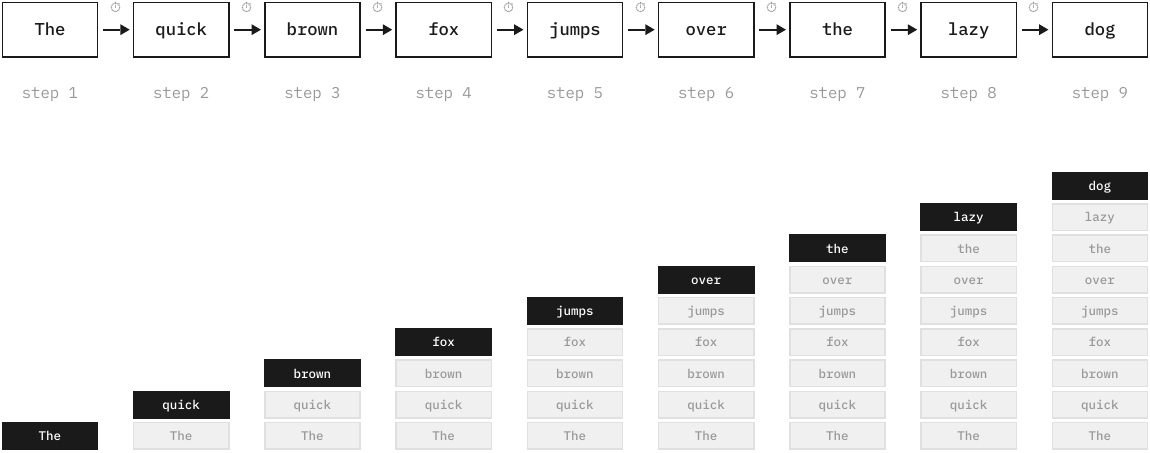}
\caption{The two bottlenecks of autoregressive inference.  Top: generating each token depends on the previous one, forming a sequential chain.  Bottom: the KV cache grows linearly with every generated token.}
\label{fig:bottlenecks}
\end{figure}

An autoregressive language model generates text one token at a time.
At each decoding step, the model runs a \emph{forward pass}: it takes the token from the previous step and passes it through all its layers---conditioning on all tokens it has generated so far---to output a probability distribution over the next token.
This token-by-token process creates two costs that scale with sequence length (Figure~\ref{fig:bottlenecks}): one in time, one in memory.

The first cost is the sequential latency of generation.
Because each step consumes the output of the previous one, no two steps can execute in parallel within a single request.
Generating $n$ tokens requires $n$ serial steps, and the wall-clock time grows with sequence length regardless of available hardware parallelism.

The second cost is the memory footprint of generation.
At each step, the model produces intermediate states that later steps reuse.
Caching these states avoids recomputing them: each decoding step appends a pair of vectors to a running pool---the \emph{KV cache}---for future reuse.
Because the KV cache grows with every generated token, it scales linearly with sequence length and dominates accelerator memory for long sequences~\cite{kim2025oaken}.

\section{Speculative Decoding}
\label{sec:prelim:spec}

Speculative decoding~\cite{leviathan2023fast,chen2023accelerating} speeds up autoregressive decoding without changing the output distribution.
The key observation is that although an autoregressive model cannot generate multiple tokens at once, it can, given a sequence of candidate tokens, verify in parallel whether it could have generated that sequence itself.
Speculative decoding exploits this asymmetry: a cheaper \emph{draft model} proposes several tokens, and the full-size \emph{target model} verifies all of them in a single forward pass.

\begin{figure}[t]
\centering
\begin{subfigure}[t]{\textwidth}
  \centering
  \includegraphics[width=\textwidth]{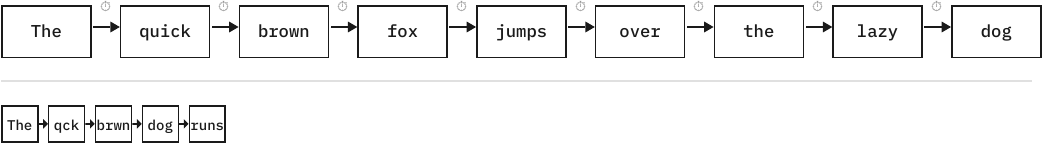}
  \caption{Draft phase.  To produce a sequence of nine tokens, a target model (top) needs nine serial forward passes on its own.  Speculative decoding reduces the number of required sequential forward passes: a cheaper draft model (bottom) proposes candidate tokens---five in this example---for the target model to verify at once.}
  \label{fig:spec-draft}
\end{subfigure}
\vspace{0.5em}
\begin{subfigure}[t]{\textwidth}
  \centering
  \includegraphics[width=\textwidth]{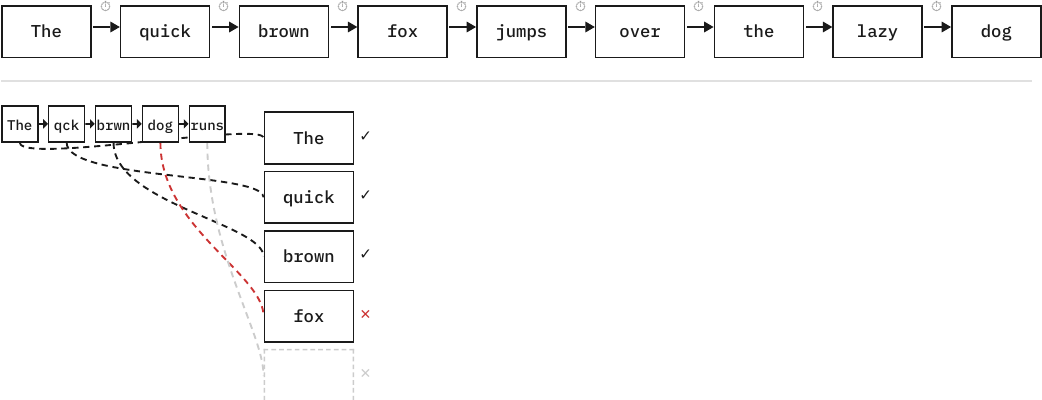}
  \caption{Verify phase.  The target model runs one forward pass over all draft tokens and compares its own distribution at each position to the draft token there.  The first three draft tokens pass (\checkmark); the fourth diverges from the target distribution (\texttimes), so the system discards it and the remaining draft token, resamples a corrected token, and starts a new round.}
  \label{fig:spec-verify}
\end{subfigure}
\caption{Speculative decoding: draft then verify.}
\label{fig:spec-decoding}
\end{figure}

The procedure alternates between two phases (Figure~\ref{fig:spec-decoding}).
In the draft phase, a cheaper draft model generates a sequence of candidate tokens.
In the verify phase, the target model runs one forward pass over the entire sequence---context plus draft tokens---producing a next-token distribution at every draft position simultaneously.
A rejection-sampling scheme then walks through the draft tokens left to right: it accepts each token when the target and draft distributions agree on it, and rejects it when they diverge.
At the first rejection, the speculative decoding algorithm discards that token and all remaining draft tokens.
Because the verify pass already computed the target distributions, the algorithm samples one additional token at no extra cost and begins the next round.

The rejection-sampling step ensures that the generated sequence follows the exact target distribution, regardless of how much the draft model diverges from the target.
The speedup depends on the \emph{acceptance rate}: the fraction of draft tokens that pass verification.
When draft and target agree on most tokens, each round accepts many tokens per target forward pass, reducing wall-clock latency.
When they disagree frequently, each round still produces at least one correct token---at the cost of the extra draft computation.

\section{Asynchronous Decoding}
\label{sec:prelim:async}

An emerging line of work improves autoregressive latency by exploiting the semantics of the generated content itself.
The key observation is that not all chunks in an LLM response depend on each other: list items and separate paragraphs often carry independent content.
If the decoding algorithm can identify the \emph{semantic dependence} between chunks---that is, which chunks semantically depend on which other chunks' content---it can decode chunks without semantic dependence concurrently, each as a separate autoregressive process~\cite{ning2023skeleton,liu2024apar}.

Skeleton-of-Thought~(SoT)~\cite{ning2023skeleton} takes a prompting approach to identify semantic dependence.
Given a user request, SoT first issues one LLM call that produces a bullet-point skeleton of the response.
It then applies regex-based pattern matching to extract the skeleton points, and issues one LLM call per point---all in parallel---to expand each point into a full paragraph.

APAR~\cite{liu2024apar} replaces the prompting strategy with finetuning.
APAR applies regex-based pattern matching to training data to identify syntactically parallel segments---list items and paragraph bodies given their first sentences---and finetunes the LLM on this restructured data.
At inference time, the finetuned model emits special tokens that mark parallel segments, and a runtime reads these tokens and spawns parallel decoding threads accordingly.
The identification of parallel structure, however, still depends on syntactic heuristics.

Both methods rely on hand-crafted regex patterns to identify which chunks can decode in parallel.
These heuristics are not learnable: they cannot improve with more data or training, they miss parallelism that does not follow list or paragraph structure, and they treat all list items as independent even when later items depend on earlier ones.

\section{Discrete Diffusion Language Models}
\label{sec:prelim:diffusion}

Discrete diffusion language models replace autoregressive decoding with diffusion denoising, predicting multiple tokens per forward pass without committing to a fixed decoding order---though how many tokens the model can reliably unmask per step remains an open problem.

\begin{figure}[t]
\centering
\includegraphics[width=\textwidth]{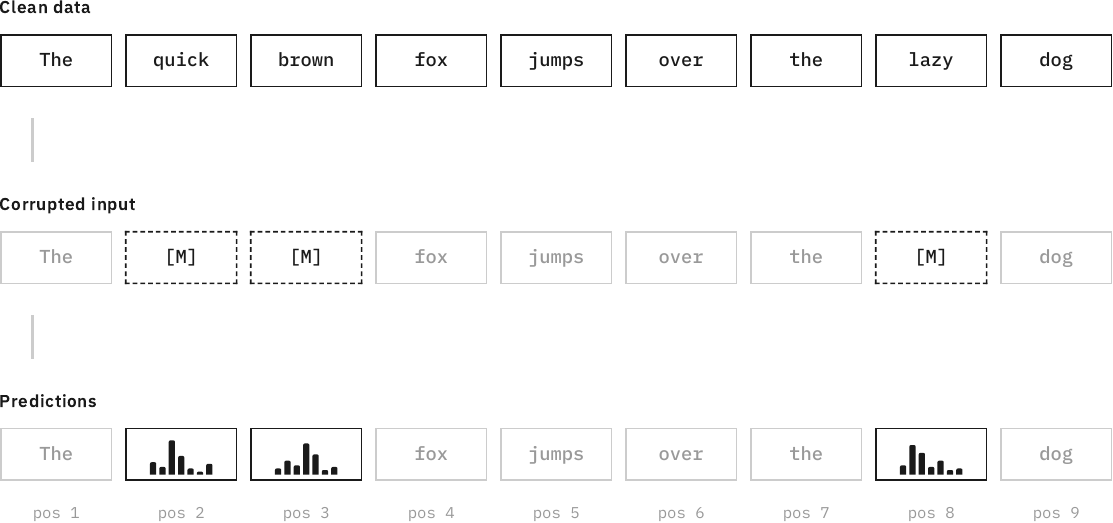}
\caption{Training a discrete diffusion model.  The forward corruption process masks a subset of tokens in a clean sequence.  The model predicts a distribution over the vocabulary at each masked position; the training objective maximizes the likelihood of the original tokens.}
\label{fig:diffusion-training}
\end{figure}

Training a discrete diffusion model requires defining a forward corruption process~\cite{austin2021structured}.
Given a clean sequence, the process samples a masking rate $t \in [0, 1]$ and independently replaces each token with a mask token with probability $t$.
The model learns to reverse this process: it takes a partially masked sequence and the masking rate $t$ and predicts every masked token simultaneously.
The training objective maximizes the log-likelihood of each masked token given the unmasked tokens~\cite{sahoo2024simple,shi2024simplified} (Figure~\ref{fig:diffusion-training}).

Generation initializes the response tokens as fully masked and iteratively unmasks them over $T$ denoising steps.
At each step, the model runs one forward pass to predict a distribution over the vocabulary at every still-masked position.
The decoding algorithm then selects a subset of masked positions and unmasks them by sampling from the predicted distributions.
The process repeats with fewer masked positions until no masks remain.
To decode a sequence of $n$ tokens, autoregressive decoding requires $n$ forward passes; a discrete diffusion model requires $\lceil n/k \rceil$, where $k$---the number of tokens to unmask per step---is a hyperparameter that controls the quality-latency trade-off.

\begin{figure}[t]
\centering
\begin{subfigure}[t]{\textwidth}
  \centering
  \includegraphics[width=\textwidth]{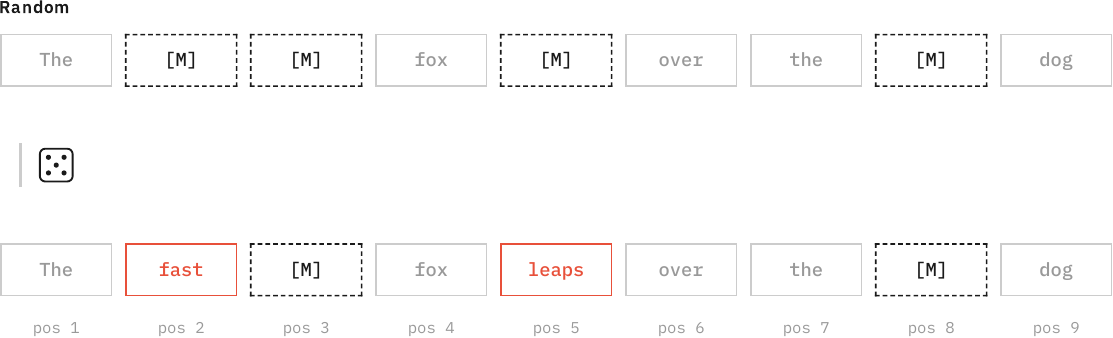}
  \caption{Random denoising order.  The decoding algorithm selects masked positions uniformly at random, regardless of the model's predictions.}
  \label{fig:denoising-random}
\end{subfigure}
\vspace{0.5em}
\begin{subfigure}[t]{\textwidth}
  \centering
  \includegraphics[width=\textwidth]{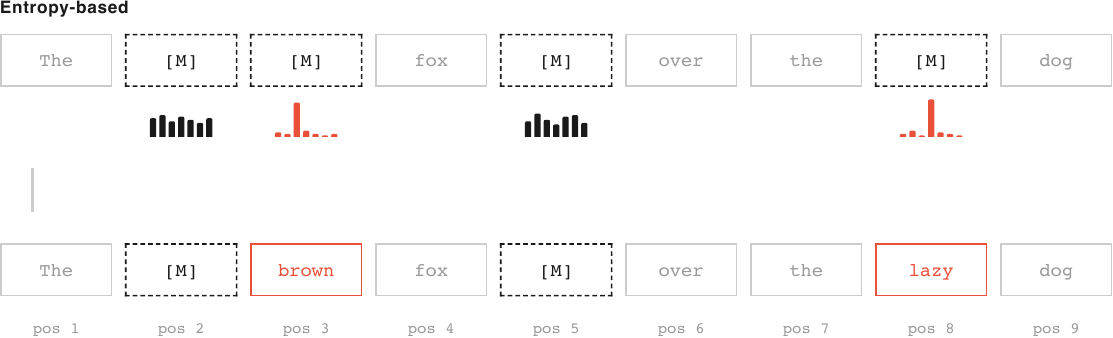}
  \caption{Entropy-based denoising order.  The algorithm unmasks positions where the model's predicted distribution has the lowest entropy (most peaked); high-entropy positions remain masked.}
  \label{fig:denoising-entropy}
\end{subfigure}
\caption{Two denoising orders applied to the same partially masked sequence.  Red marks positions that the algorithm unmasks at this step; dashes mark positions that remain masked.}
\label{fig:denoising-orders}
\end{figure}

The forward corruption process defines how training constructs noisy inputs, but it does not prescribe the order in which the decoding algorithm unmasks tokens during inference.
This choice---which positions to unmask at which step---is the \emph{denoising order}.
A random denoising order selects positions uniformly at random at each step.
An entropy-based denoising order selects the positions where the model's predicted distribution has the lowest entropy, preferring positions about which the model is most certain~\cite{ye2025dream}.
Figure~\ref{fig:denoising-orders} illustrates the difference: the random order unmasks positions 2 and 5 without considering model predictions, while the entropy-based order unmasks positions 3 and 8 where the model's distributions are most peaked.
The denoising order determines which tokens the model can condition on when predicting later tokens, and directly affects generation quality.

\section{Linear Attention}
\label{sec:prelim:linear-attn}

The KV cache grows because softmax attention needs access to every past key-value pair.
For a query $\mathbf{q}$ against $L$ past keys $\mathbf{K} \in \mathbb{R}^{L \times d}$ and values $\mathbf{V} \in \mathbb{R}^{L \times d}$, where $d$ is the head dimension, softmax attention computes
\begin{equation}
\mathrm{Attn}(\mathbf{q}, \mathbf{K}, \mathbf{V}) = \frac{\sum_{i=1}^{L} \exp(\mathbf{q}\, \mathbf{k}_i^\top / \sqrt{d}) \, \mathbf{v}_i}{\sum_{i=1}^{L} \exp(\mathbf{q}\, \mathbf{k}_i^\top / \sqrt{d})}.
\label{eq:softmax-attn}
\end{equation}
Each new query must interact with all $L$ past key-value pairs to produce a single output.  Recomputing those key-value pairs from the input at every decoding step would be prohibitively expensive, so the model stores them in a KV cache that grows by one entry per token.

Katharopoulos et al.~\cite{katharopoulos2020transformers} formulate softmax attention as a special case of a more general form with an arbitrary similarity function $\mathrm{sim}$:
\begin{equation}
\mathrm{Attn}_{\mathrm{sim}}(\mathbf{q}, \mathbf{K}, \mathbf{V}) = \frac{\sum_{i=1}^{L} \mathrm{sim}(\mathbf{q}, \mathbf{k}_i) \, \mathbf{v}_i}{\sum_{i=1}^{L} \mathrm{sim}(\mathbf{q}, \mathbf{k}_i)}.
\label{eq:gen-attn}
\end{equation}
Softmax attention uses $\mathrm{sim}(\mathbf{q}, \mathbf{k}) = \exp(\mathbf{q}\,\mathbf{k}^\top / \sqrt{d})$.
If instead the similarity function factors through a feature map $\phi$ that acts on the query and key independently---$\mathrm{sim}(\mathbf{q}, \mathbf{k}) = \phi(\mathbf{q})^\top \phi(\mathbf{k})$---then substituting into Equation~\ref{eq:gen-attn} gives
\begin{equation}
\mathrm{LinAttn}(\mathbf{q}, \mathbf{K}, \mathbf{V}) = \frac{\sum_{i=1}^{L} \phi(\mathbf{q})^\top \phi(\mathbf{k}_i) \, \mathbf{v}_i}{\sum_{i=1}^{L} \phi(\mathbf{q})^\top \phi(\mathbf{k}_i)}.
\label{eq:phi-attn}
\end{equation}
Because matrix multiplication is associative, I can regroup to factor $\phi(\mathbf{q})$ out of both sums.
Rather than scoring the query against each of the $L$ keys individually, the model first accumulates two summary statistics over the keys and values:
\begin{align}
\mathbf{S} &= \textstyle\sum_{i=1}^{L} \phi(\mathbf{k}_i)\, \mathbf{v}_i^\top \in \mathbb{R}^{d \times d}, &
\mathbf{z} &= \textstyle\sum_{i=1}^{L} \phi(\mathbf{k}_i) \in \mathbb{R}^{d}.
\label{eq:linear-attn-state}
\end{align}
Katharopoulos et al.~\cite{katharopoulos2020transformers} call $\mathbf{S}$ the \emph{attention memory} and $\mathbf{z}$ the \emph{normalizer memory}.
Given these, the output for any query is
\begin{equation}
\mathrm{LinAttn}(\mathbf{q}, \mathbf{S}, \mathbf{z}) = \frac{\phi(\mathbf{q})^\top \mathbf{S}}{\phi(\mathbf{q})^\top \mathbf{z}},
\label{eq:linear-attn}
\end{equation}
a pair of matrix-vector products independent of $L$.

During autoregressive decoding, the model maintains $\mathbf{S}$ and $\mathbf{z}$ as a running recurrence, updating each with the new key-value pair at every step.
This recurrence replaces the KV cache: instead of storing every past key-value pair ($O(Ld)$ memory that grows with each token), the model compresses all past context into a fixed-size state---$O(d^2)$ for $\mathbf{S}$ and $O(d)$ for $\mathbf{z}$---regardless of how many tokens it has generated.

\section{StreamingLLM}
\label{sec:prelim:streamingllm}

Xiao et al.~\cite{xiao2024streamingllm} observe that the first few tokens in a sequence consistently receive large attention scores across layers and heads, even when those tokens carry no special semantic content.
Xiao et al.\ call these positions \emph{attention sinks}.
Removing these tokens' intermediate states from the KV cache causes perplexity to spike.

\begin{figure}[t]
\centering
\includegraphics[width=\textwidth]{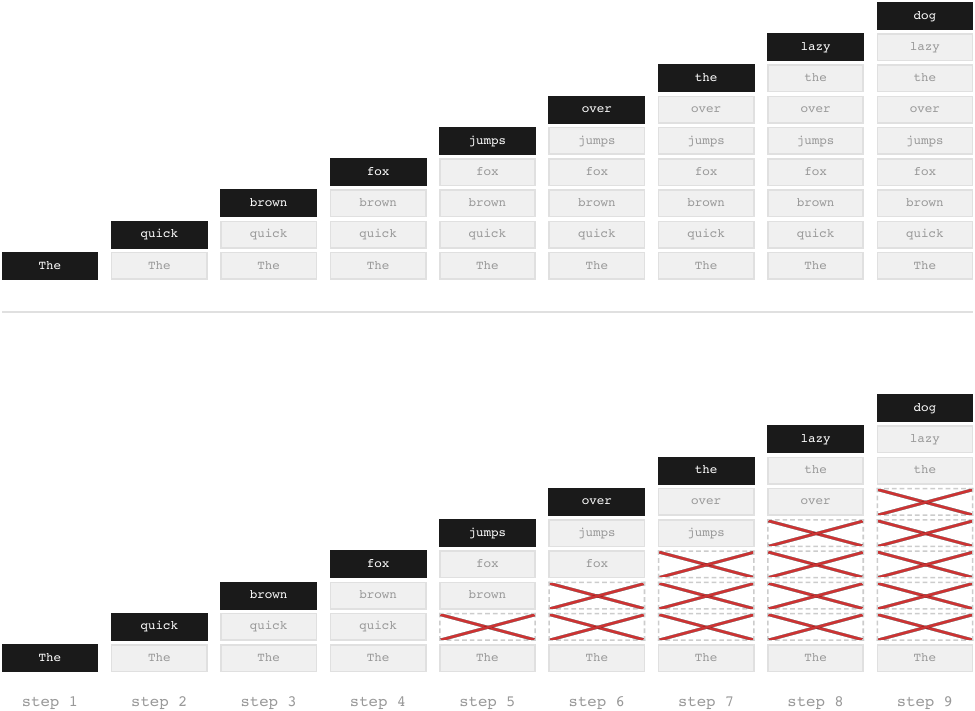}
\caption{KV cache contents at each decoding step.  Top: the standard cache grows with every token.  Bottom: StreamingLLM retains the initial sink token and a fixed window of recent tokens, evicting intermediate entries (crossed out).}
\label{fig:streamingllm}
\end{figure}

StreamingLLM addresses this by always retaining the first few tokens---the attention sinks---plus a sliding window of the most recent tokens in the KV cache (Figure~\ref{fig:streamingllm}).
This fixed-size cache maintains stable perplexity over arbitrarily long streams.
The eviction policy is entirely position-based: it discards all tokens between the initial sinks and the recent window, regardless of their content.

\section{Summary and Design Desiderata}
\label{sec:prelim:summary}

The preceding sections review techniques that each address a real bottleneck in autoregressive inference, and their designs inspire the following three desiderata that I argue any good inference optimization technique should have.

\begin{enumerate}
  \item \emph{Input-adaptive.} The technique makes data-dependent decisions at runtime, exploiting efficiency-improving opportunities in each input rather than applying a fixed strategy.
  \item \emph{Learnable.} The policy governing these decisions admits a trainable parameterization that can improve with data, rather than relying on hand-crafted heuristics.
  \item \emph{Inspectable.} The decisions are discrete and observable, enabling practitioners to inspect, debug, and understand the system's behavior.
\end{enumerate}

Table~\ref{tab:prelim-summary} evaluates each technique against these three desiderata. No existing technique satisfies all three.

\begin{table}[t]
\centering
\caption[Summary of inference techniques that this chapter reviews]{Summary of inference techniques that this chapter reviews.  \cmark\ = satisfies the property, \xmark\ = does not satisfy.}
\label{tab:prelim-summary}
\begin{tabular}{lccc}
\toprule
Technique & Input-adaptive & Learnable & Inspectable \\
\midrule
Speculative decoding (\S\ref{sec:prelim:spec}) & \xmark & \xmark & \cmark \\
Asynchronous decoding (\S\ref{sec:prelim:async}) & \cmark & \xmark & \cmark \\
Discrete diffusion (\S\ref{sec:prelim:diffusion}) & \cmark & \xmark & \cmark \\
Linear attention (\S\ref{sec:prelim:linear-attn}) & \xmark & \xmark & \xmark \\
StreamingLLM (\S\ref{sec:prelim:streamingllm}) & \xmark & \xmark & \cmark \\
\midrule
This thesis & \cmark & \cmark & \cmark \\
\bottomrule
\end{tabular}
\end{table}

Among these techniques, asynchronous decoding---SoT and APAR---stands out as an inspiring emerging line of work: it begins to use content semantics---the meaning of what the model generates---to guide inference optimization.
Yet it both falls short on parallelization, relying on fixed heuristics rather than learnable policies, and addresses only a single bottleneck when the same principle could bring impact to a broad range of problems.
This thesis develops a unified design methodology centered around semantic dependence that brings input-adaptive, learnable, and inspectable solutions to a broad range of efficiency challenges in LLM inference.

PASTA uses semantic dependence to parallelize autoregressive decoding, training the model to annotate which output chunks can generate independently. TIP uses semantic dependence to evict intermediate reasoning steps from the KV cache, reducing memory consumption while maintaining accuracy. Planned Diffusion uses semantic dependence to derive a denoising order for discrete diffusion, autoregressively generating a plan that specifies which chunks to denoise in parallel\footnote{Planned Diffusion's decomposition plan is learnable in principle---the plan tokens are model outputs amenable to reinforcement learning---but I leave designing such a learning algorithm as future work, as how best to apply RL to discrete diffusion models remains an open question.}.

\chapter{Learning to Keep a Promise: Scaling Language Model Decoding Parallelism with Learned Asynchronous Decoding}
\label{ch:pasta}

This chapter presents PASTA (\underline{PA}rallel \underline{ST}ructure \underline{A}nnotation), a system that addresses the parallel generation bottleneck.
PASTA trains LLMs to annotate semantic dependence among chunks of their output, enabling chunks without dependences to generate in parallel.

\section{Introduction}
\label{sec:pasta:intro}

Autoregressive decoding is a fundamental efficiency bottleneck in LLM inference.
Contemporary LLMs routinely require multiple seconds or even minutes of decoding time to complete user requests \cite{jiang2024mixtralexperts,touvron2023llamaopenefficientfoundation,openai2024openaio1card,deepseekai2025deepseekr1incentivizingreasoningcapability}.
This latency stems from the sequential nature of autoregressive decoding, which leads to inefficient hardware utilization during inference.
At small batch sizes, inference typically achieves less than 20\% Model Flops Utilization (MFU) \cite{pope2022efficientlyscalingtransformerinference}.

\paragraph{Parallelism from Semantics.}
Recent works like Skeleton-of-Thought (SoT) \cite{ning2023skeleton} and APAR \cite{liu2024apar} observe that not all \emph{chunks}, contiguous sequences of tokens, in an LLM response are semantically dependent on each other, and decode those that are not in parallel, a strategy I refer to as \emph{asynchronous decoding}.
Given a request, SoT first produces a bullet-point outline, then applies regular-expression-based syntactic pattern matching to extract points that it then expands in parallel.
APAR, in contrast, applies regular-expression-based syntactic pattern matching on training data to identify structures like lists and paragraphs, and finetunes an LLM to decode in parallel the item descriptions given list items and the paragraph bodies given the first sentences.

\begin{figure}[htbp]
    \centering
    \includegraphics[
  width=0.95\linewidth,
  trim=25 610 20 25,
  clip
]{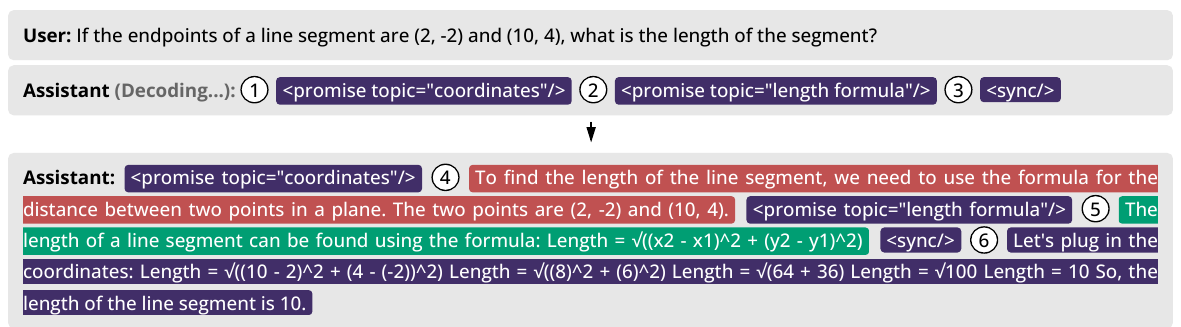}
\caption{
Example response from a \pasta{} model that the \lang{} runtime executes.
The runtime begins with only the main thread.
It first decodes \CircledText{1}, and it creates an asynchronous decoding thread, which decodes \CircledText{4} in red.
In parallel, the main thread decodes \CircledText{2}.
It creates another asynchronous decoding thread, which contains both the \promise{} tag on coordinates extraction and the \promise{} tag on length formula in its prefix, and decodes \CircledText{5} in green.
The main thread continues decoding in parallel to both threads to get \CircledText{3}.
It waits at this point until all other threads complete.
The runtime then inserts each asynchronous content after their corresponding \promise{} tags.
Finally, the runtime decodes \CircledText{6}, with both of the asynchronously decoded content in the prefix.
}
\label{fig:pasta:example}
\end{figure}

While such opportunities for parallelization broadly exist in LLM responses, relying purely on syntactic heuristics---manifested as hand-crafted regular expressions---to identify them has limitations.
First, these heuristics lack scalability, requiring manual engineering to capture more parallelism, even as more training compute becomes available.
Second, they lack robustness, failing to detect such opportunities when responses deviate from expected patterns, even by a missing punctuation mark.
These limitations motivate a learning-based approach to optimize LLMs' ability to annotate semantic dependence, enabling LLMs to find parallelization opportunities beyond fixed patterns.

\paragraph{Learned Asynchronous Decoding.}
I present \pasta{}, a system that teaches LLMs to annotate semantic dependence in their own responses.
The system consists of an annotation language that extends the model's vocabulary for marking semantic dependence, a co-designed runtime that acts on these annotations to orchestrate parallel decoding, and a training procedure that optimizes LLMs' ability to annotate semantic dependence.
Through this system, LLMs direct their own asynchronous decoding process.
In \Cref{fig:pasta:example}, I show how these components implement asynchronous decoding.

\paragraph{Annotations.}
The annotation language, \lang{} (\underline{PA}rallel \underline{ST}ructure \underline{A}nnotation \underline{LANG}uage), enables LLMs to annotate semantic dependence in their responses.
In \Cref{fig:pasta:example}, I show a \lang{}-annotated response.
The \promise{} tags serve as placeholders for content chunks that do not depend on each other semantically, such as extracting coordinates (Tag \CircledText{1}) and recalling the line segment length formula (Tag \CircledText{2}).
Each \promise{} tag includes a \topic{} attribute that concisely describes the chunk.
When further decoding steps require conditioning on tokens that other threads have not yet finished decoding, the LLM issues a \sync{} tag to indicate so, as shown at \CircledText{3} in \Cref{fig:pasta:example}.

\paragraph{Runtime.}
I develop the \lang{} runtime, which acts on \lang{} annotations to orchestrate asynchronous decoding during inference.
It launches parallel decoding threads for contents corresponding to \promise{} tags, which do not semantically depend on each other, and enforces semantic dependence at \sync{} tags.
The runtime simultaneously decodes multiple non-contiguous token chunks from the LLM, improving overall decoding latency.

\paragraph{Finetuning.}
Training an LLM to generate \lang{} annotations starts with two manual inputs: seven human-crafted demonstrations and a description of the \lang{} annotation language.
Prompting the Gemini 1.5 Flash model \cite{geminiteam2024geminifamilyhighlycapable} with these manual inputs, the \pasta{} system initiates an automated two-stage finetuning process.
In the first stage, \pasta{} uses the prompted Gemini model to create the \pastasft{} dataset by annotating the SlimOrca instruction-finetuning dataset \cite{lian2023mistralslimorca1} with \lang{} annotations that describe the semantic dependence between token chunks.
\pasta{} then finetunes an LLM on this dataset to produce a model that generates \lang{} annotations.

In the second stage, \pasta{} creates another dataset by sampling the finetuned LLM and scoring each output based on its quality and latency.
Unlike traditional uses of preference optimization to improve response quality \cite{gui2024bonbonalignmentlargelanguage,rafailov2023direct}, I adapt one such algorithm for \pasta{} to optimize for both output quality and latency.
\pasta{} applies preference optimization to the finetuned LLM on this dataset to produce a model with improved output quality and latency.
This second stage of finetuning features a quality weight hyperparameter that controls the trade-off between quality and speedup.
Through repeated iterations of the second stage, \pasta{} creates models that respond with increasingly better quality and lower latency.

\paragraph{Results.}
Varying the quality weight hyperparameter, \pasta{} produces a suite of models with different quality-latency trade-offs.
I evaluate these models on 805 representative instruction-following prompts from AlpacaEval \cite{alpaca_eval,dubois2024length}.
After one iteration of preference optimization, these models Pareto-dominate all existing asynchronous decoding methods.
Additional iterations of preference optimization further improve the quality-latency Pareto frontier, showing no signs of saturation even after two iterations.
The results demonstrate geometric mean speedups ranging from 1.21$\times$ to 1.93$\times$\footnote{Practitioners should use geometric mean to compute normalized values \cite{fleming1986not}. However, the prevailing practice in parallel decoding literature uses arithmetic averaging when reporting speedup, which would show this result as 1.57--2.6$\times$.} with corresponding quality changes of +2.2\% to $-$7.1\% respectively, measured as length-controlled win rates.

\paragraph{Contributions.}
I make the following contributions:
\begin{itemize}[leftmargin=*, topsep=0pt, itemsep=0pt]
    \item I design \lang{} to be an annotation language that enables LLMs to annotate semantic dependence between chunks of tokens in their own responses.
    \item I implement a \lang{} runtime that efficiently orchestrates asynchronous decoding based on \lang{} annotations at inference time.
    \item I develop a two-stage finetuning technique that trains LLMs to identify diverse patterns of semantic dependence in their output and express them through \lang{} annotations, while directly optimizing for both response quality and inference speedup.
    \item I evaluate the method on AlpacaEval \cite{alpaca_eval,dubois2024length}, a suite of 805 representative instruction-following prompts, and find that the method Pareto-dominates all existing asynchronous decoding methods in terms of quality and speedup.
\end{itemize}

\paragraph{Implication.}
\pasta{} demonstrates that preference optimization can incorporate latency objectives alongside quality during LLM post-training.
This result establishes learned annotation of semantic dependence as a scalable approach to reducing decoding latency.

\section{Synchronous and Asynchronous Decoding}
\label{sec:pasta:taxonomy}

To provide context for how \lang{} relates to other parallel decoding techniques for accelerating LLM decoding, I present a dichotomy of parallel decoding techniques.
A given parallel decoding strategy can fall into either \emph{synchronous} or \emph{asynchronous} decoding.
In synchronous decoding, only a single chunk actively decodes at any time while the rest of the generation halts.
In contrast, during asynchronous decoding, multiple chunks of the language model's output decode independently in parallel.

\paragraph{Synchronous Decoding.}
I consider speculative decoding \cite{leviathan2023fast,chen2023accelerating,stern2018blockwise,cai2024medusa,ankner2024hydra,he2023rest,fu2024break,spector2023accelerating,santilli2023accelerating} as a prototypical example of synchronous decoding.
It decodes multiple tokens within a single chunk in parallel, but must complete that chunk before moving on to any subsequent tokens.

\paragraph{Asynchronous Decoding.}
In contrast, works such as SoT \cite{ning2023skeleton} and APAR \cite{liu2024apar} implement asynchronous decoding techniques.
Both methods enable decoding to jump ahead in the output sequence and generate tokens before previous positions complete, resulting in multiple chunks of the output decoding in parallel.
While the \pasta{} system also implements asynchronous decoding, it improves upon previous works by employing a learning-based system to identify parallelization opportunities, instead of relying on human-defined heuristics.
By training an LLM to identify and exploit parallelization opportunities, \pasta{} Pareto-dominates previous asynchronous decoding techniques in quality and speedup.

\section{\lang{} and the Runtime}
\label{sec:pasta:language-design}

\lang{} is an XML-like annotation language that a language model uses to express semantic dependence in its own response.

\paragraph{Syntax.}
\lang{} defines three tags: a paired \async{} tag that wraps asynchronous content chunks, a \promise{} tag that declares each chunk before it appears, and a \sync{} tag that marks a synchronization point across all active threads. A \promise{} tag carries two attributes: \topic{}, a string describing the chunk's content, and \tokens{}, an integer estimating the chunk length in multiples of 10.

\begin{figure}[htbp]
    \centering
    \includegraphics[
  width=0.95\linewidth,
  trim=10 140 10 5,
  clip
]{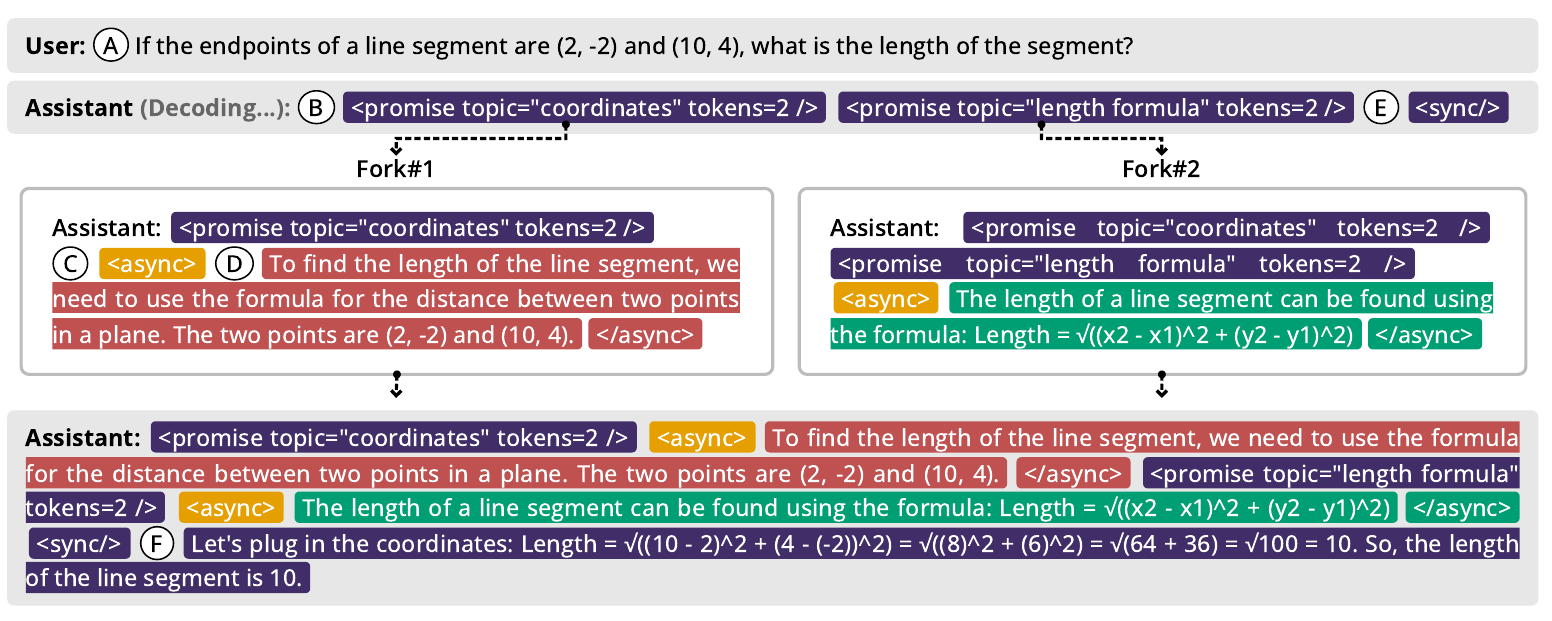}
    \caption{\lang{} runtime orchestrates parallel decoding.
    \CircledText{A} shows the user prompt.
    \CircledText{B} shows the \promise{} tag which initiates the first asynchronous decoding thread named ``Fork\#1''.
    \CircledText{C} indicates where the runtime appends an \async{} tag to the prefix of Fork\#1 with topic ``coordinates''.
    \CircledText{D} denotes the asynchronous generation by Fork\#1.
    \CircledText{E} shows the \sync{} tag where the runtime pauses to wait for all asynchronous generations.
    \CircledText{F} shows the main thread decodes the remaining content with both asynchronous generations in its prefix.
    Color shows the identity of the decoding thread (purple=main, red=Fork\#1, green=Fork\#2); orange denotes runtime-inserted tokens.}
    \label{fig:pasta:interpreter-detail}
\end{figure}

\paragraph{Runtime.}
The \lang{} runtime is the component of the PASTA system that executes \lang{} tags: a \lang{}-equipped language model initiates asynchronous decoding by generating these tags, and the runtime acts on them. I describe here the functionality of each tag and how the runtime uses them.
\Cref{fig:pasta:interpreter-detail} shows how the runtime orchestrates asynchronous decoding.

With the user query \CircledText{A} as the prefix,
the runtime decodes sequentially until encountering a \promise{} tag at tag \CircledText{B}.
The \topic{} attribute indicates the topic of the chunk that the runtime decodes asynchronously, and the \tokens{} attribute estimates the number of tokens in multiples of 10 in the \async{} tag.
Since the main thread does not condition on asynchronous content until \sync{}, these attributes provide a brief description of what each fork will produce.
The runtime then initiates a new asynchronous decoding thread named ``Fork\#1''.
The main thread continues decoding in parallel, while the new thread first appends an \async{} tag to its prefix at \CircledText{C} and then decodes content matching the specified \topic{} until reaching a \casync{} tag at \CircledText{D}.
The main thread proceeds without conditioning on any asynchronously decoded content.
Only upon encountering a \sync{} tag at \CircledText{E} does the runtime pause to wait for all asynchronous decoding threads to complete, and then enable the language model to condition on asynchronously decoded content for decoding subsequent tokens in \CircledText{F}.

\begin{figure}[htbp]
    \centering
    \includegraphics[width=.95\linewidth, clip, trim={20px 0px 20px 0px}]{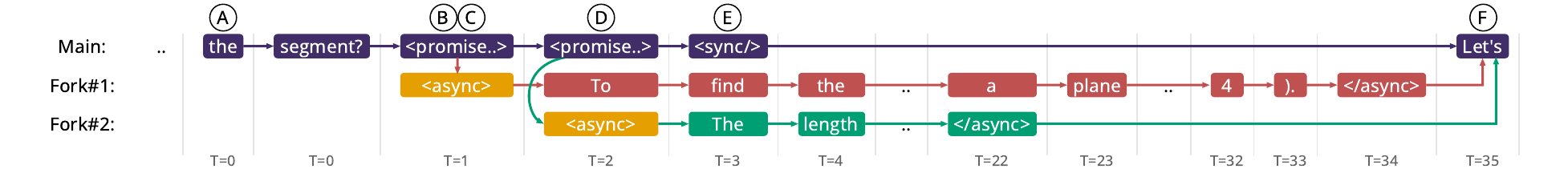}
    \caption{Decoding parallelism and attention patterns at each timestamp. At each timestamp, I show the tokens decoded in parallel in that timestamp. The directed edges between tokens show the attention relationships: each edge connects a token to the very next token that may attend to it. A token can then attend to any ancestor reachable by following these edges backward. \CircledText{A} shows the last few tokens of the user query. \CircledText{B} shows when the runtime decodes a \promise{} token, after which it immediately appends an \async{} token for Fork\#1 at \CircledText{C}. Subsequently at \CircledText{D}, Fork\#1 begins asynchronous decoding, while in parallel, the runtime creates another decoding thread (Fork\#2). At \CircledText{E}, when encountering the \sync{}, the runtime pauses the main thread until all asynchronous threads complete. Finally at \CircledText{F}, the main thread resumes decoding with both asynchronously decoded content in its prefix.
    Color shows the identity of the decoding thread (purple=main, red=Fork\#1, green=Fork\#2); orange denotes runtime-inserted tokens.}
    \label{fig:pasta:parallelism}
\end{figure}

\begin{figure}[htbp]
    \centering
    \includegraphics[width=\linewidth, clip, trim={20px 510px 0px 0px}]{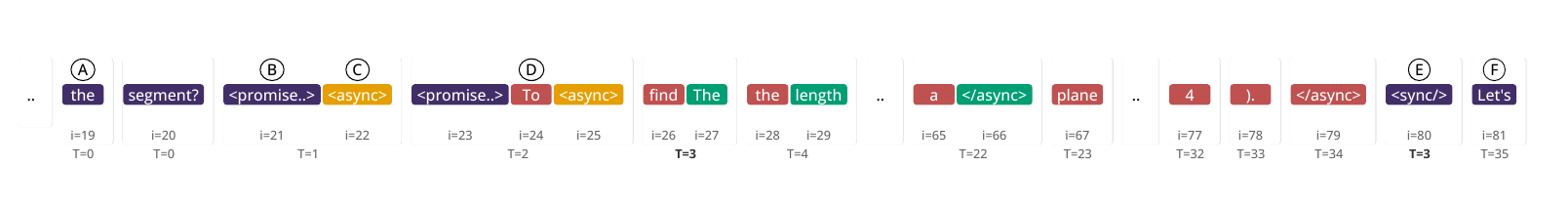}
    \caption{KV cache layout throughout parallel decoding. At \CircledText{A}, the last few tokens of the user prompt, the KV cache layout is contiguous. Starting at \CircledText{B}, the KV cache interleaves between threads, while inserting the corresponding \async{} token for the new thread at \CircledText{C}. \CircledText{D} shows parallel decoding in progress, with tokens from Fork\#1 generating while the main thread continues decoding. At \CircledText{E}, the \sync{} token (which the main thread decodes at T=3) enters the KV cache. After synchronization, the KV cache returns to a contiguous layout at \CircledText{F}.
    Color shows the identity of the decoding thread (purple=main, red=Fork\#1, green=Fork\#2); orange denotes runtime-inserted tokens.}
    \label{fig:pasta:kv-cache-management}
\end{figure}

\paragraph{Efficiency.}
The runtime implementation addresses several key challenges in efficient asynchronous decoding, with KV cache management being the core issue.
Since ML compilers often require static tensor shapes for effective compilation \cite{50530,paszke2019pytorchimperativestylehighperformance}, I assume a fixed batch size and sequence length.\footnote{I use a batch size of 1 and max sequence length of 2048 as in \textcite{gpt-fast}.}
Implementing asynchronous decoding naively as batched decoding presents two suboptimal options: 1) allocate differently sized KV cache pools and switch between them upon thread creation and termination, which wastes precious accelerator memory, or 2) pre-allocate a fixed number of decoding threads, leading to wasted memory and computation due to inactive threads.
I further discuss the drawbacks of naive option 2) in \Cref{sec:pasta:naive-interpreter}.
Instead, I store KV cache from all asynchronous decoding threads in a single pre-allocated memory pool where threads interleave their KV cache entries.

\Cref{fig:pasta:parallelism,fig:pasta:kv-cache-management} illustrate the approach.
I denote decoding timestamp with T and index in the KV cache pool with i.
When only one thread is active at T=0 (\CircledText{A}), it appends new KV cache sequentially to the pool at i=19--20.
When the main thread (purple) decodes a \promise{} token at T=1 (\CircledText{B}), the runtime immediately appends an \async{} token to signal the start of a new thread Fork\#1 (\CircledText{C}).
At T=2 (\CircledText{D}), Fork\#1 begins asynchronous decoding, while the runtime initiates another asynchronous decoding thread Fork\#2.
At T=3 (\CircledText{E}), the runtime decodes 3 tokens in parallel from 3 active threads. The \sync{} token does not enter the KV cache pool immediately: it signals the main thread to pause and wait for the other threads to complete. The remaining two tokens enter the KV cache pool in neighboring positions (i=26--27).
The two threads (green/red) continue to decode in parallel, taking turns appending tokens to the same KV cache pool (T=3--22). This interleaving allows both threads to share a single memory pool.

To prevent cross-thread interference with this interleaved KV cache layout, I use attention masks to ensure threads cannot attend to each other's tokens before synchronization. \Cref{fig:pasta:parallelism} visualizes these attention relationships: each directed edge connects a token to the very next token that may attend to it, and a token can attend to any ancestor reachable by following these edges backward.
A token in an asynchronous thread can only attend to tokens within its own thread and tokens from the main thread that existed before the runtime spawned the thread. For example, the token \texttt{find} (at T=3, red) attends to \texttt{To} (at T=2, red) and \texttt{segment?} (at T=0, purple), but cannot attend to the second \async{} token (at T=2, orange), the second \promise{} token (at T=2, purple), or the \sync{} token (at T=3, purple).

Once the runtime decodes the \sync{} token in the main thread at T=3 (\CircledText{E}), it pauses the main thread to synchronize, waiting for both forks to complete: Fork\#1 and Fork\#2 decode their \casync{} tokens at T=34 and T=22 respectively. After the wait is over at T=34, the runtime inserts the \sync{} token into the KV cache pool at i=80, and the main thread resumes decoding (\CircledText{F}) while conditioning on both forks' asynchronous generations.

While prior works like radix attention \cite{zheng2023efficiently} enable multiple decoding threads to share attention to a common prefix, PASTA additionally enables a single decoding thread to attend to multiple asynchronously decoded threads as described above.

\subsection{Naive Interpreter Implementation}
\label{sec:pasta:naive-interpreter}
\begin{figure}[p]
    \centering
    \begin{subfigure}{\linewidth}
        \centering
        \includegraphics[width=\linewidth, trim=10px 140px 10px 5px, clip]{figs/pasta/forking.html.pdf}
        \caption{\lang{} interpreter orchestrates parallel decoding.
        \CircledText{A} shows the user query.
        \CircledText{B} shows the \promise{} tag which initiates the first asynchronous decoding thread named ``Fork\#1''.
        \CircledText{C} indicates where the interpreter appends an \async{} tag to the prefix of Fork\#1, signaling Fork\#1 should complete the promised content with topic ``coordinates''.
        \CircledText{D} denotes the asynchronous generation by Fork\#1.
        \CircledText{E} shows the \sync{} tag where the interpreter pauses to wait for all asynchronous generations.
        \CircledText{F} shows the main thread decodes the remaining content with both asynchronous generations in its prefix.
        }
        \label{fig:pasta:interpreter-detail-refresh}
    \end{subfigure}
    \begin{subfigure}[t]{\linewidth}
        \centering
        \includegraphics[width=\linewidth, trim=5 540 225 5, clip]{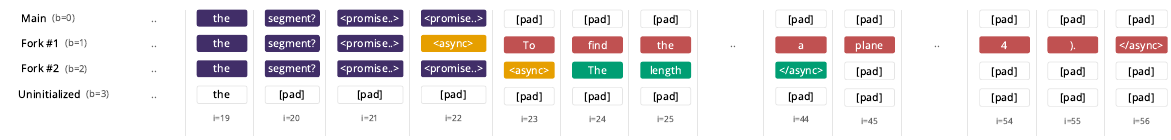}
        \subcaption{Naive KV cache layout before synchronization. The naive approach implements asynchronous decoding through batched decoding, with a pre-allocated fixed-size KV cache pool supporting a maximum number of parallel threads, which I set to 4 in this figure.
        Each decoding thread operates as an independent batch item, with shared prefix duplicated for each thread.
        The runtime stores the KV cache pool in row-major layout with shape (max batch size $\times$ sequence length), shown here before synchronization.
        The figure shows KV cache pool contents immediately after decoding \sync{} at \CircledText{E}.}
        \label{fig:pasta:naive-before}
    \end{subfigure}
    \vspace*{3em}
    \begin{subfigure}[t]{\textwidth}
    \centering
    \includegraphics[width=\linewidth, trim=5 520 110 -10, clip]{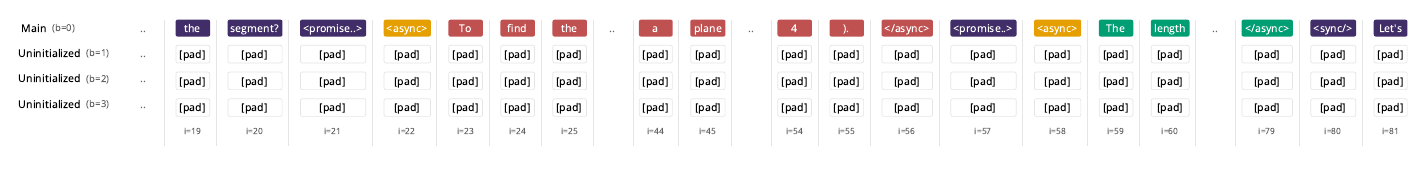}
    \subcaption{Naive KV cache layout after synchronization. For asynchronous content to be available to subsequent decoding steps, the naive interpreter must copy and insert each thread's KV cache (shown in red and green) after their corresponding \promise{} tokens in the main thread's KV cache row. After insertion, the interpreter adds the \sync{} token and subsequent decoding can attend to the asynchronous content. The interpreter must then mark rows corresponding to terminated threads (Fork \#1, Fork \#2) as uninitialized to free memory. The figure shows the KV cache pool contents two steps into \CircledText{F}.}
    \label{fig:pasta:naive-after}
    \end{subfigure}
    \caption{Example Naive Interpreter Implementation.}
    \label{fig:pasta:naive-figures}
\end{figure}

In this section, I illustrate the KV cache layout for the naive implementation before and after synchronization in \Cref{fig:pasta:naive-before} and \Cref{fig:pasta:naive-after}.
I repeat the full model output in \Cref{fig:pasta:interpreter-detail-refresh}.
Here I illustrate the inefficiencies associated with implementing asynchronous decoding as batched decoding, where I pre-allocate the KV cache pool assuming a fixed number of decoding threads of 4.

In \Cref{fig:pasta:naive-before}, I show the KV cache content before synchronization at \CircledText{E}. The naive batched implementation requires duplicating the prefix for each asynchronous thread, which runs as an independent batch item. Since the naive implementation sizes the KV cache pool for the maximum possible number of parallel threads, many rows often remain unused. This wastes both accelerator memory and computation, as the KV cache pool's shape determines the shape of attention computation.

In \Cref{fig:pasta:naive-after}, I show the KV cache content after synchronization, two decoding steps into \CircledText{F}. The naive interpreter must copy and insert Fork\#1 and Fork\#2's asynchronous generations after their corresponding \promise{} tags in the main thread's KV cache row, then mark the rows of terminated threads as uninitialized. This naive implementation suffers from wasted accelerator memory during allocation, wasted computation from oversized attention, and overhead from KV cache movement during synchronization.

\section{Training}
\label{sec:pasta:training}

\begin{figure}[htbp]
    \centering
    \includegraphics[width=0.45\linewidth]{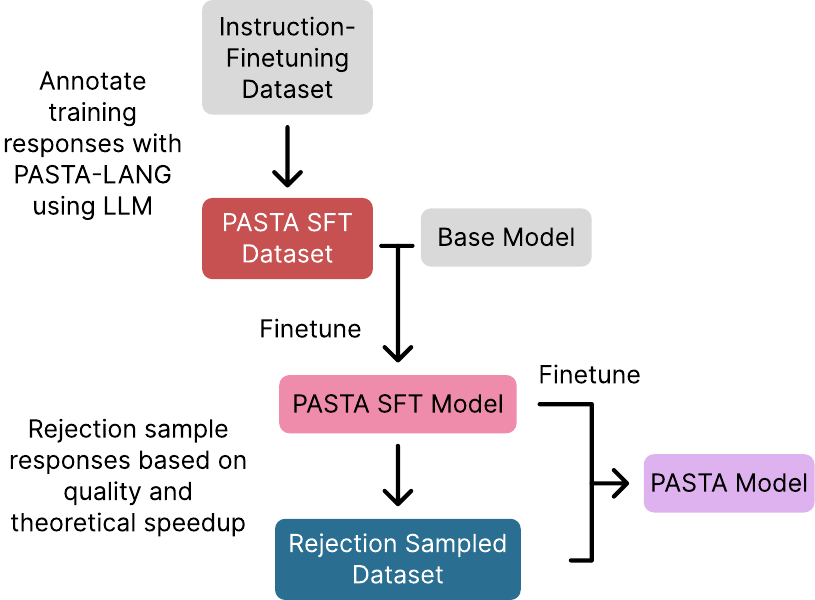}
    \caption{\lang{} dataset creation and model training.}
    \label{fig:pasta:flow}
\end{figure}

I present a two-stage finetuning process that trains an LLM to annotate semantic dependence in its own responses using \lang{}.
\Cref{fig:pasta:flow} illustrates the \pasta{} system pipeline for dataset construction and model training to produce a \lang{}-equipped model.

\paragraph{Building a \pastasft{} Dataset.}
The \pasta{} system begins by constructing an initial finetuning dataset, which I refer to as the \pastasft{} dataset.
It prompts Gemini 1.5 Flash \cite{geminiteam2024geminifamilyhighlycapable} to add \lang{} annotations to responses from an instruction-following dataset.
It provides Gemini 1.5 Flash with 7 human-annotated examples and a description of the syntax and semantics of \lang{}, and labels a 100K response subset of the SlimOrca dataset \cite{lian2023mistralslimorca1}.\footnote{Due to API limitations, I only successfully annotated 87K/100K instruction-response pairs as the \pastasft{} dataset.}
The annotation prompt is available in \Cref{app:pasta:dataset-prompts}.

\paragraph{Training a \pastasft{} Model.}
\pasta{} then finetunes the base LLM on the \pastasft{} dataset, producing what I call the \pastasft{} model.
Since the model must decode content after \promise{} tags without access to the corresponding \async{} content, I implement three key modifications to the standard next-token prediction finetuning algorithm.

First, the attention mask prevents the tokens after a \promise{} from attending to content within its \async{} tags until the model emits a \sync{} tag.
This ensures asynchronous blocks decode independently.
Second, the position IDs of tokens after a \promise{} depend on the length of the corresponding \async{} block, which the model does not yet know at decoding time.
To handle this, I train the model to predict the length of each \async{} block in the \tokens{} attribute of its \promise{} tag.
This prediction enables assigning position IDs to tokens that follow the \promise{} tag.
Finally, to enable the model to continue decoding past \promise{} tags, I set the next-token prediction target at each \promise{} to be the first token after its corresponding \async{} block.
This enables the model to skip past \promise{} tags.

\paragraph{\lang{} Preference Optimization.}
\label{para:pasta:pref-pairs}
\pasta{} further improves the \pastasft{} model with a second stage of training that directly optimizes for quality and speedup achieved by the \lang{} annotations.

First, \pasta{} trains a baseline sequential model \baseline{} by fine-tuning the same base model on the \pastasft{} dataset without the annotations.
Then, for each prompt in the \pastasft{} dataset, \pasta{} samples $N$ different \lang{}-annotated responses from the \pastasft{} model using a temperature $T.$
It then scores each of the $N$ sampled responses.
I set $N = 10$ and temperature $T=1.$
The score for ranking the sampled responses is a combination of:
\begin{enumerate}[leftmargin=*, topsep=0pt, itemsep=0pt]
    \item Response \emph{theoretical speedup}---the ratio between (1) the total number of tokens in \baseline{}'s response, and (2) the length of the longest sequence that \pastasft{} must decode sequentially. This measures the maximum achievable speedup of \pastasft{} over \baseline{}.
     \item Response \emph{quality}---for each sampled response, \pasta{} computes a confidence-weighted win-loss ratio by comparing it against the \baseline{}'s response and the original SlimOrca response.
     Each comparison appears in both orders. Gemini 1.5 Pro judges each pair, providing a preference and a confidence score between 0 and 1.
     The final ratio is the confidence-weighted sum of wins divided by the confidence-weighted sum of losses.
\end{enumerate}

Each sampled response receives a score:
$\text{speedup} + \lambda \times \text{quality}$, where $\lambda$ is the \emph{quality weight}.
For each prompt, \pasta{} selects the highest and lowest-scoring response as the preferred and rejected example, respectively.
I explore alternative scoring methods in \Cref{sec:pasta:preference-score}.

While the methodology is compatible with any LLM preference optimization algorithm, in this work I use BoNBoN optimization \cite{gui2024bonbonalignmentlargelanguage} as it was the state-of-the-art algorithm at the time of this work.
The BoNBoN algorithm trains a model to approximate the best-of-N response distribution by combining an SFT loss on the preferred example with an IPO preference loss \cite{azar2024general} between the preferred and rejected responses.
Specifically, the BoNBoN objective is:
\begin{align*}
\mathcal{L}_{\rm BoNBoN}(\theta; D, \theta_\text{init}) =\; &\mathbb{E}_{x, y^+, y^- \sim D} [-\alpha\log p_\theta( y^+ | x)
\\&+(1 - \alpha)(
\log{\frac{p_\theta( y^+ | x)}{p_\theta( y^- | x)}}
\\&- \log{\frac{p_{\theta_\text{init}}( y^+ | x)}{p_{\theta_\text{init}}( y^- | x)}} -
\frac{1}{\beta}
)^2]
\end{align*}
where $\theta$ and $\theta_{\text{init}}$ denote the trained and initial model weights, $x$ is a prompt, $y^+$ and $y^-$ are the preferred and rejected \lang{}-annotated responses, and $\alpha$ weights the SFT and IPO loss contributions.
I set $\theta_\text{init}$ to be the \pastasft{} model.
I use a learning rate of 5E-7 and set $\alpha$ to $0.005$ as recommended by \textcite{gui2024bonbonalignmentlargelanguage}.

\section{Experimental Setup}
\label{sec:pasta:setup}

\paragraph{Baselines.}
I compare against the following baselines:
standard autoregressive decoding from \baseline{}, APAR decoding \cite{liu2024apar}, and SoT decoding \cite{ning2023skeleton}.
APAR and SoT are examples of asynchronous decoding techniques relying on hand-crafted syntactic heuristics.
I evaluate sequential autoregressive decoding using the \baseline{} model finetuned on the \pastasft{} dataset without \lang{} annotations.
To evaluate APAR decoding, I preprocess the same \pastasft{} dataset following the official APAR methodology, applying regex and filters to extract structured data \cite{liu2024apar}.
I then train an APAR model on this dataset.

\paragraph{Models and Hyperparameters.}
For all experiments, I use Gemma 7B \cite{gemmateam2024gemmaopenmodelsbased} as the base model.
I finetune all models using a batch size of 8, a learning rate that decays linearly from 1e-5 to 0, and train for a total of 4 epochs.
I choose these hyperparameters as they maximize the quality of the baseline model.\footnote{For reference, the baseline model achieves a 38\% length controlled win rate against Gemma-7B-it, Google's officially released instruction-tuned Gemma-7b model.
This comparison demonstrates that despite the smaller-scale experimental setup, the baseline model performs competitively against the state of the art.}
I selected the learning rate and epochs based on which combination led to the highest quality \baseline{} model as measured by length-controlled win-rate on AlpacaEval.
I performed a grid search over all combinations of learning rates in $[1\text{e-}4, 1\text{e-}5, 1\text{e-}6]$ and epochs in $[1, 4, 8]$.
Training with a learning rate of $1\text{e-}5$ for $4$ epochs resulted in the highest win-rate model.

There are two \pasta{}-specific hyperparameters: $\lambda$, the quality weight used for building preference pairs (\Cref{para:pasta:pref-pairs}); and $r$, the number of BoNBoN preference optimization iterations.
I train multiple \pasta{} models, with $\lambda=1,2,4,8$ and set $r=2$.
I refer to each model as \pastabon{$r$}{$\lambda$}.
I also train \pasta{} models optimizing exclusively for quality, denoted \pastabon{$r$}{$\infty$}.

\paragraph{Hardware and Software.}
I evaluate decoding performance using PyTorch with torch.compile optimization set to maximum auto-tuning mode \cite{paszke2019pytorchimperativestylehighperformance}.
All experiments run on H100 GPUs using greedy decoding.
I use a batch size of 1 as is common with parallel decoding literature \cite{leviathan2023fast}.
To avoid measuring compilation overhead, I decode each request twice and only take the timing of the second run.

\begin{figure}[t]
  \centering
  \begin{subfigure}[t]{0.32\linewidth}
    \centering
    \includegraphics[width=\linewidth]{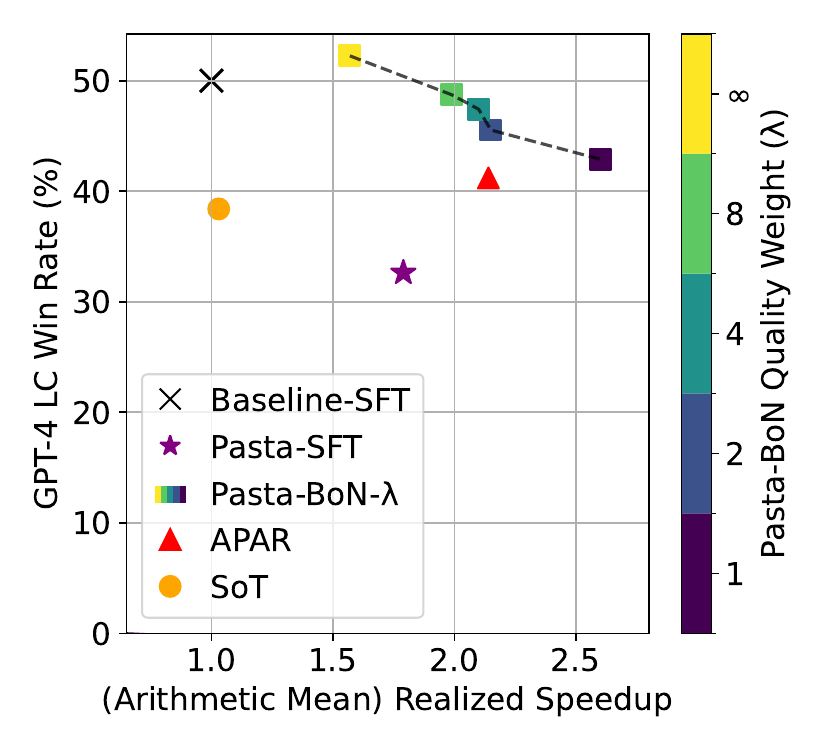}
    \subcaption{Realized speedup. \pasta{} models achieve a Pareto-optimal quality--speedup trade-off.}
    \label{fig:pasta:quality-parallelism-rs-arith}
  \end{subfigure}
  \hfill
  \begin{subfigure}[t]{0.32\linewidth}
    \centering
    \includegraphics[width=\linewidth]{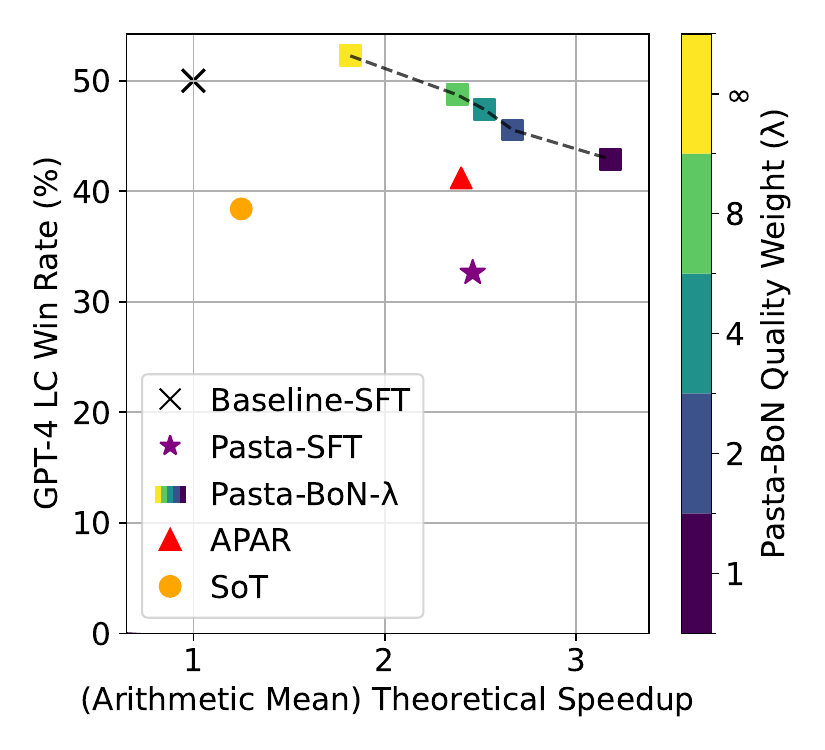}
    \subcaption{Theoretical speedup. Realized speedup with the \lang{} interpreter is close to the theoretical limit.}
    \label{fig:pasta:theoretical-speedup-arith}
  \end{subfigure}
  \hfill
  \begin{subfigure}[t]{0.32\linewidth}
    \centering
    \includegraphics[width=\linewidth]{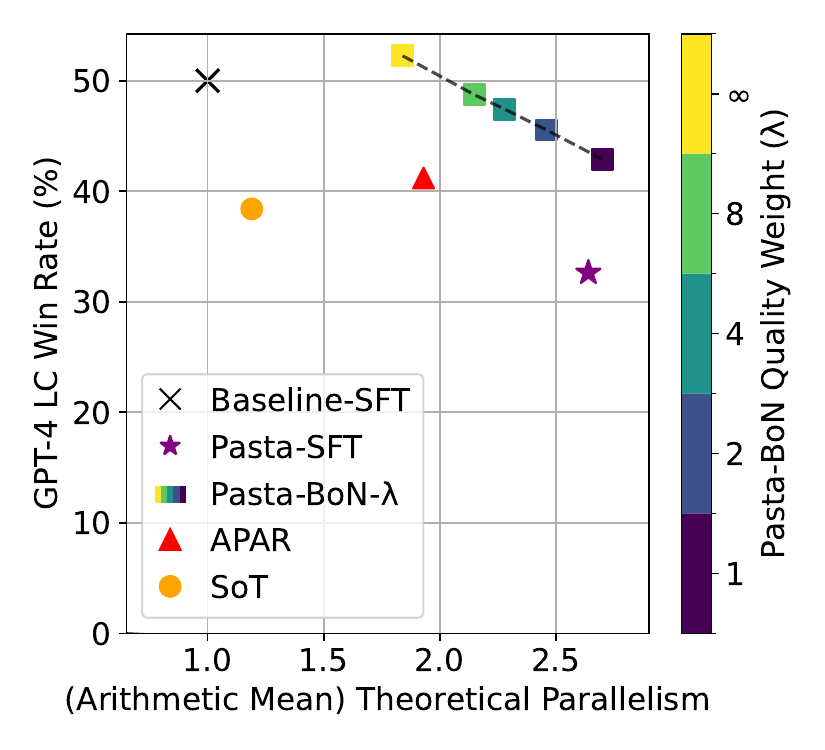}
    \subcaption{Theoretical parallelism. \pasta{} outputs exhibit a high degree of parallelism.}
    \label{fig:pasta:parallelism-arith}
  \end{subfigure}
  \caption{PASTA evaluation metrics computed using arithmetic mean (for comparison with prevailing literature practice).}
  \label{fig:pasta:arithmean-plots}
\end{figure}

\paragraph{Evaluation.}
I evaluate all models on the AlpacaEval benchmark \cite{dubois2024length,alpaca_eval}, an open-ended suite of 805 representative prompts.
For each decoding method, I compute the following metrics:
\begin{enumerate}[leftmargin=*, topsep=0pt, itemsep=0pt]
    \item I measure the \textit{realized speedup} against the baseline as the ratio between the wall-clock decoding times: baseline model time divided by test model time, both using the \lang{} runtime.
    \item I calculate the \textit{theoretical speedup} against the baseline as the ratio between (1) the total number of tokens in the baseline response and (2) the length of the longest sequence that the test model must decode sequentially, as described in \Cref{sec:pasta:training}.
    \item I measure the \textit{theoretical parallelism} in the model output: the ratio of (1) the total number of non-control tokens to (2) the length of the longest sequence of tokens that the model must decode sequentially.
    \item I measure the \textit{quality} as the length-controlled, LLM-as-a-judge win-rate using AlpacaEval benchmark \cite{alpaca_eval,dubois2024length} when compared to the baseline model.
    I use Gemini 1.5 Pro \cite{geminiteam2024geminifamilyhighlycapable} as the judge model for development and GPT4 as the judge for evaluation, to prevent reward hacking.
\end{enumerate}
I aggregate the speedup and parallelism over each prompt in the dataset using the geometric mean.\footnote{The arithmetic mean of ratios can lead to inconsistent results depending on the baseline used to compute the ratio and is therefore not appropriate for this use \cite{fleming1986not}. However, the prevailing practice in parallel decoding literature uses arithmetic mean when reporting speedup. I provide the speedup computed with arithmetic mean for reference in \Cref{fig:pasta:arithmean-plots}.}

\label{sec:pasta:results}
\begin{figure}[htbp]
    \centering
    \begin{minipage}[b]{0.32\linewidth}
        \centering
        \includegraphics[width=\linewidth, trim=15px 0px 15px 0, clip]{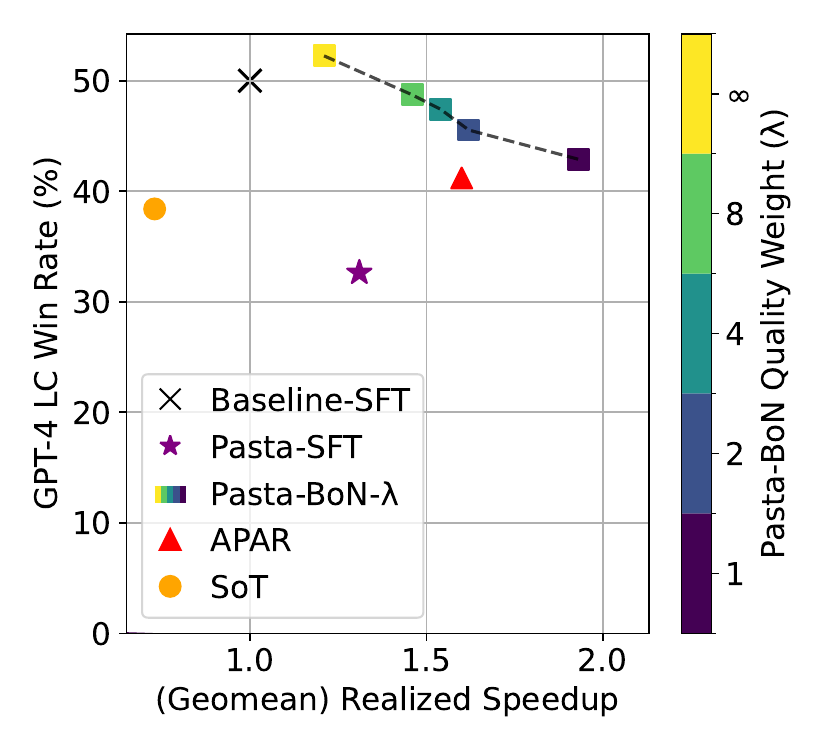}
        \label{fig:pasta:quality-parallelism-rs}
    \end{minipage}
    \hfill
    \begin{minipage}[b]{0.32\linewidth}
        \centering
        \includegraphics[width=\linewidth, trim=15px 0px 15px 0, clip]{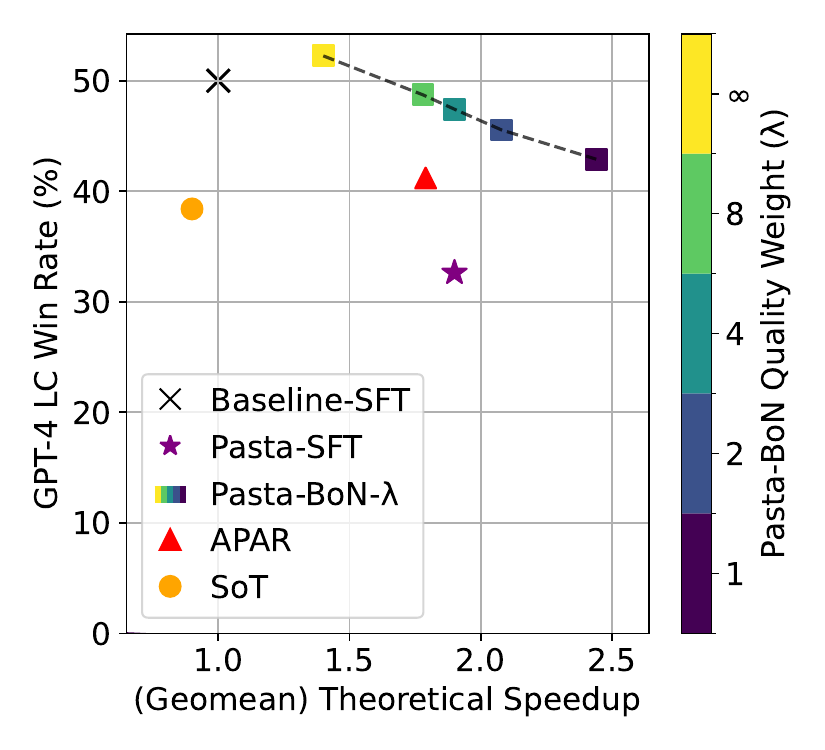}
        \label{fig:pasta:theoretical-speedup}
    \end{minipage}
    \hfill
    \begin{minipage}[b]{0.32\linewidth}
        \centering
        \includegraphics[width=\linewidth, trim=15px 0px 15px 0, clip]{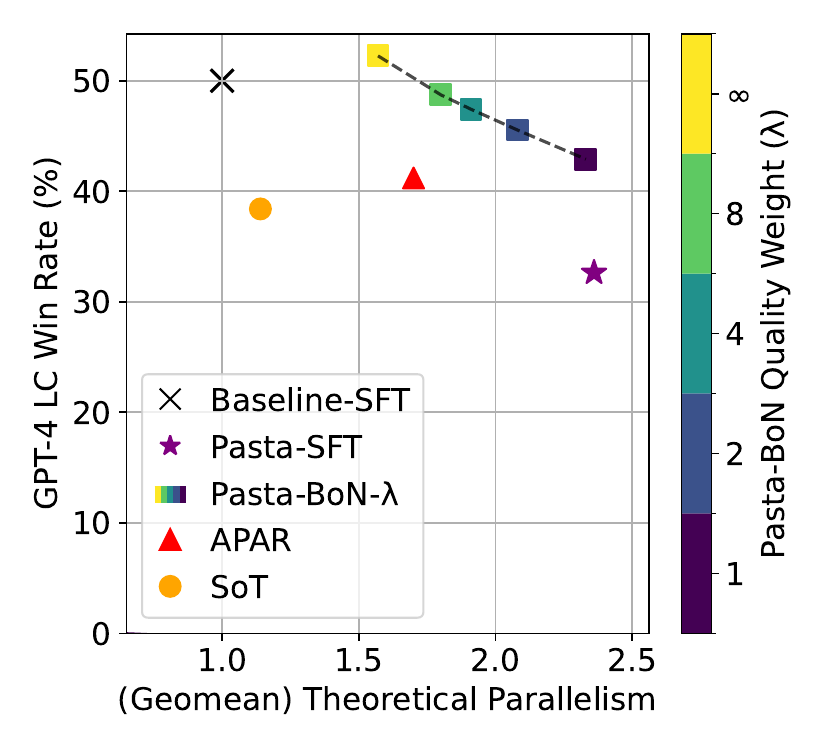}
        \label{fig:pasta:parallelism-plot}
    \end{minipage}
    \caption{Left (Realized Speedup). \pasta{} models achieve Pareto-optimal quality-latency trade-off compared to asynchronous decoding strategies with hand-crafted heuristics.
    Middle (Theoretical Speedup). The realized speedup using \lang{} runtime is close to the theoretical speedup.
    Right (Theoretical Parallelism). \pasta{} responses exhibit high parallelism.}
    \label{fig:pasta:comparison-plots}
\end{figure}

The left, middle and right plots in \Cref{fig:pasta:comparison-plots} show how response quality trades off against realized speedup, theoretical speedup, and parallelism respectively across \pasta{} and baseline models.
I mark the baseline (50\% win rate with no speedups) with an X.
An ideal asynchronous decoding strategy should match this baseline in quality while surpassing it in speedup;
the closer to the top right corner of the plot, the better the technique.

\paragraph{Pareto-Optimality.}
The left plot in \Cref{fig:pasta:comparison-plots} shows that \pasta{} models achieve Pareto-optimal trade-off between quality and realized speedup.
The best-quality model (\pastabon{2}{$\infty$}) achieved 52.3\% win rate at 1.21$\times$ speedup, whereas the best-speedup model (\pastabon{2}{1}) achieved 42.9\% win rate at 1.93$\times$ speedup.
The \pastabon{2}{2} model Pareto-dominates all prior asynchronous decoding techniques, achieving superior quality-latency trade-off.

The APAR model and \pastabon{2}{2} achieved similar speedups (1.6$\times$ vs 1.62$\times$), but \pastabon{2}{2} obtained a 4.3\% higher win rate.
This result stems from APAR's reliance on syntactic heuristics, which cannot reliably identify parallelization opportunities and can lead to false positives.
Similarly, while APAR and \pastabon{2}{1} showed comparable win rates (41.2\% vs 42.9\%), \pastabon{2}{1} delivered a 20.6\% higher speedup.
This superior speedup stems from the flexibility of \lang{}'s annotation, which enables asynchronous decoding at any position in the output.

Notably, I do not observe any speedup by SoT \cite{ning2023skeleton}, when applied to \baseline{}.
SoT, as a prompt-based method, requires the base model to have strong instruction-following ability to perform well.
I validate this by applying SoT to the stronger official instruction-finetuned Gemma-IT model from Google.
With this generous implementation, SoT achieved a 1.61$\times$ speedup while dropping its win rate by 12\%.
In contrast, \pastabon{2}{2} achieved 1.62$\times$ speedup with only a 5\% drop to win rate.

\paragraph{Role of $\lambda.$}
The quality weight $\lambda$ serves as an effective control knob for the trade-off between quality and speedup. A lower weight results in more aggressive optimization for speedup at the cost of reduced quality.

\paragraph{Theoretical speedup.}
Comparing the left (realized speedup) and middle (theoretical speedup) plots in \Cref{fig:pasta:comparison-plots} shows that the combination of \lang{} runtime and \pasta{} models delivers realized speedup close to theoretical optimal.

\paragraph{Parallelism.}
The right plot in \Cref{fig:pasta:comparison-plots} shows that \pasta{} models achieve a high degree of theoretical parallelism.
Notably, while the \pastasft{} model starts with high theoretical parallelism, this does not translate into significant decoding speedup.
This disconnect occurs because the model generates redundant parallel branches, inflating theoretical parallelism without improving decoding speed.
Preference optimization mitigates this pathology, as \emph{Pasta-BoN} models deliver speedups commensurate with their theoretical parallelism.

\paragraph{Conclusion.}
These results validate the self-orchestrating approach for the parallel generation bottleneck: \pasta{} produces Pareto-dominant performance by learning to annotate semantic dependence rather than relying on hand-crafted heuristics.
\pasta{} also enables flexible trade-off between speedup improvements and response quality.

\paragraph{Arithmetic Mean Evaluation.}
The prevailing practice in parallel decoding literature uses arithmetic mean to compute average speedup.
However, practitioners should use geometric mean when averaging normalized values such as speedup against a baseline \cite{fleming1986not}.
As such, I report geometric mean above but include here in \Cref{fig:pasta:quality-parallelism-rs-arith,fig:pasta:theoretical-speedup-arith,fig:pasta:parallelism-arith} the results computed using arithmetic mean as reference.
Notably, by definition, arithmetic mean is larger than or equal to geometric mean, therefore \Cref{fig:pasta:quality-parallelism-rs-arith,fig:pasta:theoretical-speedup-arith,fig:pasta:parallelism-arith} show notably higher speedup and parallelism than the geometric mean results.
Furthermore, \pasta{} models still Pareto-dominate using arithmetic mean.

\section{Sensitivity Analysis}
\label{sec:pasta:sensitivity}

In this section, I investigate how three key design choices of the \pasta{} system impact model quality and speedup:
a) the number of preference optimization iterations,
b) the position ID assignment strategy, and
c) the scoring method for \lang{}-annotated responses.

\begin{figure}[htbp]
    \centering
    \begin{minipage}[b]{0.32\linewidth}
        \centering
        \includegraphics[width=\linewidth, trim=15px 0px 15px 15px, clip]{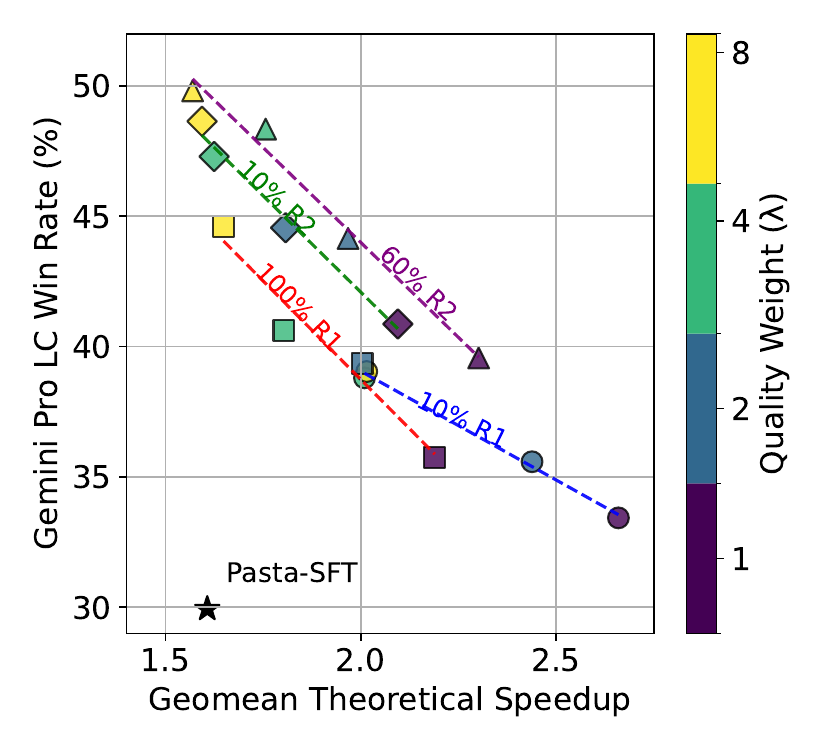}
        \label{fig:pasta:scalability}
    \end{minipage}
    \hfill
    \begin{minipage}[b]{0.32\linewidth}
        \centering
        \includegraphics[width=\linewidth, trim=0px 0px 0px 15px, clip]{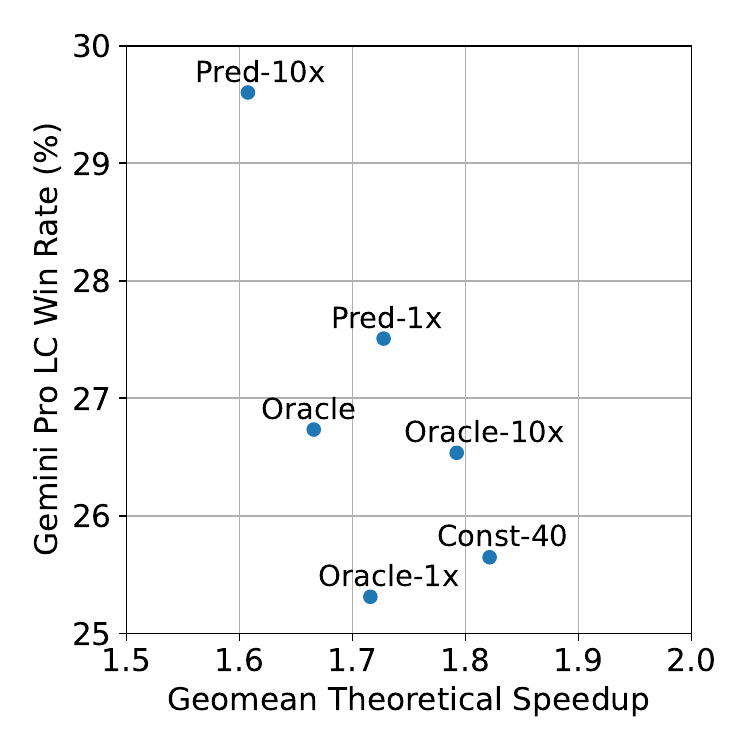}
        \label{fig:pasta:positional-encoding}
    \end{minipage}
    \hfill
    \begin{minipage}[b]{0.32\linewidth}
        \centering
        \includegraphics[width=\linewidth, trim=0px 0px 0px 15px, clip]{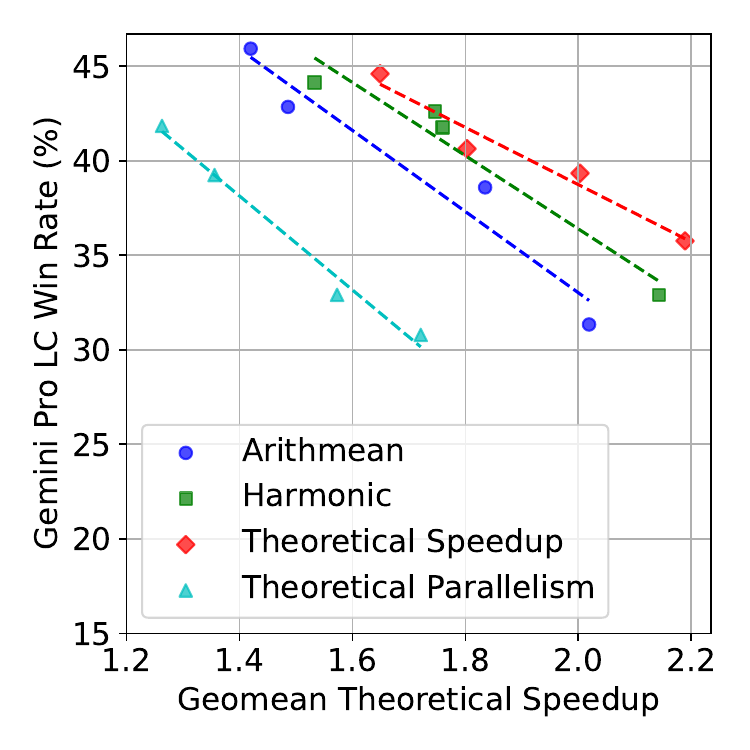}
        \label{fig:pasta:optimization-objective}
    \end{minipage}
    \caption{Left (Scalability). Increasing the number of rounds of preference optimization continuously improves the quality-latency trade-off.
    Middle (Position ID Assignment). Analysis of different methods for computing position IDs during decoding. LLM-based prediction of the position IDs in multiples of ten (Pred-10x) achieves the highest quality without significantly sacrificing speedup.
    Right (Preference Score). Analysis of different metrics for decoding efficiency used in calculating the \lang{} preference scores. Optimizing for the theoretical speedup achieves both high theoretical speedup and length-controlled win rate.}
    \label{fig:pasta:sensitivity-analysis-plots}
\end{figure}

\subsection{Number of Iterations}
\label{sec:pasta:num-iterations}

I examine how the number of preference optimization iterations affects the quality-latency trade-off in \pasta{} models;
I observe continuous improvements as training compute increases and find distinct optimization dynamics at different preference optimization iterations.

\paragraph{Methodology.}
I analyze the impact of preference optimization on the quality-latency trade-off of \pasta{} models at 5 different stages:
(1) initial (aka \pastasft{}), (2) 10\% into Round 1, (3) 100\% into Round 1, (4) 10\% into Round 2, and (5) 60\% into Round 2.\footnote{I stopped at 60\% of Round 2 due to time constraints.}
A single star represents the initial stage, while results from later stages contain four points each, corresponding to quality weights ($\lambda$) of 1, 2, 4, and 8.
I visualize each stage's data using distinct colors in \Cref{fig:pasta:scalability}.
To illustrate the difference in quality-latency trade-off between stages, I compute linear fits for each stage after \pastasft{} and plot the linear fit using the same color as each stage.

\paragraph{Results.}
\Cref{fig:pasta:scalability} demonstrates the scalability of preference optimization, as increased training compute continuously improves the Pareto frontier toward better quality-latency trade-offs (i.e., top right corner).
As is common when scaling LLMs with more training compute \cite{kaplan2020scalinglawsneurallanguage}, I observe diminishing returns, though I do not observe saturation after two rounds of preference optimization.

The linear fits of quality-latency trade-off within each stage reveal distinct optimization dynamics.
The initial 10\% of Round 1 preference optimization aggressively optimizes for speedup, moving the group of points for that stage to the right of the plot.
However, from 10\% Round 2 to 60\% Round 2, I observe an emphasis on quality improvements, shifting the group of points upward.
Overall, preference optimization effectively navigates the trade-off space by exploring both quality improvement and speedup improvement.

\paragraph{Conclusion.}
Preference optimization is a scalable technique that improves the quality-latency trade-off with increased training compute.
This scalability is a direct benefit of the learned annotation approach: unlike hand-crafted syntactic heuristics, which require manual engineering to improve, preference optimization directly converts computational resources into a better quality-latency trade-off.

\subsection{Position ID Assignment}
\label{sec:pasta:positional-embedding}

Asynchronous decoding methods introduce uncertainty over the true position of tokens during generation because the runtime generates the output in a non-sequential manner.
Namely, the main decoding thread is unaware of the true length of any previously occurring \async{} blocks that the runtime has not yet synchronized.
Thus, the main thread must estimate the number of tokens in an \async{} block before continuing generation.
If the predicted number of tokens for the \async{} block does not match the true number of tokens generated, this can lead to errors as the position IDs after synchronization will either not increase monotonically (predicted too few) or contain a gap (predicted too many).

\paragraph{Methodology.}
To minimize the error between the true and predicted position IDs, I compare three different approaches for position ID assignment:

\textit{Fixed-length}: I assume that each \async{} block has a fixed length.
In the experiments, I set this length to forty tokens as this is slightly larger than the median \async{} block length in the training data.

\textit{Length Prediction}: I train the model to predict the length of each \async{} block, and then use the model's predictions during decoding.
I evaluate two variants that predict \async{} block lengths at different granularities:
1) Pred-1X: Predict the \async{} token length exactly; and
2) Pred-10X: Predict the \async{} token length as a multiple of ten for coarser granularity.

\textit{Oracle}:
I use the ground truth length of each \async{} block (i.e.\ the position IDs of the tokens in the block if the runtime decoded the chunk sequentially) to assign position IDs.
While the oracle position IDs are infeasible to obtain during deployment, I evaluate the performance of oracle ID decoding to serve as a reference point for the performance of decoding with no error in the position ID calculations.
I evaluate two different granularities for the oracle position IDs:
1) Oracle-1X: Using the true \async{} block length; and
2) Oracle-10X: Using the true \async{} block length rounded to the nearest multiple of ten.
Both Oracle-1X and Oracle-10X include the predicted length in the \promise{} tag.
I also evaluate a third variant, Oracle, which uses the exact \async{} block lengths to offset position IDs but does not include the length in the \promise{} tag, isolating the effect of length information on quality.

\paragraph{Results.}
\Cref{fig:pasta:positional-encoding} presents both the response quality and speedup for models finetuned on the \pastasft{} dataset using each of the different position ID estimation techniques.
Length prediction performs the best, achieving quality and speedup metrics matching (or even slightly exceeding) the Oracle.
Unlike fixed-length or oracle baselines, length prediction is admissible to preference optimization because the model generates the length estimate as part of its output.
Among the two prediction variants, I adopt Pred-10X as the position ID assignment strategy because its coarser granularity makes the prediction task easier.

\subsection{\lang{} Preference Score}
\label{sec:pasta:preference-score}

When performing \lang{} preference optimization, I compute a preference score for each \lang{}-annotated response as a weighted combination of a quality term and a decoding efficiency term, which reflects the improvement in decoding speed for a response.
While the ultimate goal is to increase decoding speed, it is not obvious \emph{a priori} that directly optimizing for the speedup will produce the desired behavior. Such an objective could lead to degenerate solutions such as producing short responses.

\paragraph{Methodology.}
To determine the most effective decoding efficiency term, I investigate four separate metrics for the decoding efficiency of a response:
(a) the theoretical speedup of the response only;
(b) the harmonic mean of the theoretical speedup and theoretical parallelism of the response;
(c) the arithmetic mean of the theoretical speedup and theoretical parallelism of the response;
(d) the theoretical parallelism of the response only.

I recompute the preference score for each training response using each of the efficiency metrics presented above.
I then perform a single round of BoNBoN training on each of the different preference-labeled datasets.

\paragraph{Results.}
\Cref{fig:pasta:optimization-objective} presents the response quality and speedup for each model trained to optimize a different decoding efficiency metric.
Intuitively, one might expect the harmonic mean to be the most effective since it encourages balanced optimization by inducing a larger weight on the weaker metric.
However, I find that optimizing for harmonic mean performs similarly to optimizing directly for speedup. I hypothesize this is because the LLM-based quality evaluation naturally favors longer, more detailed responses, preventing the model from artificially increasing speedup through response truncation. As expected, optimizing solely for theoretical parallelism leads to poor speedup, demonstrating the importance of including speedup in the objective.
Based on these results, I adopt theoretical speedup only as the efficiency metric.

\section{Related Work}
\label{sec:pasta:related-work}

After discussing parallel decoding in \Cref{sec:pasta:taxonomy}, I now survey other relevant research areas.

\paragraph{Agent Planning/Tool Use.}
LLM-based agents solve complex tasks by planning and invoking tools such as web search and external APIs \cite{yao2023react,schick2023toolformer,shen2023hugginggpt,liang2023taskmatrixai,lu2023chameleon}.
\pasta{} applies this paradigm to inference itself: the model uses \lang{} annotations to invoke the \pasta{} runtime as a tool, directing how its own output generates in parallel.

\paragraph{Approximate Parallelization.}
This work extends the idea of approximate parallelization \cite{10.1145/1993498.1993555,10.1145/2414729.2414738}.
\textcite{10.1145/1993498.1993555} proposed a framework that enables programmers to annotate breakable data dependences in a program and
developed a compiler and runtime that exploits these annotations to automatically parallelize otherwise sequential regions of code.
\textcite{10.1145/2414729.2414738} opportunistically relaxes synchronization primitives in a parallel program to improve parallelism, producing program outputs that are acceptably close to the original.
Similarly, PASTA breaks the sequential decoding process of LLMs into approximately parallelizable components and exploits the parallelism to improve decoding efficiency.
PASTA differs in that instead of relying on end-user annotation or compiler analysis, it teaches LLMs to annotate semantic dependence in their own output, enabling the co-designed runtime to exploit the resulting parallelism.

\section{Conclusion}
\label{sec:pasta:summary}

This chapter introduces \pasta{}, a system that teaches LLMs to annotate semantic dependence between chunks in their own responses, enabling chunks without dependences to generate in parallel.
\pasta{} validates the self-orchestrating paradigm for the parallel generation bottleneck: evaluation on AlpacaEval demonstrates that the approach Pareto-dominates existing asynchronous decoding methods in terms of quality and speedup, and the improvements continue with additional training compute, showing no signs of saturation.

The key insight that LLMs can annotate semantic dependence to direct their own inference transfers beyond parallel generation.
In the next chapter, I apply this same principle to a different inference bottleneck: memory management through learned context eviction in Thinking in Place.

\chapter{Thinking in Place}
\label{ch:tip}

The previous chapter trains models to exploit semantic dependence for parallel generation: PASTA parallelizes autoregressive decoding by annotating which response chunks can generate independently. Parallel generation accelerates decoding but leaves the KV cache footprint unchanged---every generated token remains in memory for the duration of decoding, long after many reasoning steps become \emph{dead}; that is, future reasoning steps no longer need them. This chapter turns semantic dependence toward a different bottleneck: memory. I present Thinking in Place (TIP), a method where the model learns to annotate dead reasoning steps and the runtime evicts their KV cache entries.

\section{Introduction}
\label{sec:tip:intro}

Reasoning models solve hard problems by generating long chains of thought \parencite{wei2022cot,nye2021scratchpads}, routinely producing tens of thousands of tokens per request \parencite{zhang2025deepseekr1}. As \Cref{ch:prelim} describes, each generated token appends a pair of key--value vectors to a running cache, so the KV cache grows linearly with the number of generated tokens. Long reasoning traces therefore strain accelerator memory: KV cache storage and bandwidth often limit batch size and throughput in serving systems \parencite{kwon2023efficient,zhang2023h2o}.

Recent work mitigates this cost by evicting tokens that appear unimportant under heuristic rules or scores \parencite{zhang2023h2o,wang2025sagekv,dong2026foresightkv}. During chain-of-thought reasoning, however, the model rarely needs the full text of every earlier step for the remainder of the computation. Once the model uses an intermediate result, many earlier steps become semantically irrelevant to future decoding, but standard decoding keeps them in the KV cache and keeps them visible to attention. The model therefore stores the entire trace in memory even though only a small fraction of it remains semantically relevant.

\begin{figure}[t]
\centering
\includegraphics[width=0.9\linewidth]{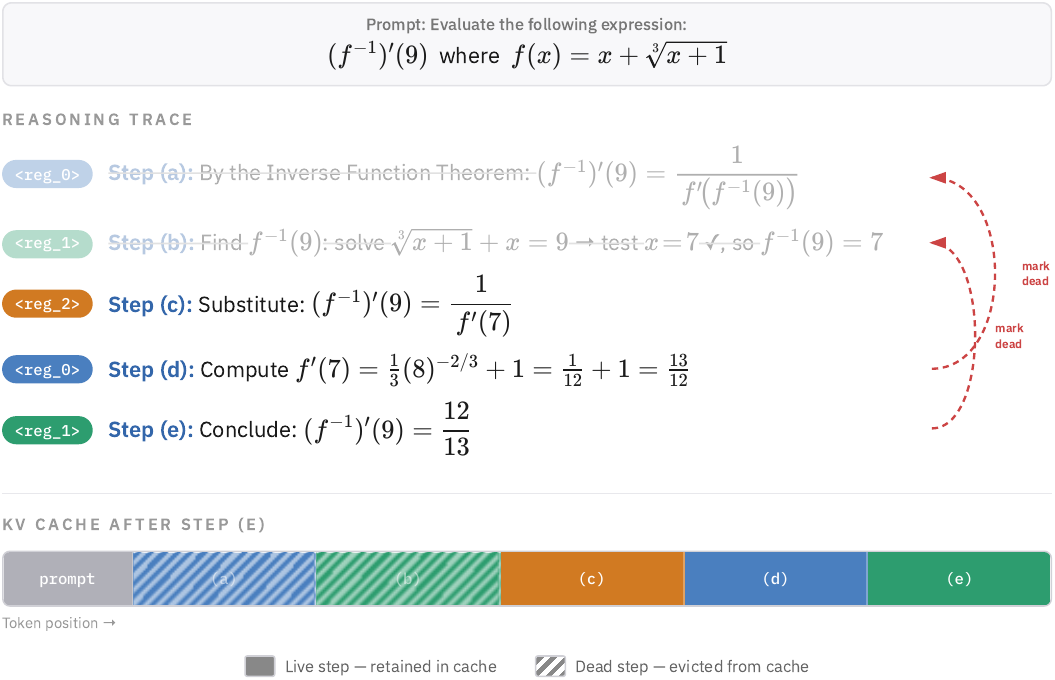}
\caption{Learned context eviction on a five-step calculus problem with $N=3$ thought registers.
The model prefixes each reasoning step with a thought register token (\texttt{<reg\_0>}, \texttt{<reg\_1>}, or \texttt{<reg\_2>}), assigning the step to that thought register (top).
At Step~(d), the model reuses \texttt{<reg\_0>}, which makes Step~(a) a dead step; at Step~(e), it reuses \texttt{<reg\_1>}, marking Step~(b) as dead.
The KV cache pool state (bottom, crossed-out segments) shows the resulting evictions, leaving only three live steps in context.
With $N$ thought registers the model keeps at most $N$ reasoning steps live, regardless of how many total steps it produces.}
\label{fig:tip:registers}
\end{figure}

\paragraph{Learned Context Eviction.}
I present Thinking in Place (TIP), a system for learned context eviction that consists of three components: a set of special tokens that let the model express which reasoning steps to evict, a runtime that removes the corresponding KV cache entries, and a training procedure that optimizes these eviction decisions against final-answer accuracy via reinforcement learning.
Where existing methods score token importance with heuristic surrogates, TIP optimizes task performance directly---the model learns to evict reasoning steps only when doing so does not hurt the final answer.
\Cref{fig:tip:registers} illustrates how these components interact.

\paragraph{Thought Registers.}
TIP augments the model's vocabulary with $N$ special tokens \texttt{<reg\_0>}, \ldots, \texttt{<reg\_\{N-1\}>}, called \textit{thought registers}, and the model prefixes each reasoning step with one of these tokens to assign the step to that register.
When the model reuses a thought register, it marks the reasoning step previously assigned to that register as dead, freeing its KV cache entries for eviction.
Because the model draws from a pool of $N$ registers, at most $N$ reasoning steps remain live at any time---bounding context growth by construction.
\Cref{fig:tip:registers} illustrates the mechanism.

\paragraph{Runtime.}
\looseness=-1
I build a runtime on top of paged attention~\parencite{kwon2023efficient}, which organizes KV cache memory into fixed-size blocks, making eviction a matter of dropping blocks.
When the model reuses a thought register token, the runtime evicts the dead step's KV cache blocks: it removes them from the attention kernel, so subsequent tokens never attend over the discarded reasoning.

\paragraph{Training.}
Training a TIP model proceeds in two stages.
In the first stage, I collect reasoning traces annotated with thought register assignments and fine-tune the model to emit thought register tokens at the boundaries of reasoning steps.
This stage teaches the model the syntax of thought registers---where to place tokens and how to signal that a reasoning step is dead.
In the second stage, I apply reinforcement learning with a reward that combines final-answer correctness with a KV-cache penalty for retaining too many tokens.
This stage optimizes the model's reuse decisions directly against task performance: the model learns to mark reasoning steps as dead early, enabling the runtime to evict their KV cache pages.

\paragraph{Results.}
On AIME 2024 and AIME 2025, TIP Pareto-dominates both the Dense baseline and heuristic-eviction baselines (StreamingLLM, Paged H2O) across accuracy, live KV cache size, throughput, and decode time.
Against the Dense baseline on AIME 2024, TIP raises accuracy from 14.4\% to 20.8\% while roughly halving the live KV cache, shortens total decode time by 28--35\%, and increases throughput by 21--30\%.

\paragraph{Contributions.}
I make the following contributions:
\begin{itemize}[leftmargin=*, topsep=0pt, itemsep=0pt]
    \item I introduce learned context eviction, a mechanism in which the model itself marks reasoning steps it no longer needs so the system can discard those steps from the KV cache during generation.
    \item I implement a paged-attention-based runtime that evicts dead steps' KV cache blocks so subsequent tokens never attend over discarded reasoning steps. This implementation modifies only which blocks the attention kernel reads and requires no modification to the paged attention kernel.
    \item I train TIP in two stages. I first annotate reasoning traces with special tokens denoting eviction decisions, and fine-tune the model on these traces to teach it the annotation syntax. I then apply reinforcement learning to optimize the model's eviction decisions against a reward that combines task accuracy with a KV cache penalty.
    \item I evaluate TIP on AIME 2024 and AIME 2025 and find that it Pareto-dominates both standard autoregressive decoding and heuristic-eviction baselines on accuracy, live KV cache size, throughput, and decode time.
\end{itemize}

\section{Method}
\label{sec:tip:method}

Training a model to evict its own reasoning steps requires four ingredients: special tokens that let the model express eviction decisions, a runtime that implements these evictions during decoding, supervised data that marks when each reasoning step becomes dead, and a reinforcement learning stage that optimizes these decisions against final-answer accuracy.

\subsection{Thought-Register Mechanism}
\label{sec:tip:ids}

I define a \emph{reasoning step} as a unit of thought in a chain-of-thought output: one idea, one calculation, or one deduction.
TIP augments the model's vocabulary with $N$ special \emph{thought register tokens}: \texttt{<reg\_0>}, \texttt{<reg\_1>}, \ldots, \texttt{<reg\_\{N-1\}>}.
Each token names a \emph{thought register}---a logical slot that holds one reasoning step at a time.
The model prefixes each reasoning step with a thought register token.
When two steps share the same thought register, the earlier step becomes dead and only the later one remains.
Because the model draws from $N$ thought registers, at most $N$ reasoning steps are live at any time.

\begin{figure}[p]
\centering
\includegraphics[width=0.9\linewidth]{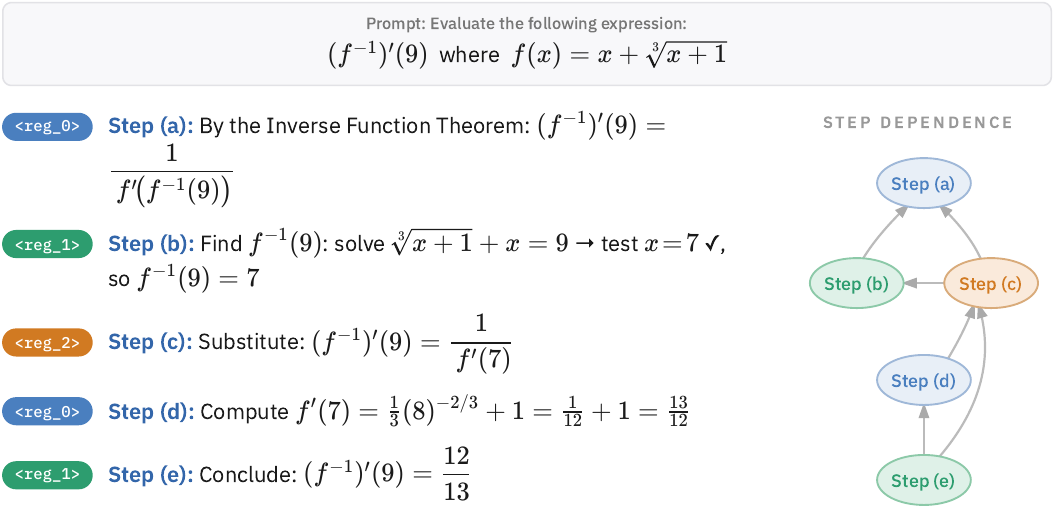}
\caption{Thought register mechanism on the example from \Cref{fig:tip:registers} ($N=3$). The model prefixes each reasoning step with a thought register token. When it reuses a thought register, the previous occupant becomes dead and the runtime evicts the dead step's KV cache pages. The liveness graph (right) shows which steps share a thought register: Step~(d) reuses \texttt{<reg\_0>} and marks Step~(a) dead; Step~(e) reuses \texttt{<reg\_1>} and marks Step~(b) dead. At most $N$ reasoning steps remain live at any time.}
\label{fig:tip:registers-method}
\end{figure}

\paragraph{Generation.}
\Cref{fig:tip:registers} walks through a five-step solution with $N=3$ thought registers.
The model emits a thought register token before each reasoning step, assigning the step to that thought register.
The model marks a reasoning step as dead by reusing the corresponding thought register token; the runtime observes this reuse and begins accumulating the new reasoning step in that thought register.
Each reasoning step runs from its thought register token up to the next thought register token or end-of-sequence token; steps are variable-length.

\paragraph{Liveness Invariant.}
Two invariants hold throughout decoding.
First, prompt tokens are always live.
Second, reasoning steps cycle through thought registers: each thought register holds at most one live step, so the number of live reasoning steps never exceeds $N$.

\subsection{Runtime}
\label{sec:tip:runtime}

Production LLM inference systems manage the KV cache with \emph{paged attention}~\parencite{kwon2023efficient}, which groups KV cache entries into fixed-size blocks and allocates or frees each block independently.
The TIP runtime builds on this mechanism: it interprets thought register tokens to decide which pages to evict, controlling which pages the attention kernel reads at each step.

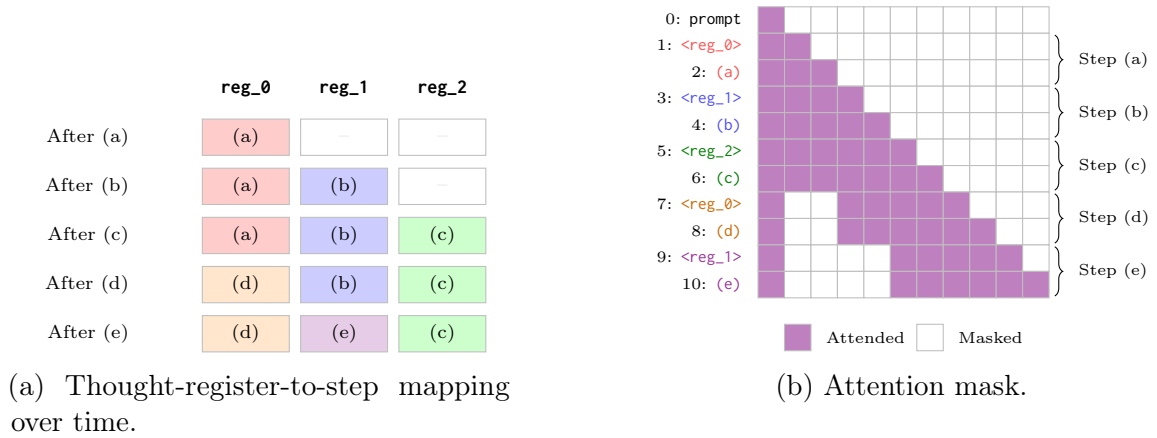
\begin{figure}[p]
\centering
\begin{subfigure}[t]{0.40\linewidth}
\centering
\begin{tikzpicture}[
    font={\fontsize{7.5}{8.5}\selectfont},
    cell/.style={rectangle, draw=gray!50, minimum width=1.15cm, minimum height=0.48cm, align=center, inner sep=1pt},
    head/.style={minimum width=1.15cm, minimum height=0.48cm, align=center, font={\fontsize{7.5}{8.5}\selectfont\bfseries}},
    rlabel/.style={anchor=east, font={\fontsize{7.5}{8.5}\selectfont}},
    stepa/.style={cell, fill=red!20},
    stepb/.style={cell, fill=blue!20},
    stepc/.style={cell, fill=green!20},
    stepd/.style={cell, fill=orange!20},
    stepe/.style={cell, fill=violet!20},
    empty/.style={cell, fill=white, text=gray!30},
]
\def\cw{1.3}
\def\rh{0.65}

\node[head] at (\cw, 0) {\texttt{reg\_0}};
\node[head] at (2*\cw, 0) {\texttt{reg\_1}};
\node[head] at (3*\cw, 0) {\texttt{reg\_2}};

\node[rlabel] at (-0.1, -\rh) {After (a)};
\node[stepa] at (\cw, -\rh) {(a)};
\node[empty] at (2*\cw, -\rh) {--};
\node[empty] at (3*\cw, -\rh) {--};

\node[rlabel] at (-0.1, -2*\rh) {After (b)};
\node[stepa] at (\cw, -2*\rh) {(a)};
\node[stepb] at (2*\cw, -2*\rh) {(b)};
\node[empty] at (3*\cw, -2*\rh) {--};

\node[rlabel] at (-0.1, -3*\rh) {After (c)};
\node[stepa] at (\cw, -3*\rh) {(a)};
\node[stepb] at (2*\cw, -3*\rh) {(b)};
\node[stepc] at (3*\cw, -3*\rh) {(c)};

\node[rlabel] at (-0.1, -4*\rh) {After (d)};
\node[stepd] at (\cw, -4*\rh) {(d)};
\node[stepb] at (2*\cw, -4*\rh) {(b)};
\node[stepc] at (3*\cw, -4*\rh) {(c)};

\node[rlabel] at (-0.1, -5*\rh) {After (e)};
\node[stepd] at (\cw, -5*\rh) {(d)};
\node[stepe] at (2*\cw, -5*\rh) {(e)};
\node[stepc] at (3*\cw, -5*\rh) {(c)};

\end{tikzpicture}
\caption{Thought-register-to-step mapping over time.}
\label{fig:tip:kv-state}
\end{subfigure}%
\hfill
\begin{subfigure}[t]{0.57\linewidth}
\centering
\begin{tikzpicture}[
    cell/.style={rectangle, draw=lightgray, minimum size=0.35cm},
    attended/.style={cell, fill=violet!50},
    masked/.style={cell, fill=white},
    font={\fontsize{8.6}{9}\selectfont}
]
\def\cs{0.35}

\def\matrixdata{
    1, 0, 0, 0, 0, 0, 0, 0, 0, 0, 0,
    1, 1, 0, 0, 0, 0, 0, 0, 0, 0, 0,
    1, 1, 1, 0, 0, 0, 0, 0, 0, 0, 0,
    1, 1, 1, 1, 0, 0, 0, 0, 0, 0, 0,
    1, 1, 1, 1, 1, 0, 0, 0, 0, 0, 0,
    1, 1, 1, 1, 1, 1, 0, 0, 0, 0, 0,
    1, 1, 1, 1, 1, 1, 1, 0, 0, 0, 0,
    1, 0, 0, 1, 1, 1, 1, 1, 0, 0, 0,
    1, 0, 0, 1, 1, 1, 1, 1, 1, 0, 0,
    1, 0, 0, 0, 0, 1, 1, 1, 1, 1, 0,
    1, 0, 0, 0, 0, 1, 1, 1, 1, 1, 1
}

\begin{scope}
    \foreach \cell [count=\n from 0] in \matrixdata {
        \pgfmathtruncatemacro{\y}{floor(\n / 11)}
        \pgfmathtruncatemacro{\x}{mod(\n, 11)}
        \ifnum\cell=1
            \node[attended] at (\x*\cs, -\y*\cs) {};
        \else
            \node[masked] at (\x*\cs, -\y*\cs) {};
        \fi
    }
\end{scope}

\node[anchor=east] at (-0.25, 0) {\fontsize{6.5}{7}\selectfont 0: \texttt{prompt}};
\node[anchor=east] at (-0.25, -1*\cs) {\fontsize{6.5}{7}\selectfont 1: \textcolor{red!70}{\texttt{<reg\_0>}}};
\node[anchor=east] at (-0.25, -2*\cs) {\fontsize{6.5}{7}\selectfont 2: \textcolor{red!70}{\texttt{(a)}}};
\node[anchor=east] at (-0.25, -3*\cs) {\fontsize{6.5}{7}\selectfont 3: \textcolor{blue!70}{\texttt{<reg\_1>}}};
\node[anchor=east] at (-0.25, -4*\cs) {\fontsize{6.5}{7}\selectfont 4: \textcolor{blue!70}{\texttt{(b)}}};
\node[anchor=east] at (-0.25, -5*\cs) {\fontsize{6.5}{7}\selectfont 5: \textcolor{green!50!black}{\texttt{<reg\_2>}}};
\node[anchor=east] at (-0.25, -6*\cs) {\fontsize{6.5}{7}\selectfont 6: \textcolor{green!50!black}{\texttt{(c)}}};
\node[anchor=east] at (-0.25, -7*\cs) {\fontsize{6.5}{7}\selectfont 7: \textcolor{orange!80!black}{\texttt{<reg\_0>}}};
\node[anchor=east] at (-0.25, -8*\cs) {\fontsize{6.5}{7}\selectfont 8: \textcolor{orange!80!black}{\texttt{(d)}}};
\node[anchor=east] at (-0.25, -9*\cs) {\fontsize{6.5}{7}\selectfont 9: \textcolor{violet!80}{\texttt{<reg\_1>}}};
\node[anchor=east] at (-0.25, -10*\cs) {\fontsize{6.5}{7}\selectfont 10: \textcolor{violet!80}{\texttt{(e)}}};

\draw [decorate,decoration={brace,amplitude=3pt},xshift=2pt]
    (10*\cs+0.18, -1*\cs+0.15) -- (10*\cs+0.18, -2*\cs-0.15)
    node [midway,xshift=5pt,anchor=west] {\fontsize{6}{6}\selectfont Step (a)};

\draw [decorate,decoration={brace,amplitude=3pt},xshift=2pt]
    (10*\cs+0.18, -3*\cs+0.15) -- (10*\cs+0.18, -4*\cs-0.15)
    node [midway,xshift=5pt,anchor=west] {\fontsize{6}{6}\selectfont Step (b)};

\draw [decorate,decoration={brace,amplitude=3pt},xshift=2pt]
    (10*\cs+0.18, -5*\cs+0.15) -- (10*\cs+0.18, -6*\cs-0.15)
    node [midway,xshift=5pt,anchor=west] {\fontsize{6}{6}\selectfont Step (c)};

\draw [decorate,decoration={brace,amplitude=3pt},xshift=2pt]
    (10*\cs+0.18, -7*\cs+0.15) -- (10*\cs+0.18, -8*\cs-0.15)
    node [midway,xshift=5pt,anchor=west] {\fontsize{6}{6}\selectfont Step (d)};

\draw [decorate,decoration={brace,amplitude=3pt},xshift=2pt]
    (10*\cs+0.18, -9*\cs+0.15) -- (10*\cs+0.18, -10*\cs-0.15)
    node [midway,xshift=5pt,anchor=west] {\fontsize{6}{6}\selectfont Step (e)};

\node[attended, label={[label distance=0.05cm, font=\fontsize{6}{6}\selectfont]right:Attended}] at (1*\cs, -12*\cs) {};
\node[masked, label={[label distance=0.05cm, font=\fontsize{6}{6}\selectfont]right:Masked}] at (6*\cs, -12*\cs) {};

\end{tikzpicture}
\caption{Attention mask.}
\label{fig:tip:attention-mask}
\end{subfigure}
\caption{KV cache state and attention mask for the trace in \Cref{fig:tip:registers}.
(a)~tracks which reasoning step each thought register holds after each step completes.
(b)~shows the attention mask over the full token sequence, where filled cells are attended and white cells are masked.
Consider Step~(d): the model reuses \texttt{<reg\_0>} and Step~(a) becomes dead.
In~(a), the ``After~(d)'' row shows Step~(d) replacing Step~(a) in the \texttt{reg\_0} column while Steps~(b) and~(c) remain in their thought registers.
In~(b), rows~7--8 (\texttt{<reg\_0>} and Step~(d)) lose access to rows~1--2 (the dead Step~(a)): the white cells show that once a step becomes dead, the reused thought register token and all subsequent tokens can no longer attend to it.}
\label{fig:tip:mechanism}
\end{figure}

\paragraph{Paged Attention.}
vLLM~\parencite{kwon2023efficient} introduces paged attention, a memory management technique for LLM inference that solves a GPU memory waste problem.
During autoregressive decoding, the inference system maintains a KV cache for each request that grows token by token, and must decide how to allocate memory for it.
A naive approach reserves a large contiguous block upfront for the maximum possible sequence length, wasting most of the allocation because sequences rarely reach the maximum.
As requests arrive and leave, free memory fragments into small scattered gaps too small to serve new requests, even when total free memory would suffice.
Paged attention borrows the idea of paging from operating systems: instead of requiring one contiguous allocation per sequence, it divides KV cache memory into small, fixed-size blocks called \emph{pages}, each holding the keys and values for a fixed number of tokens.
A block table maps each sequence's logical token positions to the physical pages that store them, so a sequence's KV cache can span non-contiguous memory.
This design eliminates fragmentation because the allocator can assign any free page to any sequence, and limits over-allocation to at most one partially filled page per sequence because the inference system allocates pages incrementally as it generates tokens rather than reserving memory speculatively.
Paging lets the attention kernel skip any page whose tokens all belong to dead reasoning steps.
TIP exploits this mechanism for speedup.

\paragraph{Decode Loop.}
The runtime maintains a \emph{thought register table} that maps each thought register to the start and end positions of its current reasoning step and derives page retention from this table: the runtime retains a page if and only if it contains at least one live step's tokens.
On each decode step, the runtime samples a new token and checks whether the token is a thought register token.
If not, the current reasoning step grows by one token.
If it is, the runtime updates the thought register table: it replaces the previous reasoning step's position range with the new step's and evicts any page that no longer contains a live step's tokens.
\Cref{fig:tip:mechanism} illustrates this transition: when the model emits \texttt{<reg\_0>} a second time at Step~(d), Step~(a) becomes dead and the runtime evicts any pages that no longer contain a live step's tokens, then begins accumulating Step~(d) into thought register~0.

\paragraph{Page-Granular Eviction.}
When a step becomes dead, the runtime marks any page whose tokens all now belong to dead steps as evicted; the attention kernel skips evicted pages entirely.
Pages that mix dead and live steps remain retained, and all their tokens---including the dead step's---stay visible to attention.
I make a deliberate choice not to suppress dead tokens on retained pages, as doing so would require maintaining a per-token liveness mask on every retained page.
In \Cref{sec:tip:analysis:interp}, I present empirical justification for this design decision.
\Cref{fig:tip:attention-mask} illustrates the block attention mask for the trace in \Cref{fig:tip:registers}, under the simplifying assumption that each reasoning step aligns to a page boundary.

\paragraph{Incremental Mask Cache.}
The FlexAttention kernel requires a block-sparse attention mask that marks each page as retained or evicted; the kernel reads only retained pages.
Most decode steps extend the active thought register by one token within an already-retained page, leaving the set of retained pages unchanged.
The runtime detects this case and reuses the previous step's mask, rebuilding it only when the set of retained pages changes.

\subsection{Data Curation}
\label{sec:tip:data}

Training the model to use thought registers requires supervised data in which reasoning steps have thought register assignments.
I construct this data from a 5K subset of DeepMath-103K~\cite{deepmath} in two stages: step dependence annotation and thought register assignment.

\paragraph{Step Dependence Annotation.}
Given a reasoning trace, I prompt Gemini Flash to segment it into discrete reasoning steps (\Cref{sec:tip:ids}) and to annotate the \emph{dependence structure} among steps---that is, for each step, which previous steps it relies on.
This annotation produces a directed acyclic graph (DAG) over reasoning steps, where an edge from step $i$ to step $j$ indicates that step $i$ depends on the content of step $j$.
Steps that no future step depends on are candidates for eviction.
I include the full annotation prompts in \Cref{app:tip:annotation-prompts}.

\paragraph{Thought Register Assignment.}
Given the step dependence DAG and a budget of $N$ thought registers, I design a greedy algorithm to assign thought registers to reasoning steps:

\begin{enumerate}[leftmargin=*, topsep=0pt, itemsep=0pt]
    \item Walk through the reasoning steps in order. Maintain a set of $N$ available thought registers, initially all available.
    \item Assign each new step a thought register from the available set.
    \item After processing each step, check which live steps have no remaining future dependents. Mark their thought registers as available for reassignment.
    \item If no thought register is available when the algorithm must assign a new step, evict the step whose next dependent is furthest in the future (a B\'{e}l\'{a}dy-style optimal replacement policy \parencite{belady1966study}).
\end{enumerate}

This procedure inserts thought register tokens into each reasoning trace, encoding the eviction decisions that the dependence DAG prescribes.

\subsection{Training}
\label{sec:tip:training}

TIP training starts from a Qwen3-4B base model and proceeds in two stages.

\paragraph{Supervised Fine-Tuning (SFT).}
\looseness=-1
I fine-tune the model for 2 epochs on the 5K DeepMath-103K~\cite{deepmath} solutions after the curation pipeline (\Cref{sec:tip:data}) segments them into steps, annotates a dependence DAG, and inserts thought register tokens.
I use the Adam optimizer with a learning rate of $1 \times 10^{-5}$, a batch size of 4, and bfloat16 precision.
This stage trains the model to emit thought register tokens and write reasoning steps within a fixed thought register budget.

\paragraph{Reinforcement Learning (GRPO).}
Following SFT, I apply Group Relative Policy Optimization (GRPO) on the DeepScaleR dataset~\cite{deepscaler2025} for 50 training steps.
I use a group size of 16 with 16 rollouts per prompt and a learning rate of $1 \times 10^{-6}$ on a 4$\times$H200 node; training takes approximately 2 days.

\paragraph{Reward.}
The reward combines a correctness term with a KV-cache regularizer.
I prompt the model to produce its final answer inside a $\texttt{\textbackslash boxed\{\}}$ expression, extract the integer from there, and compare it against the ground truth after normalizing both.
A correct answer receives a base reward of 1.0; an incorrect but properly formatted answer receives 0.1; a response with no extracted answer receives 0.
I multiply the base reward by a KV regularization factor $1 - \bar{k} / L$, where $\bar{k}$ is the number of live KV tokens the model attends to averaged over decoding steps and $L$ is the maximum sequence length.
This factor linearly penalizes KV cache usage: a rollout that attends to the full context receives zero reward regardless of correctness, while one that evicts aggressively retains nearly the full base reward.

\section{Experimental Evaluation}
\label{sec:tip:experiments}

\looseness=-1
I evaluate TIP under paged attention~\cite{kwon2023efficient}, the memory layout used by modern inference servers. I implement all methods---TIP and baselines alike---on top of FlexAttention~\cite{dong2024flexattention} so they share the same attention kernel and differ only in their eviction policy. I measure accuracy against three efficiency metrics: live KV cache size, aggregate throughput, and decode time.

\paragraph{Datasets.}
I evaluate on AIME 2024 and AIME 2025, the official competition releases of the American Invitational Mathematics Examination. Each year contributes 30 problems (AIME-I and AIME-II combined), giving 60 problems total. Every problem admits a single integer answer. I prompt the model to place its final answer in a $\texttt{\textbackslash boxed\{\}}$ expression, extract the integer, and compare it against the official answer via exact match.

\paragraph{Decoding Strategies.}
I compare four decoding strategies.
(1) \emph{Dense} performs standard autoregressive decoding from the SFT model trained on DeepMath-103K without thought registers---the KV cache grows linearly with the number of generated tokens.
(2) \emph{StreamingLLM} (SLLM)~\cite{xiao2024streamingllm} applies a fixed sliding window over the KV cache combined with a handful of \emph{attention sinks}: the first four tokens of the sequence, which language models consistently attend to regardless of semantic content and which the runtime must preserve to avoid accuracy collapse.
(3) \emph{Paged H2O} (PH2O) is the paged-attention variant of H2O~\cite{zhang2023h2o}, which splits a fixed per-request budget evenly between the most recent tokens and the most important tokens according to accumulated attention scores. Under paged attention, PH2O evicts at page granularity and is therefore conceptually identical to ChunkKV~\parencite{liu2025chunkkv}.
(4) \emph{TIP} uses the model after supervised fine-tuning with $N=16$ thought registers followed by GRPO.
I evaluate SLLM and PH2O at three total budgets: 1024, 2048, and 4096 tokens.
I select StreamingLLM and H2O as baselines because both operate compatibly at page granularity under paged attention, matching the production serving setting TIP targets.

\paragraph{Inference Setup.}
For each problem in each benchmark, I sample 16 responses per decoding strategy with temperature $0.6$ and top-$p$ $0.95$, following the recommended settings from the Qwen documentation. I decode at a batch size of 16 with a page size of 64, the smallest size I found to work reliably in FlexAttention on H200. I report all metrics per GPU.

\paragraph{Metrics.}
I report four metrics for each decoding strategy:
(i) \emph{accuracy}---the fraction of samples per problem that produce the correct final answer, averaged across the benchmark;
(ii) \emph{live KV cache}---the mean number of live KV tokens visible to attention during decoding, averaged over the full generation;
(iii) \emph{throughput}---aggregate generated tokens per second across the decoding batch; and
(iv) \emph{decode time}---total wall-clock time to decode one benchmark.

\begin{figure}[t]
\centering
\includegraphics[width=\linewidth]{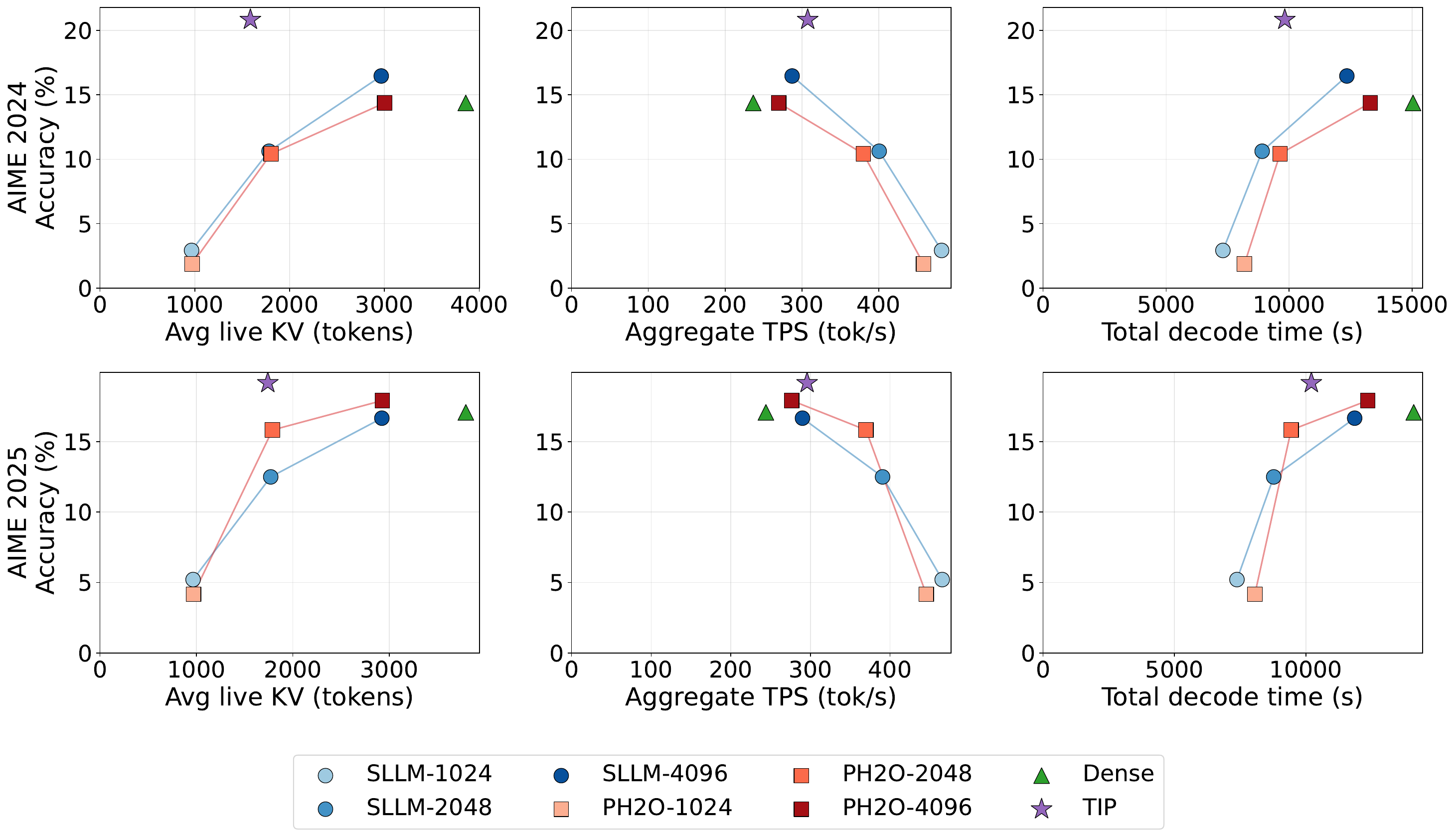}
\caption{Accuracy vs.\ live KV cache size (left), aggregate throughput (middle), and total decode time (right) on AIME 2024 (top) and AIME 2025 (bottom). TIP exceeds the Dense baseline accuracy while roughly halving the live KV cache and shortening decode time, and outperforms StreamingLLM (SLLM) and Paged H2O (PH2O) baselines at comparable memory footprints. I evaluate SLLM and PH2O at budgets of 1024, 2048, and 4096 tokens (shown as progressively darker shades).}
\label{fig:tip:accuracy-vs-kv}
\end{figure}

\paragraph{Live KV cache.}
The left column of \Cref{fig:tip:accuracy-vs-kv} plots mean accuracy versus average live KV cache size.
TIP exceeds Dense's accuracy while using roughly half the live KV cache.
Concretely, TIP reaches 20.8\% accuracy at 1586 live KV tokens on AIME 2024 and 19.2\% at 1740 tokens on AIME 2025, against Dense's 14.4\% at 3858 tokens and 17.1\% at 3794 tokens---a 59\% and 54\% cache reduction paired with 6.4-point and 2.1-point accuracy gains.
TIP also dominates the heuristic-eviction baselines at comparable budgets: SLLM-2048 and PH2O-2048 use slightly more cache than TIP (1783 and 1804 tokens on AIME 2024; 1770 and 1788 on AIME 2025) yet trail TIP by roughly 10 accuracy points on AIME 2024 and 7 / 3.4 points on AIME 2025.

\paragraph{Throughput.}
The middle column of \Cref{fig:tip:accuracy-vs-kv} plots mean accuracy versus aggregate throughput.
TIP achieves higher throughput than Dense while reaching higher accuracy.
Concretely, TIP runs at 308 tok/s on AIME 2024 and 296 tok/s on AIME 2025, against Dense's 237 and 244 tok/s---a 30\% and 21\% throughput gain.
At the 4096-token budget where SLLM and PH2O reach their highest accuracy, TIP also matches or beats their throughput while delivering 1.3--4.4 more accuracy points: 308 vs.\ 287 tok/s against SLLM-4096 on AIME 2024 and 296 vs.\ 277 tok/s against PH2O-4096 on AIME 2025.

\paragraph{Decode time.}
The right column of \Cref{fig:tip:accuracy-vs-kv} plots mean accuracy versus total decode time.
TIP shortens decode time relative to Dense while reaching higher accuracy.
Concretely, TIP completes the benchmark in 9{,}825s on AIME 2024 and 10{,}209s on AIME 2025, against Dense's 15{,}039s and 14{,}104s---a 35\% and 28\% reduction.
At the 4096-token budget where SLLM and PH2O reach their highest accuracy, TIP also finishes faster while delivering 1.3--4.4 more accuracy points: 9{,}825s vs.\ 12{,}349s against SLLM-4096 on AIME 2024 and 10{,}209s vs.\ 12{,}353s against PH2O-4096 on AIME 2025.

\paragraph{TIP Overhead.}
The throughput gain is smaller than the live KV reduction would suggest, for three reasons.
First, page fragmentation means the reduction in live KV cache size does not translate directly to a proportionate reduction in KV cache pages.
Second, every decoding step pays a bookkeeping cost to update the liveness of each reasoning step occupying a thought register, while no other baseline technique pays this overhead.
Third, reduced KV cache only reduces the execution time of dot-product attention, which is only one component of the forward pass.

\paragraph{Takeaway.}
TIP Pareto-dominates Dense and the best-accuracy budgets of SLLM and PH2O on accuracy, live KV cache, throughput, and decode time.

\section{Additional Analysis}
\label{sec:tip:analysis}

I present additional analysis on two components of TIP: the SFT data annotation method (\Cref{sec:tip:data}) and the runtime interpreter that maps thought register tokens to attention masks (\Cref{sec:tip:runtime}).

\subsection{SFT Data Annotation}
\label{sec:tip:analysis:data}

I examine two design decisions in the thought register allocation algorithm that produces the initial SFT data (\Cref{sec:tip:data}).

\begin{figure}[t]
\centering
\includegraphics[width=\linewidth]{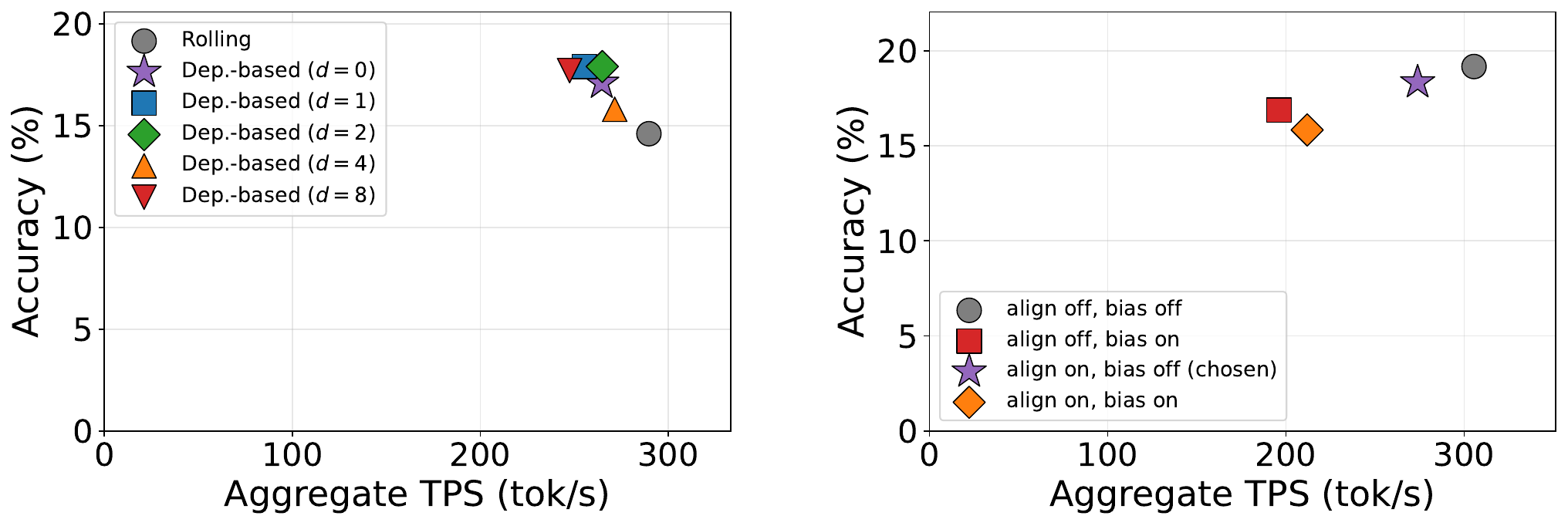}
\caption[Throughput vs.\ accuracy for the annotation-policy and interpreter-design ablations]{Throughput vs.\ accuracy (mean-of-16) on AIME 2025. (a)~Annotation policy: dependence-based variants cluster in the upper left (higher accuracy, modest throughput cost); rolling sits in the lower right. (b)~Interpreter design: bias on shifts each align column leftward by 22--36\% in throughput; align off shifts each bias column upward by 0.9--1.1 points in accuracy. A star marks the chosen setting (align off, bias off).}
\label{fig:tip:analysis-ablation}
\end{figure}

\paragraph{Method.}
I investigate two questions:
(a) does the dependence-based eviction policy outperform a na\"{i}ve rolling baseline, and
(b) does extending the eviction criterion from direct dependents to transitive dependents improve SFT model accuracy?
For (a), the \emph{rolling} baseline cycles through registers in order and always evicts the oldest, ignoring the dependence DAG entirely.
For (b), the algorithm in \Cref{sec:tip:data} considers only each step's direct dependents ($d{=}0$); I generalize this to transitive dependents by introducing a depth parameter $d$ that bounds how many hops the algorithm traverses in the DAG, and evaluate $d \in \{0, 1, 2, 4, 8\}$.
All other training and evaluation settings follow \Cref{sec:tip:training,sec:tip:experiments}.
\Cref{fig:tip:analysis-ablation}a plots aggregate throughput against accuracy for all configurations.

\paragraph{Rolling vs.\ dependence-based.}
The dependence-based policy ($d{=}0$) reaches 17.1\% versus 14.6\% for rolling, a 2.5-point gain.
Throughput decreases modestly: 264.6 tok/s versus 289.7 tok/s for rolling, an 8.7\% reduction.
The dependence DAG provides a meaningfully better training signal than a na\"{i}ve rolling policy.

\paragraph{Dependence depth.}
Increasing the dependence depth does not improve accuracy: the $d{=}1$ and $d{=}2$ variants both reach 17.9\%, the highest in the ablation.
Across depths, mean-of-16 spans 15.8--17.9\% with no monotone trend.
\paragraph{Takeaway.}
The dependence-based policy yields a consistent accuracy gain over rolling at a modest throughput cost.
Direct dependents alone produce effective training data; extending to transitive dependents does not improve accuracy.

\subsection{Interpreter Design}
\label{sec:tip:analysis:interp}

I analyze two implementation choices inside the TIP runtime:
(a) \emph{page alignment}---whether the runtime aligns each thought register's reasoning step to a KV cache page boundary, and
(b) \emph{dead-token suppression}---whether dead tokens on retained pages stay visible to attention.

\paragraph{Method.}
I evaluate four runtime configurations on AIME 2025, one for each combination of page alignment (on or off) and dead-token suppression (on or off), keeping all other sampling hyperparameters consistent with \Cref{sec:tip:experiments}.
I report the effect of each choice relative to the baseline with both disabled.
\Cref{fig:tip:analysis-ablation}b plots aggregate throughput against mean accuracy (mean-of-16) for each configuration.

\paragraph{Page Alignment.}
A reasoning step can start at any token position, so two or more thought registers can share a KV cache page.
Evicting one then cannot free the shared page without also evicting the other's live tokens---the page is fragmented.
With page alignment, the runtime rounds each reasoning step's start up to the nearest page boundary, so that each thought register occupies a disjoint set of pages.
Evicting a thought register then cleanly frees all of its pages.
The trade-off is that rounding up introduces a gap of unused positions between the end of one reasoning step and the page-aligned start of the next, reducing the model's effective context length.
The unaligned configuration beats the aligned configuration by 0.9 points on mean-of-16 (19.2\% vs.\ 18.3\%).
I disable page alignment: the accuracy gain outweighs the fragmentation cost, and preserving the full effective context length is more valuable than clean page-level eviction.

\paragraph{Dead-Token Suppression.}
If any token on a page is live, the runtime must retain the entire page.
Whether the dead tokens on a retained page receive attention remains an independent design choice.
Suppressing them corresponds precisely to the intended semantics: dead tokens belong to evicted reasoning steps and carry no information relevant to future generation.
However, maintaining and applying a per-token liveness bias within retained pages is costly.
Enabling suppression drops throughput from 305.5 to 196.1 tok/s (35.8\% reduction).
I therefore disable dead-token suppression.

\paragraph{Interaction.}
Enabling both page alignment and dead-token suppression simultaneously does not recover the losses of either choice: the combined configuration reaches 211.9 tok/s at 15.8\% mean-of-16, underperforming the baseline (305.5 tok/s, 19.2\%) on both metrics.

\paragraph{Takeaway.}
The TIP runtime disables both page alignment and dead-token suppression to maintain high accuracy at high throughput.

\section{Related Work}
\label{sec:tip:related}

\paragraph{KV-Cache Eviction.}
KV-cache eviction methods select entries for removal by scoring their importance, and differ primarily in whether that scoring function is hand-designed or learned.
Most methods define a fixed scoring function over cached tokens:
StreamingLLM retains a sliding window of recent tokens plus a small set of initial tokens \parencite{xiao2024streamingllm},
H2O retains tokens with the highest cumulative attention scores \parencite{zhang2023h2o},
Ada-KV allocates budgets adaptively across attention heads \parencite{feng2024adakv},
and RaaS evicts tokens by least-recent attention above a threshold \parencite{hu2025raas}.
Recent work replaces the hand-designed criterion with a learned one:
\textcite{moschella2025kvp} train per-head ranking agents to predict future token utility,
TRIM-KV learns a retain-or-drop gate at token creation time \parencite{bui2025trimkv},
and LookaheadKV trains modules to approximate future importance without draft generation \parencite{ahn2026lookaheadkv}.
These learned methods improve over heuristics but still optimize a surrogate for token importance rather than task performance itself.
TIP removes the surrogate entirely: reinforcement learning against final-answer accuracy trains the model to discover which evictions hurt the answer, without an intermediate importance estimator.

\paragraph{Semantically Driven Inference Optimization.}
Several methods exploit semantic dependence structure in the model's output to guide inference decisions.
Skeleton-of-Thought identifies parallelizable chunks via a planning prompt and expands them concurrently \parencite{ning2023skeleton};
APAR fine-tunes models to mark parallelizable segments in their own output \parencite{liu2024apar};
PASTA learns semantic dependence structure and uses it to parallelize generation \parencite{jin2025pasta};
and SPRINT extends asynchronous decoding to multi-step reasoning with alternating planning and execution rounds \parencite{biju2025sprint}.
Planned Diffusion derives semantic dependence structure from a generated plan to decompose diffusion sampling into independently denoised chunks \parencite{jin2026pd}.
TIP extends this principle from scheduling parallel generation to managing the KV cache.

\paragraph{KV-Cache Compression.}
A complementary line of work reduces memory by compressing the representation of cached entries rather than evicting them.
KIVI applies asymmetric 2-bit quantization to keys and values \parencite{liu2024kivi};
KVQuant develops RoPE-aware ultra-low-precision quantization for very long contexts \parencite{hooper2024kvquant};
and Palu projects keys and values into a low-rank latent space \parencite{chang2025palu}.
These methods reduce bytes per entry; TIP reduces the number of entries retained.
The two approaches are orthogonal and composable: a model can evict obsolete reasoning steps with TIP and store the remaining entries in fewer bits.

\section{Conclusion}
\label{sec:tip:conclusion}

This chapter introduces TIP, a system for learned context eviction: the model learns to annotate dead reasoning steps, and the runtime evicts the corresponding KV cache pages.
The evaluation on AIME 2024 and AIME 2025 shows that TIP improves accuracy over autoregressive decoding while cutting the average live KV cache roughly in half and shortening total decode time by 28--35\%, whereas heuristic-eviction baselines sacrifice accuracy at matched memory budgets.
The key insight---that a model can learn semantic dependence among its own reasoning steps and annotate it for the runtime to act on---extends the self-orchestrating paradigm from parallel generation (\Cref{ch:pasta,ch:pd}) to context eviction.
Together, PASTA, TIP, and Planned Diffusion demonstrate that models can learn to orchestrate inference execution---parallel generation and context eviction---by annotating semantic dependence in the text they produce.

\chapter{Planned Diffusion}
\label{ch:pd}

\looseness=-1
The previous two chapters train models to annotate semantic dependence for autoregressive inference: PASTA exploits dependence to parallelize decoding, and TIP exploits dependence to evict dead reasoning steps from the KV cache. Discrete diffusion language models, by contrast, produce multiple tokens per step. This chapter asks whether semantic dependence annotations remain useful for models that natively support parallel generation. I present \pd{}, which uses semantic dependence annotations to improve the quality-latency trade-off of parallel sampling for discrete diffusion language models.

\section{Introduction}
\label{sec:pd:intro}

Discrete diffusion language models~\cite{sahoo2024simple} generate tokens in parallel through iterative denoising, but sampling from them requires a \emph{denoising order} (\Cref{sec:pd:prelim}): a strategy for deciding which tokens to unmask at each step.
Existing approaches rely on heuristics such as random order or confidence-based thresholds.
These heuristics require many denoising steps to match autoregressive quality, yielding a suboptimal trade-off between latency and quality~\cite{wu2025fastdllm}.

\paragraph{Semantic Dependence for Scheduling.}
Typical language model responses contain semantically independent chunks. For example, in a response with a bulleted list, the model can denoise each bullet point concurrently~\cite{ning2023skeleton}. \pd{} exploits this by having the model identify independent chunks and denoise them in parallel.

\begin{figure}[t]
    \centering
    \includegraphics[width=\linewidth,clip,trim=25 195 25 10]{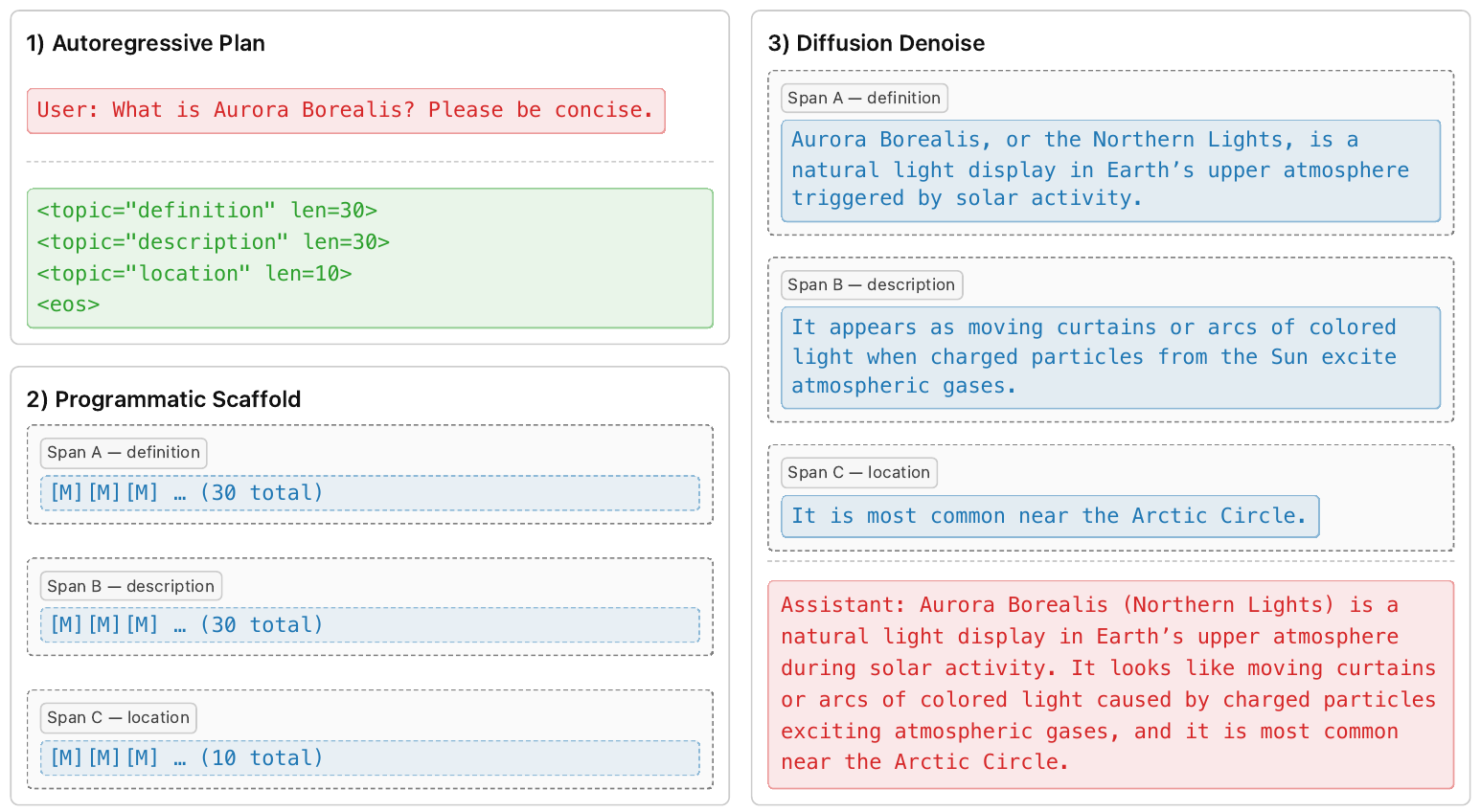}
    \caption{A real example of \pd{}. (1) The model first generates a sequential plan using semantic dependence annotations to define the structure and length of independent text chunks. (2) The runtime then translates this plan into a scaffold, initializing each chunk with a corresponding number of mask tokens. (3) The model denoises all chunks in parallel with diffusion, generating text for each section to produce the complete response.}
    \label{fig:pd:intro}
\end{figure}

\paragraph{Two-Stage Generation.}
First, in a sequential planning stage, the model operates autoregressively to generate a high-level \emph{execution plan} composed of semantic dependence annotations.
This plan partitions the task into a set of semantically independent sub-tasks.
Second, in a parallel diffusion stage, the model executes this plan, simultaneously generating the text for all planned chunks.
This single-model, hybrid approach offers a distinct architectural advantage over other acceleration techniques, such as speculative decoding~\cite{leviathan2023fast}, which requires a separate draft model.
To the best of my knowledge, this is the first text-only model trained with both discrete diffusion and autoregressive objectives.
\Cref{fig:pd:intro} presents a sample generation produced by the \pd{} model.

\paragraph{Contributions.}
I make the following contributions:
\begin{itemize}[leftmargin=*, topsep=0pt, itemsep=0pt]
    \item I introduce \pd{}, a new parallel generation technique that decomposes text generation into a sequential planning stage and a parallel diffusion stage.
    \item I design the annotation language, training method, and inference algorithm that enable a single model to perform this hybrid generation process.
    \item I demonstrate that \pd{} achieves a Pareto-optimal trade-off between quality and latency on AlpacaEval, a suite of 805 instruction-following prompts; it achieves a 1.27$\times$ to 1.81$\times$ speedup over autoregressive decoding while incurring only a 0.87\% to 5.4\% drop in win rate, respectively.
    \item I show that \pd{}'s instruction-following quality continues to improve with more finetuning compute, while the autoregressive baseline plateaus.
    \item I present sensitivity analysis validating that the planning mechanism is minimal and reliable, and show that simple runtime knobs offer tunable control over the quality--latency trade-off.
\end{itemize}

\section{Preliminaries}
\label{sec:pd:prelim}

Generative language models learn a probability distribution over sequences of discrete tokens. In this work, I focus on two main paradigms: autoregressive and discrete diffusion.

\paragraph{Autoregression.}
Autoregressive models are the standard for sequential text generation. They factorize the joint probability of a token sequence $x=(x_1, x_2,\cdots)$ as a product of conditional distributions. I denote this distribution $p_{\text{AR}}$:
\begin{equation}
\label{eq:pd:prelim-ar}
p_{\text{AR}}(x) = \prod_{i=1}^{|x|} p_\theta(x_i | x_{<i})
\end{equation}
where $p_\theta$ is a parameterized conditional distribution over tokens.

\paragraph{Discrete Diffusion.}
Discrete diffusion models learn to reverse a fixed data corruption process that gradually introduces noise into a clean sequence. For text, the corruption process replaces tokens with a special \emph{mask} token~\cite{austin2021structured}. Let $x^0$ be a clean sequence of tokens. The forward corruption process $q$ produces a noisy version $x^t$
at a timestep $t\in[0,1]$. The distribution of a corrupted sequence $x^t$ follows:
\begin{equation}
\label{eq:pd:forward}
    q_{t\mid 0}(x^t_i \mid x^0_i) =
    \begin{cases}
  t, & \text{if } x^t_i = \text{[MASK]} \\
  1-t, & \text{if } x^t_i = x^0_i  \\
  0 & \text{otherwise}
\end{cases}
~~~q_{t\mid 0}(x^t \mid x^0) = \prod_i q_{t\mid 0}(x^t_i\mid x^0_i)
\end{equation}
Because the true posterior over a forward noising process is intractable to compute exactly~\cite{lou2024discretediffusionmodelingestimating}, diffusion models approximate the reverse process by maximizing a variational lower bound on the log-likelihood~\cite{sahoo2024simple}.
\begin{equation}
\label{eq:pd:elbo}
    \log p_{\text{D}}(x^0) \geq \mathbb{E}_{t\sim U(0,1),x^t \sim q(x^t \mid x^0)} \frac{1}{t} \sum_i \mathbbm{1}(x^t_i=\text{[MASK]})\log
    p_{\theta}(x^0_i \mid x^t)
\end{equation}
Unlike autoregressive decoding, this objective decomposes over individual positions, permitting parallel token sampling. For brevity, when referring to diffusion I let $p_{\text{D}}(x)$ denote the data distribution implied by the learned reverse process.
Because the training objective decomposes over individual positions (\Cref{eq:pd:elbo}), for any disjoint sets of observed positions $\mathcal{O}$ and queried positions $\mathcal{Q} \subseteq [L]$, the conditional distribution over $\mathcal{Q}$ factorizes as
\begin{equation}
\label{eq:pd:conditional}
    p_{\text{D}}(x_{\mathcal{Q}} \mid x_{\mathcal{O}})
    = \prod_{i \in \mathcal{Q}} p_{\text{D}}(x_i \mid x_{\mathcal{O}}).
\end{equation}
In \pd{}, I condition both distributions on prior context tokens $c = (c_1, c_2, \cdots)$; I denote these conditional distributions as $p_{\text{AR}}(\cdot \mid c)$ and $p_{\text{D}}(\cdot \mid c)$.

\paragraph{Denoising Order.}
Sampling from a discrete diffusion model requires selecting a \emph{denoising order}~\cite{kim2025trainworstplanbest, turok2026duel}.
I define an \emph{ordered partition} of positions $[L] = \{1, \ldots, L\}$ as a tuple
$\sigma = (\sigma_1, \ldots, \sigma_T)$ of non-empty disjoint subsets with
$\bigcup_{t=1}^T \sigma_t = [L]$. I write $\sigma_{<t} = \sigma_1 \cup \cdots
\cup \sigma_{t-1}$ for positions revealed before step $t$. Positions within the
same subset decode in \emph{parallel}; positions in different subsets decode
\emph{sequentially}. A denoising order selects an ordered partition of $[L]$, determining how the joint distribution factors across decoding steps:
\begin{equation}
\label{eq:pd:diffusion-order}
    p_{\text{D}}(x \mid c;\,\sigma)
    = \prod_{t=1}^{T} p_{\text{D}}\!\left(x_{\sigma_t} \mid x_{\sigma_{<t}}, c\right).
\end{equation}
This follows from substituting $\mathcal{O} = \sigma_{<t}$ and
$\mathcal{Q} = \sigma_t$ into \Cref{eq:pd:conditional}. I first describe two standard denoising orders; both reveal one position per step ($|\sigma_t|=1$ for all $t$).
Let $\pi$ be a permutation of $[L]$ drawn uniformly at random. The uniform
ordered partition is
\begin{equation}
    \sigma_t^{\mathrm{Unif}} := \{\pi(t)\} \qquad t = 1, \ldots, L
\end{equation}
The ELBO samples $t$ uniformly and masks each position independently, treating all unmasking orders as equally likely. Uniform random unmasking therefore matches the training objective in \Cref{eq:pd:elbo}, but in practice it underperforms entropy-ordered unmasking~\cite{israel2025accelerating}. The entropy-ordered partition selects at each step $t$ the position with lowest predicted entropy:
\begin{equation}
\label{eq:pd:entropy-order}
    \sigma_t^{\mathrm{Ent}}
    := \operatorname*{arg\,min}_{i \,\notin\, \sigma_{<t}}
      H\!\bigl(p_{\mathrm{D}}(x_i \mid x_{\sigma_{<t}})\bigr) \qquad t = 1, \ldots, L
\end{equation}
where $H$ denotes Shannon entropy. Unlike $\sigma^{\mathrm{Unif}}$, the
partition is not fixed in advance but depends on the tokens revealed at each
step. \textcite{ye2025dream} use this as their default strategy.

\section{Planned Diffusion}
\label{sec:pd:method}

\pd{} alternates between autoregressive planning and parallel diffusion denoising. Each \emph{iteration} consists of a single planning stage followed by a single diffusion stage; later iterations may condition on the outputs of earlier ones. The plan segments the output into chunks and concisely describes each chunk's semantic content; the diffusion model then fills in the chunks in parallel. \pd{} composes two levels of ordering: across chunks, it decodes in parallel; within each chunk, it uses a separate denoising order.

I first formalize this two-stage structure. I then detail: (i) an annotation pipeline that augments instruction-following data with semantic dependence annotations, (ii) a training objective that combines autoregressive and diffusion losses, and (iii) an inference procedure that translates the model's semantic dependence annotations into a denoising order.

\subsection{Formal Description}
\label{sec:pd:formal}

I formalize a single iteration of \pd{}. Beginning from prior tokens
$c$, the model autoregressively generates a \emph{plan} $z$, which specifies
$K$ chunks, each with a semantic description, and predicted lengths $l_1, \ldots, l_K$. This induces a total sequence
length $L = \sum_{k=1}^{K} l_k$ and a partition of $[L]$ into sets of contiguous indices $s_1, \ldots, s_K$, where each $s_k = \{L_{k-1}+1 \,,\ldots,\, L_k\}$ is the set of indices starting from the end of the previous set until index $L_k = \sum_{j=1}^{k} l_j$ with $L_0 = 0$. I define an inner denoising order for each chunk by letting $\sigma^{(k)} = (\sigma^{(k)}_1, \ldots, \sigma^{(k)}_{T_k})$ be any ordered partition of $s_k$ into $T_k \leq l_k$ steps. The composed denoising order $\sigma^{\mathrm{Plan}}(z)$ merges the per-chunk denoising orders into a single denoising order over all $L$ positions:
\begin{equation}
\label{eq:pd:plan-order}
    \sigma^{\mathrm{Plan}}_t(z) = \bigcup_{\{k\,:\,T_k \geq t\}} \sigma^{(k)}_t, \qquad t = 1, \ldots, \max_k\, T_k
\end{equation}
which reveals positions from every chunk $s_k$ with $T_k \geq t$
simultaneously at step $t$. The diffusion phase generates content
$x \in \mathcal{V}^{L}$, and the joint distribution over plan and content is:
\begin{equation}
\label{eq:pd:joint}
    p_{\mathrm{PD}}(z, x \mid c;\,\sigma^{\mathrm{Plan}}(z))
    = \underbrace{p_{\mathrm{AR}}(z \mid c)}_{\text{Planning}}
      \cdot \underbrace{p_{\mathrm{D}}\!\left(x \mid z,\, c;\,\sigma^{\mathrm{Plan}}(z)\right)}_{\text{Diffusion}}
\end{equation}
Because $p_{\mathrm{D}}$ accepts any valid ordered partition of $[L]$, \pd{} decouples the choice of inner order $\sigma^{(k)}$ from the outer order over chunks. Setting each
$\sigma^{(k)} = \sigma^{\mathrm{Ent}}$ restricted to $s_k$ recovers the
entropy strategy applied independently within each chunk, as I do in experiments.

\Cref{alg:pd:algorithm} presents the full \pd{} sampling algorithm, which can perform multiple iterations of planning and diffusion.

\begin{algorithm}[t]
\caption{Planned Diffusion}
\label{alg:pd:algorithm}
\small
\begin{algorithmic}[1]
\Function{Planned\_Diffusion}{$c$}
    \Loop
        \State Sample plan $z \sim p_{\text{AR}}(\cdot \mid c)$ \Comment{Autoregressively generate plan}
        \State Parse $z$ to get $K$ chunks $s_1, \ldots, s_K$ with lengths $l_1, \ldots, l_K$
        \For{$k = 1, \ldots, K$} \textbf{in parallel}
            \State Sample $x_{s_k} \sim p_{\text{D}}\!\left(\cdot \mid z,\, c;\, \sigma^{(k)}\right)$ \Comment{Diffuse each chunk in parallel}
        \EndFor
        \State $x \gets (x_{s_1}, \ldots, x_{s_K})$ \Comment{Concatenate chunks together}
        \State $c \gets (c,\, z,\, x)$ \Comment{Append plan and output to prior tokens}
        \If{$z[\text{end}] = \pdeos{}$}
            \State \textbf{break} \Comment{Final planning token triggers end of generation}
        \EndIf
    \EndLoop
    \State $c \gets \func{RemoveControlTags}(c)$ \Comment{Strip planning tokens from final output}
    \State \Return $c$
\EndFunction
\end{algorithmic}
\end{algorithm}

\subsection{Data}
\label{sec:pd:data}

I first define the semantic dependence annotations that \pd{} uses to express its plan, then describe how I create annotated training data.

\paragraph{Semantic Dependence Annotations.}
During the sequential planning stage, the model uses a paired \pdtag{topic}\ldots\pdctag{topic} tag structure to define each chunk that the diffusion stage will generate in parallel.
Within this structure, the model generates a concise description of the chunk's content (e.g., ``definition'' in \Cref{fig:pd:intro}) and its predicted length (e.g., ``30'' in \Cref{fig:pd:intro}).
During the parallel diffusion stage, the model generates the tokens for each planned chunk within a corresponding \pdasync{}\ldots\pdcasync{} tag pair. Finally, the \pdsync{} tag conveys semantic dependence: tokens that follow \pdsync{} depend on the content of preceding \pdasync{} chunks, so the sampling algorithm continues sequential planning only after the model fills those chunks. I add all annotation tokens to the model's vocabulary for training and inference and strip them during post-processing of final outputs.

\paragraph{Finetuning Dataset.}
I annotate the SlimOrca instruction-finetuning dataset~\cite{SlimOrca} for parallel generation, following the LLM-based annotation approach in \Cref{ch:pasta}.
I prompt a \textsc{Gemini} model with the syntax and semantics of the semantic dependence annotations (\Cref{app:pd:annotation-prompt}).
\textsc{Gemini} inserts the annotations into the response from each instruction-response pair in the dataset.
The opening \pdasync{} carries two attributes: \texttt{topic} (a concise label, $\leq 3$ words) and \texttt{tokens} (a coarse length estimate, e.g., multiples of 10 tokens).
At inference time, \texttt{tokens} determines the number of mask tokens to allocate to each chunk; at training time, I use the ground-truth chunk length directly. To train robustness to length prediction error, I add 0--10 tokens of stochastic padding to each chunk during preprocessing.
I impose that every non-annotation token lies inside exactly one \pdasync{}\ldots\pdcasync{} chunk (no nesting, no overlap).
I validate well-formedness (balanced tags, coverage, non-overlap, attribute types/ranges) and discard malformed cases.
\Cref{fig:pd:attention-mask} contains a concrete example of the annotation language used by \pd{}.

\subsection{Training}
\label{sec:pd:training}

\paragraph{Training Loss.}
The goal of training is to maximize the joint probability over plan tokens and their content. Given an annotated dataset $\mathcal{D}$ where $Y \in \mathcal{D}$ is a single clean example, I decompose $Y$ into sets of planning tokens $Z$ and content tokens $X$, such that $Y = Z \cup X$.  $X^t$ is a noised sequence of tokens under noise distribution $q_{t\mid 0}$ (\Cref{eq:pd:forward}). Thus, $X^t$ contains masked tokens with probability $t$. Let $f_\theta$ denote the \pd{} model; $f_\theta(x,i)$ denotes the model's prediction at position $i$ given input $x$. Finally, let $M_i(X)$ be an attention masking function that takes as input a sequence $X$ and outputs a subset $M_i(X) \subseteq X$ of the sequence that $f_\theta$ has access to at a particular index $i$. I describe the attention masking in the following paragraph. With $\mathrm{CE}$ as the cross-entropy loss, the overall training objective is:
\begin{equation}
\label{eq:pd:loss}
    \mathcal{L}(\theta) =
    \operatorE\limits_{\substack{Y \sim \mathcal{D} \\ t \sim U(0,1)}}
    \frac{1}{|Y|}
    \sum_{y_i \in Y}
    \underbrace{\mathbbm{1}(y_i \in Z)
    \mathrm{CE}\!\left( f_\theta(y_{<i}, i),\, y_i \right)
    }_{\text{Autoregressive}}
    +
    \underbrace{
    \frac{1}{t}
    \mathbbm{1}(y_i \in X) \mathrm{CE}\! \left( f_\theta(M_i(X^t\cup Z), i),\, y_i \right)
    }_{\text{Diffusion}}
\end{equation}
In the training objective, the same noise parameterized by $t$ applies to chunks across multiple iterations of planning and diffusion, where future diffusion chunks condition on previously sampled diffusion chunks. This decision follows from the interpretation of a diffusion model as an any-order autoregressive model capable of supporting arbitrary conditional queries at inference time~\cite{shi2025simplifiedgeneralizedmaskeddiffusion}. \textcite{nie2025large} use the same technique to train Llada 7B with a diffusion objective but use it for semi-autoregressive block sampling at inference time.

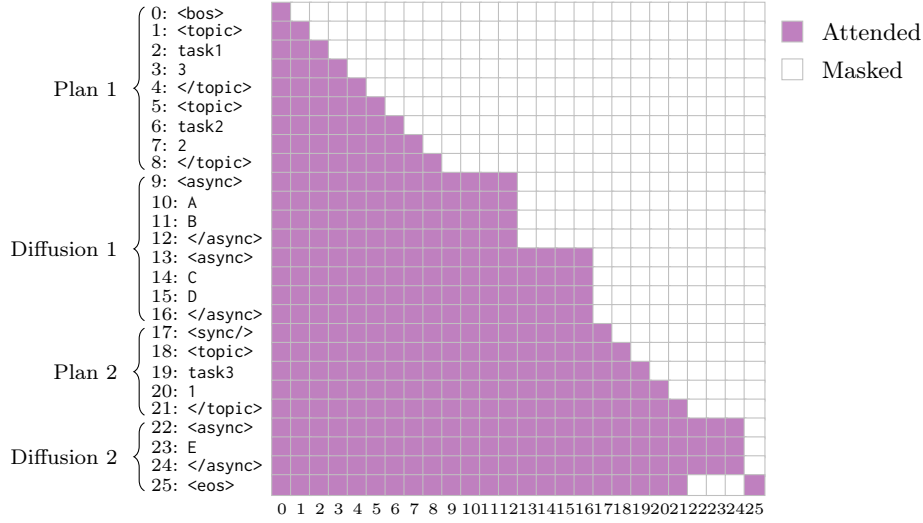
\begin{figure}[t]
\centering%
\begin{tikzpicture}[
    cell/.style={
        rectangle,
        draw=lightgray,
        minimum size=0.25cm,
    },
    attended/.style={
        cell,
        fill=violet!50,
    },
    masked/.style={
        cell,
        fill=white,
    },
    font={\fontsize{8.6}{9}\selectfont}
]

\def\cellsize{0.25cm}
\def\horizontalshift{0cm}

\begin{scope}[xshift=\horizontalshift]

    \def\matrixdata{
    1, 0, 0, 0, 0, 0, 0, 0, 0, 0, 0, 0, 0, 0, 0, 0, 0, 0, 0, 0, 0, 0, 0, 0, 0, 0,
    1, 1, 0, 0, 0, 0, 0, 0, 0, 0, 0, 0, 0, 0, 0, 0, 0, 0, 0, 0, 0, 0, 0, 0, 0, 0,
    1, 1, 1, 0, 0, 0, 0, 0, 0, 0, 0, 0, 0, 0, 0, 0, 0, 0, 0, 0, 0, 0, 0, 0, 0, 0,
    1, 1, 1, 1, 0, 0, 0, 0, 0, 0, 0, 0, 0, 0, 0, 0, 0, 0, 0, 0, 0, 0, 0, 0, 0, 0,
    1, 1, 1, 1, 1, 0, 0, 0, 0, 0, 0, 0, 0, 0, 0, 0, 0, 0, 0, 0, 0, 0, 0, 0, 0, 0,
    1, 1, 1, 1, 1, 1, 0, 0, 0, 0, 0, 0, 0, 0, 0, 0, 0, 0, 0, 0, 0, 0, 0, 0, 0, 0,
    1, 1, 1, 1, 1, 1, 1, 0, 0, 0, 0, 0, 0, 0, 0, 0, 0, 0, 0, 0, 0, 0, 0, 0, 0, 0,
    1, 1, 1, 1, 1, 1, 1, 1, 0, 0, 0, 0, 0, 0, 0, 0, 0, 0, 0, 0, 0, 0, 0, 0, 0, 0,
    1, 1, 1, 1, 1, 1, 1, 1, 1, 0, 0, 0, 0, 0, 0, 0, 0, 0, 0, 0, 0, 0, 0, 0, 0, 0,
    1, 1, 1, 1, 1, 1, 1, 1, 1, 1, 1, 1, 1, 0, 0, 0, 0, 0, 0, 0, 0, 0, 0, 0, 0, 0,
    1, 1, 1, 1, 1, 1, 1, 1, 1, 1, 1, 1, 1, 0, 0, 0, 0, 0, 0, 0, 0, 0, 0, 0, 0, 0,
    1, 1, 1, 1, 1, 1, 1, 1, 1, 1, 1, 1, 1, 0, 0, 0, 0, 0, 0, 0, 0, 0, 0, 0, 0, 0,
    1, 1, 1, 1, 1, 1, 1, 1, 1, 1, 1, 1, 1, 0, 0, 0, 0, 0, 0, 0, 0, 0, 0, 0, 0, 0,
    1, 1, 1, 1, 1, 1, 1, 1, 1, 1, 1, 1, 1, 1, 1, 1, 1, 0, 0, 0, 0, 0, 0, 0, 0, 0,
    1, 1, 1, 1, 1, 1, 1, 1, 1, 1, 1, 1, 1, 1, 1, 1, 1, 0, 0, 0, 0, 0, 0, 0, 0, 0,
    1, 1, 1, 1, 1, 1, 1, 1, 1, 1, 1, 1, 1, 1, 1, 1, 1, 0, 0, 0, 0, 0, 0, 0, 0, 0,
    1, 1, 1, 1, 1, 1, 1, 1, 1, 1, 1, 1, 1, 1, 1, 1, 1, 0, 0, 0, 0, 0, 0, 0, 0, 0,
    1, 1, 1, 1, 1, 1, 1, 1, 1, 1, 1, 1, 1, 1, 1, 1, 1, 1, 0, 0, 0, 0, 0, 0, 0, 0,
    1, 1, 1, 1, 1, 1, 1, 1, 1, 1, 1, 1, 1, 1, 1, 1, 1, 1, 1, 0, 0, 0, 0, 0, 0, 0,
    1, 1, 1, 1, 1, 1, 1, 1, 1, 1, 1, 1, 1, 1, 1, 1, 1, 1, 1, 1, 0, 0, 0, 0, 0, 0,
    1, 1, 1, 1, 1, 1, 1, 1, 1, 1, 1, 1, 1, 1, 1, 1, 1, 1, 1, 1, 1, 0, 0, 0, 0, 0,
    1, 1, 1, 1, 1, 1, 1, 1, 1, 1, 1, 1, 1, 1, 1, 1, 1, 1, 1, 1, 1, 1, 0, 0, 0, 0,
    1, 1, 1, 1, 1, 1, 1, 1, 1, 1, 1, 1, 1, 1, 1, 1, 1, 1, 1, 1, 1, 1, 1, 1, 1, 0,
    1, 1, 1, 1, 1, 1, 1, 1, 1, 1, 1, 1, 1, 1, 1, 1, 1, 1, 1, 1, 1, 1, 1, 1, 1, 0,
    1, 1, 1, 1, 1, 1, 1, 1, 1, 1, 1, 1, 1, 1, 1, 1, 1, 1, 1, 1, 1, 1, 1, 1, 1, 0,
    1, 1, 1, 1, 1, 1, 1, 1, 1, 1, 1, 1, 1, 1, 1, 1, 1, 1, 1, 1, 1, 1, 0, 0, 0, 1
    }

    \begin{scope}[xstep=\cellsize, ystep=\cellsize]
        \foreach \cell [count=\n from 0] in \matrixdata {
            \pgfmathtruncatemacro{\y}{floor(\n / 26)}
            \pgfmathtruncatemacro{\x}{mod(\n, 26)}

            \ifnum\cell=1
                \node[attended] at (\x*\cellsize, -\y*\cellsize) {};
            \else
                \node[masked] at (\x*\cellsize, -\y*\cellsize) {};
            \fi
        }
    \end{scope}

    \foreach \y in {0,...,25} {
        \node at (\y*\cellsize, -25.4*\cellsize) [anchor=north] {\fontsize{6}{6}\selectfont\y};
    }

    \node at (-7.5*\cellsize, 0*\cellsize) [anchor=west] {\fontsize{7}{7}\selectfont 0: \texttt{\textless{}bos\textgreater{}}};
    \node at (-7.5*\cellsize, -1*\cellsize) [anchor=west] {\fontsize{7}{7}\selectfont 1: \texttt{\textless{}topic\textgreater{}}};
    \node at (-7.5*\cellsize, -2*\cellsize) [anchor=west] {\fontsize{7}{7}\selectfont 2: \texttt{task1}};
    \node at (-7.5*\cellsize, -3*\cellsize) [anchor=west] {\fontsize{7}{7}\selectfont 3: \texttt{3}};
    \node at (-7.5*\cellsize, -4*\cellsize) [anchor=west] {\fontsize{7}{7}\selectfont 4: \texttt{\textless{}/topic\textgreater{}}};
    \node at (-7.5*\cellsize, -5*\cellsize) [anchor=west] {\fontsize{7}{7}\selectfont 5: \texttt{\textless{}topic\textgreater{}}};
    \node at (-7.5*\cellsize, -6*\cellsize) [anchor=west] {\fontsize{7}{7}\selectfont 6: \texttt{task2}};
    \node at (-7.5*\cellsize, -7*\cellsize) [anchor=west] {\fontsize{7}{7}\selectfont 7: \texttt{2}};
    \node at (-7.5*\cellsize, -8*\cellsize) [anchor=west] {\fontsize{7}{7}\selectfont 8: \texttt{\textless{}/topic\textgreater{}}};

    \node at (-7.5*\cellsize, -9*\cellsize) [anchor=west] {\fontsize{7}{7}\selectfont 9: \texttt{\textless{}async\textgreater{}}};
    \node at (-7.5*\cellsize, -10*\cellsize) [anchor=west] {\fontsize{7}{7}\selectfont 10: \texttt{A}};
    \node at (-7.5*\cellsize, -11*\cellsize) [anchor=west] {\fontsize{7}{7}\selectfont 11: \texttt{B}};
    \node at (-7.5*\cellsize, -12*\cellsize) [anchor=west] {\fontsize{7}{7}\selectfont 12: \texttt{\textless{}/async\textgreater{}}};
    \node at (-7.5*\cellsize, -13*\cellsize) [anchor=west] {\fontsize{7}{7}\selectfont 13: \texttt{\textless{}async\textgreater{}}};
    \node at (-7.5*\cellsize, -14*\cellsize) [anchor=west] {\fontsize{7}{7}\selectfont 14: \texttt{C}};
    \node at (-7.5*\cellsize, -15*\cellsize) [anchor=west] {\fontsize{7}{7}\selectfont 15: \texttt{D}};
    \node at (-7.5*\cellsize, -16*\cellsize) [anchor=west] {\fontsize{7}{7}\selectfont 16: \texttt{\textless{}/async\textgreater{}}};

    \node at (-7.5*\cellsize, -17*\cellsize) [anchor=west] {\fontsize{7}{7}\selectfont 17: \texttt{\textless{}sync/\textgreater{}}};
    \node at (-7.5*\cellsize, -18*\cellsize) [anchor=west] {\fontsize{7}{7}\selectfont 18: \texttt{\textless{}topic\textgreater{}}};
    \node at (-7.5*\cellsize, -19*\cellsize) [anchor=west] {\fontsize{7}{7}\selectfont 19: \texttt{task3}};
    \node at (-7.5*\cellsize, -20*\cellsize) [anchor=west] {\fontsize{7}{7}\selectfont 20: \texttt{1}};
    \node at (-7.5*\cellsize, -21*\cellsize) [anchor=west] {\fontsize{7}{7}\selectfont 21: \texttt{\textless{}/topic\textgreater{}}};

    \node at (-7.5*\cellsize, -22*\cellsize) [anchor=west] {\fontsize{7}{7}\selectfont 22: \texttt{\textless{}async\textgreater{}}};
    \node at (-7.5*\cellsize, -23*\cellsize) [anchor=west] {\fontsize{7}{7}\selectfont 23: \texttt{E}};
    \node at (-7.5*\cellsize, -24*\cellsize) [anchor=west] {\fontsize{7}{7}\selectfont 24: \texttt{\textless{}/async\textgreater{}}};
    \node at (-7.5*\cellsize, -25*\cellsize) [anchor=west] {\fontsize{7}{7}\selectfont 25: \texttt{\textless{}eos\textgreater{}}};

    \draw [decorate,decoration={brace,amplitude=4pt},xshift=-2pt]
    (-7*\cellsize, -8*\cellsize-0.3*\cellsize) --
    (-7*\cellsize, 0*\cellsize+0.3*\cellsize)
    node [midway,xshift=-6pt,anchor=east] {\fontsize{8}{8}\selectfont Plan 1};

    \draw [decorate,decoration={brace,amplitude=4pt},xshift=-2pt]
    (-7*\cellsize, -21*\cellsize-0.3*\cellsize) --
    (-7*\cellsize, -17*\cellsize+0.3*\cellsize)
    node [midway,xshift=-6pt,anchor=east] {\fontsize{8}{8}\selectfont Plan 2};

    \draw [decorate,decoration={brace,amplitude=4pt},xshift=-2pt]
    (-7*\cellsize, -16*\cellsize-0.3*\cellsize) --
    (-7*\cellsize, -9*\cellsize+0.3*\cellsize)
    node [midway,xshift=-6pt,anchor=east] {\fontsize{8}{8}\selectfont Diffusion 1};

    \draw [decorate,decoration={brace,amplitude=4pt},xshift=-2pt]
    (-7*\cellsize, -25*\cellsize-0.3*\cellsize) --
    (-7*\cellsize, -22*\cellsize+0.3*\cellsize)
    node [midway,xshift=-6pt,anchor=east] {\fontsize{8}{8}\selectfont Diffusion 2};

    \node[attended, label={[label distance=0.1cm]right:Attended}] at (27*\cellsize, -1*\cellsize) {};
    \node[masked, label={[label distance=0.1cm]right:Masked}] at (27*\cellsize, -3*\cellsize) {};

\end{scope}

\end{tikzpicture}
    \caption{The attention mask for \pd{} combines causal and bidirectional attention. Sequential planning stages use causal attention. \pdasync{} chunks use bidirectional attention for parallel denoising, with concurrent chunks enabled to cross-attend. After a \pdsync{} token, subsequent tokens can attend to all prior tokens. For illustrative purposes, this example shortens the diffusion chunks.}
    \label{fig:pd:attention-mask}
\end{figure}

\paragraph{Attention Mask.}
I implement the following rules via the attention mask $M_i$. Planning tokens receive causal attention (Plan 1 and 2 in \Cref{fig:pd:attention-mask}). Diffusion tokens use bidirectional attention, as required for diffusion-based parallel denoising (Diffusion 1 and 2 in \Cref{fig:pd:attention-mask}).
In \Cref{sec:pd:experiments}, I also show that restricting bidirectional dense attention within each chunk is another way to train \pd{}. I refer to this variant as \pdsa{}, which enforces full independence between chunks. The attention mask for \pdsa{} appears in \Cref{fig:pd:sparse-attention-mask}.

\subsection{Inference}
\label{sec:pd:inference}

\paragraph{Variable-Length Denoising.}
\looseness=-1 Typically, diffusion models generate with a fixed number of denoising steps. More denoising steps improve quality but reduce speed. Unlike vanilla diffusion, \pd{} does not generate a predetermined number of tokens, so the number of denoising steps cannot be fixed. I define a parameter $r$ called the \emph{step ratio}. Given a generation length $|x|$, the number of denoising steps is $s= r \cdot |x|$. For a plan $z$ and multiple diffusion chunks of lengths $l_1, \ldots, l_K$, the step ratio defines the step count as $s= r \cdot \max_k l_k$. The number of denoising steps depends only on the length of the longest chunk, because tokens in each chunk denoise in parallel to other chunks. Higher step ratio corresponds to higher quality and slower generation, while lower step ratio leads to faster generation but lower quality.

\section{Experimental Evaluation}
\label{sec:pd:experiments}

I experimentally assess the performance of \pd{}, focusing on its trade-off between generation quality and latency.
The results show that \pd{} expands the latency-quality Pareto frontier for text generation when compared to autoregressive and diffusion baselines.
Furthermore, I demonstrate that the method scales better with additional compute: \pd{} continues to improve with more training, whereas the performance of the autoregressive baseline plateaus.

\paragraph{Training Setup.}
\textcite{qwen2025qwen25technicalreport} first pre-train Dream-7B-Base autoregressively and \textcite{ye2025dream} further pre-train it with a diffusion objective; I fine-tune this checkpoint.
I train with AdamW~\cite{kingma2017adammethodstochasticoptimization,loshchilov2019decoupledweightdecayregularization}, peak learning rate $5\times10^{-5}$ with linear decay, and bfloat16 precision.
I use per-GPU batch size $1$ and global batch size $4$.
Because autoregressive and diffusion language models have different optimal epoch counts~\cite{prabhudesai2025diffusionbeatsautoregressivedataconstrained}, I sweep epochs over $\{2, 4, 8, 16\}$.
I fine-tune on Gemini-annotated SlimOrca instruction-following data (\Cref{sec:pd:data}); I keep the semantic dependence annotations for \pd{} and strip them for autoregressive and diffusion baselines.
Training runs on $4\times$H200 (141\,GB) with PyTorch~\cite{imambi2021pytorch} and HuggingFace~\cite{wolf2020huggingfacestransformersstateoftheartnatural}.

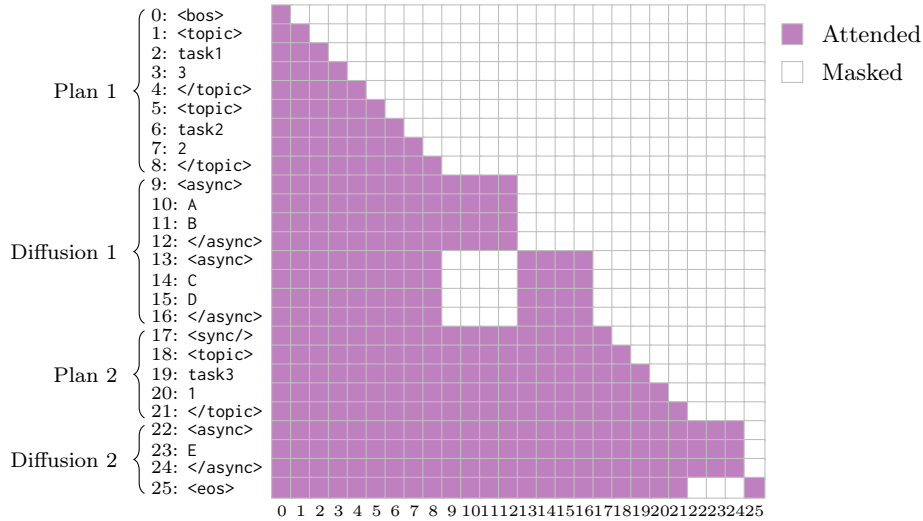
\begin{figure}[h!]
\centering%
\begin{tikzpicture}[
    cell/.style={
        rectangle,
        draw=lightgray,
        minimum size=0.25cm,
    },
    attended/.style={
        cell,
        fill=violet!50,
    },
    masked/.style={
        cell,
        fill=white,
    },
    font={\fontsize{8.6}{9}\selectfont}
]

\def\cellsize{0.25cm}
\def\horizontalshift{0cm}

\begin{scope}[xshift=\horizontalshift]

    \def\matrixdata{
    1, 0, 0, 0, 0, 0, 0, 0, 0, 0, 0, 0, 0, 0, 0, 0, 0, 0, 0, 0, 0, 0, 0, 0, 0, 0,
    1, 1, 0, 0, 0, 0, 0, 0, 0, 0, 0, 0, 0, 0, 0, 0, 0, 0, 0, 0, 0, 0, 0, 0, 0, 0,
    1, 1, 1, 0, 0, 0, 0, 0, 0, 0, 0, 0, 0, 0, 0, 0, 0, 0, 0, 0, 0, 0, 0, 0, 0, 0,
    1, 1, 1, 1, 0, 0, 0, 0, 0, 0, 0, 0, 0, 0, 0, 0, 0, 0, 0, 0, 0, 0, 0, 0, 0, 0,
    1, 1, 1, 1, 1, 0, 0, 0, 0, 0, 0, 0, 0, 0, 0, 0, 0, 0, 0, 0, 0, 0, 0, 0, 0, 0,
    1, 1, 1, 1, 1, 1, 0, 0, 0, 0, 0, 0, 0, 0, 0, 0, 0, 0, 0, 0, 0, 0, 0, 0, 0, 0,
    1, 1, 1, 1, 1, 1, 1, 0, 0, 0, 0, 0, 0, 0, 0, 0, 0, 0, 0, 0, 0, 0, 0, 0, 0, 0,
    1, 1, 1, 1, 1, 1, 1, 1, 0, 0, 0, 0, 0, 0, 0, 0, 0, 0, 0, 0, 0, 0, 0, 0, 0, 0,
    1, 1, 1, 1, 1, 1, 1, 1, 1, 0, 0, 0, 0, 0, 0, 0, 0, 0, 0, 0, 0, 0, 0, 0, 0, 0,
    1, 1, 1, 1, 1, 1, 1, 1, 1, 1, 1, 1, 1, 0, 0, 0, 0, 0, 0, 0, 0, 0, 0, 0, 0, 0,
    1, 1, 1, 1, 1, 1, 1, 1, 1, 1, 1, 1, 1, 0, 0, 0, 0, 0, 0, 0, 0, 0, 0, 0, 0, 0,
    1, 1, 1, 1, 1, 1, 1, 1, 1, 1, 1, 1, 1, 0, 0, 0, 0, 0, 0, 0, 0, 0, 0, 0, 0, 0,
    1, 1, 1, 1, 1, 1, 1, 1, 1, 1, 1, 1, 1, 0, 0, 0, 0, 0, 0, 0, 0, 0, 0, 0, 0, 0,
    1, 1, 1, 1, 1, 1, 1, 1, 1, 0, 0, 0, 0, 1, 1, 1, 1, 0, 0, 0, 0, 0, 0, 0, 0, 0,
    1, 1, 1, 1, 1, 1, 1, 1, 1, 0, 0, 0, 0, 1, 1, 1, 1, 0, 0, 0, 0, 0, 0, 0, 0, 0,
    1, 1, 1, 1, 1, 1, 1, 1, 1, 0, 0, 0, 0, 1, 1, 1, 1, 0, 0, 0, 0, 0, 0, 0, 0, 0,
    1, 1, 1, 1, 1, 1, 1, 1, 1, 0, 0, 0, 0, 1, 1, 1, 1, 0, 0, 0, 0, 0, 0, 0, 0, 0,
    1, 1, 1, 1, 1, 1, 1, 1, 1, 1, 1, 1, 1, 1, 1, 1, 1, 1, 0, 0, 0, 0, 0, 0, 0, 0,
    1, 1, 1, 1, 1, 1, 1, 1, 1, 1, 1, 1, 1, 1, 1, 1, 1, 1, 1, 0, 0, 0, 0, 0, 0, 0,
    1, 1, 1, 1, 1, 1, 1, 1, 1, 1, 1, 1, 1, 1, 1, 1, 1, 1, 1, 1, 0, 0, 0, 0, 0, 0,
    1, 1, 1, 1, 1, 1, 1, 1, 1, 1, 1, 1, 1, 1, 1, 1, 1, 1, 1, 1, 1, 0, 0, 0, 0, 0,
    1, 1, 1, 1, 1, 1, 1, 1, 1, 1, 1, 1, 1, 1, 1, 1, 1, 1, 1, 1, 1, 1, 0, 0, 0, 0,
    1, 1, 1, 1, 1, 1, 1, 1, 1, 1, 1, 1, 1, 1, 1, 1, 1, 1, 1, 1, 1, 1, 1, 1, 1, 0,
    1, 1, 1, 1, 1, 1, 1, 1, 1, 1, 1, 1, 1, 1, 1, 1, 1, 1, 1, 1, 1, 1, 1, 1, 1, 0,
    1, 1, 1, 1, 1, 1, 1, 1, 1, 1, 1, 1, 1, 1, 1, 1, 1, 1, 1, 1, 1, 1, 1, 1, 1, 0,
    1, 1, 1, 1, 1, 1, 1, 1, 1, 1, 1, 1, 1, 1, 1, 1, 1, 1, 1, 1, 1, 1, 0, 0, 0, 1
    }

    \begin{scope}[xstep=\cellsize, ystep=\cellsize]
        \foreach \cell [count=\n from 0] in \matrixdata {
            \pgfmathtruncatemacro{\y}{floor(\n / 26)}
            \pgfmathtruncatemacro{\x}{mod(\n, 26)}

            \ifnum\cell=1
                \node[attended] at (\x*\cellsize, -\y*\cellsize) {};
            \else
                \node[masked] at (\x*\cellsize, -\y*\cellsize) {};
            \fi
        }
    \end{scope}

    \foreach \y in {0,...,25} {
        \node at (\y*\cellsize, -25.4*\cellsize) [anchor=north] {\fontsize{6}{6}\selectfont\y};
    }

    \node at (-7.5*\cellsize, 0*\cellsize) [anchor=west] {\fontsize{7}{7}\selectfont 0: \texttt{\textless{}bos\textgreater{}}};
    \node at (-7.5*\cellsize, -1*\cellsize) [anchor=west] {\fontsize{7}{7}\selectfont 1: \texttt{\textless{}topic\textgreater{}}};
    \node at (-7.5*\cellsize, -2*\cellsize) [anchor=west] {\fontsize{7}{7}\selectfont 2: \texttt{task1}};
    \node at (-7.5*\cellsize, -3*\cellsize) [anchor=west] {\fontsize{7}{7}\selectfont 3: \texttt{3}};
    \node at (-7.5*\cellsize, -4*\cellsize) [anchor=west] {\fontsize{7}{7}\selectfont 4: \texttt{\textless{}/topic\textgreater{}}};
    \node at (-7.5*\cellsize, -5*\cellsize) [anchor=west] {\fontsize{7}{7}\selectfont 5: \texttt{\textless{}topic\textgreater{}}};
    \node at (-7.5*\cellsize, -6*\cellsize) [anchor=west] {\fontsize{7}{7}\selectfont 6: \texttt{task2}};
    \node at (-7.5*\cellsize, -7*\cellsize) [anchor=west] {\fontsize{7}{7}\selectfont 7: \texttt{2}};
    \node at (-7.5*\cellsize, -8*\cellsize) [anchor=west] {\fontsize{7}{7}\selectfont 8: \texttt{\textless{}/topic\textgreater{}}};

    \node at (-7.5*\cellsize, -9*\cellsize) [anchor=west] {\fontsize{7}{7}\selectfont 9: \texttt{\textless{}async\textgreater{}}};
    \node at (-7.5*\cellsize, -10*\cellsize) [anchor=west] {\fontsize{7}{7}\selectfont 10: \texttt{A}};
    \node at (-7.5*\cellsize, -11*\cellsize) [anchor=west] {\fontsize{7}{7}\selectfont 11: \texttt{B}};
    \node at (-7.5*\cellsize, -12*\cellsize) [anchor=west] {\fontsize{7}{7}\selectfont 12: \texttt{\textless{}/async\textgreater{}}};
    \node at (-7.5*\cellsize, -13*\cellsize) [anchor=west] {\fontsize{7}{7}\selectfont 13: \texttt{\textless{}async\textgreater{}}};
    \node at (-7.5*\cellsize, -14*\cellsize) [anchor=west] {\fontsize{7}{7}\selectfont 14: \texttt{C}};
    \node at (-7.5*\cellsize, -15*\cellsize) [anchor=west] {\fontsize{7}{7}\selectfont 15: \texttt{D}};
    \node at (-7.5*\cellsize, -16*\cellsize) [anchor=west] {\fontsize{7}{7}\selectfont 16: \texttt{\textless{}/async\textgreater{}}};

    \node at (-7.5*\cellsize, -17*\cellsize) [anchor=west] {\fontsize{7}{7}\selectfont 17: \texttt{\textless{}sync/\textgreater{}}};
    \node at (-7.5*\cellsize, -18*\cellsize) [anchor=west] {\fontsize{7}{7}\selectfont 18: \texttt{\textless{}topic\textgreater{}}};
    \node at (-7.5*\cellsize, -19*\cellsize) [anchor=west] {\fontsize{7}{7}\selectfont 19: \texttt{task3}};
    \node at (-7.5*\cellsize, -20*\cellsize) [anchor=west] {\fontsize{7}{7}\selectfont 20: \texttt{1}};
    \node at (-7.5*\cellsize, -21*\cellsize) [anchor=west] {\fontsize{7}{7}\selectfont 21: \texttt{\textless{}/topic\textgreater{}}};

    \node at (-7.5*\cellsize, -22*\cellsize) [anchor=west] {\fontsize{7}{7}\selectfont 22: \texttt{\textless{}async\textgreater{}}};
    \node at (-7.5*\cellsize, -23*\cellsize) [anchor=west] {\fontsize{7}{7}\selectfont 23: \texttt{E}};
    \node at (-7.5*\cellsize, -24*\cellsize) [anchor=west] {\fontsize{7}{7}\selectfont 24: \texttt{\textless{}/async\textgreater{}}};
    \node at (-7.5*\cellsize, -25*\cellsize) [anchor=west] {\fontsize{7}{7}\selectfont 25: \texttt{\textless{}eos\textgreater{}}};

    \draw [decorate,decoration={brace,amplitude=4pt},xshift=-2pt]
    (-7*\cellsize, -8*\cellsize-0.3*\cellsize) --
    (-7*\cellsize, 0*\cellsize+0.3*\cellsize)
    node [midway,xshift=-6pt,anchor=east] {\fontsize{8}{8}\selectfont Plan 1};

    \draw [decorate,decoration={brace,amplitude=4pt},xshift=-2pt]
    (-7*\cellsize, -21*\cellsize-0.3*\cellsize) --
    (-7*\cellsize, -17*\cellsize+0.3*\cellsize)
    node [midway,xshift=-6pt,anchor=east] {\fontsize{8}{8}\selectfont Plan 2};

    \draw [decorate,decoration={brace,amplitude=4pt},xshift=-2pt]
    (-7*\cellsize, -16*\cellsize-0.3*\cellsize) --
    (-7*\cellsize, -9*\cellsize+0.3*\cellsize)
    node [midway,xshift=-6pt,anchor=east] {\fontsize{8}{8}\selectfont Diffusion 1};

    \draw [decorate,decoration={brace,amplitude=4pt},xshift=-2pt]
    (-7*\cellsize, -25*\cellsize-0.3*\cellsize) --
    (-7*\cellsize, -22*\cellsize+0.3*\cellsize)
    node [midway,xshift=-6pt,anchor=east] {\fontsize{8}{8}\selectfont Diffusion 2};

    \node[attended, label={[label distance=0.1cm]right:Attended}] at (27*\cellsize, -1*\cellsize) {};
    \node[masked, label={[label distance=0.1cm]right:Masked}] at (27*\cellsize, -3*\cellsize) {};

\end{scope}

\end{tikzpicture}
    \caption{The sparse attention mask for \pdsa{}. Unlike the dense attention mask in \Cref{fig:pd:attention-mask}, concurrent \pdasync{} chunks cannot attend to each other during denoising, enforcing full independence between chunks.}
    \label{fig:pd:sparse-attention-mask}
\end{figure}

\paragraph{Decoding Strategies.}
I compare seven decoding strategies.
(1) \emph{Autoregressive} samples tokens sequentially from the autoregressive model.
(2) \emph{Diffusion} samples masked tokens in parallel from the diffusion model; I configure the number of denoising steps to equal the number of new tokens as this produces the highest quality generation~\cite{lou2024discretediffusionmodelingestimating,shi2025simplifiedgeneralizedmaskeddiffusion,sahoo2024simple}.
(3) \fastdllm{}~\cite{wu2025fastdllm} samples from the same diffusion model with an inference-time optimization, using half the number of new tokens as denoising steps and a default confidence threshold of 0.9 from \textcite{wu2025fastdllm}.
(4) \pastasft{}~\cite{jin2025pasta} uses semantic dependence annotations to implement semantic parallelism in a purely autoregressive setting. I compare to the SFT-trained version of \pasta{}, as opposed to RL-trained, for fair comparison with \pd{}, which I train only with SFT.
(5) \emph{Skeleton-of-Thought}~\cite{ning2023skeleton} autoregressively samples a bullet-point outline, then applies regular-expression-based syntactic pattern matching to extract points for parallel expansion.
(6) \pdsft{} (ours) samples a plan autoregressively from the \pd{} model, then samples masked tokens within each chunk in parallel from the same model; for each chunk, I configure denoising steps to equal its predicted token count.
\looseness=-1
(7) \pdsa{} (ours) is a variant of \pd{} that treats concurrent chunks as fully independent conditioned on the plan and denoises them with block-sparse attention (cf.\ \Cref{fig:pd:sparse-attention-mask}), reducing computational cost at the expense of model expressivity and GPU utilization efficiency.

\paragraph{Inference Setup.}
I sample with temperature $0.2$ and top-$p$ $0.95$ following \textcite{ye2025dream}, and cap sequence length at $2048$ tokens.
Inference runs on the same H200 hardware configuration.

\paragraph{Benchmark and Metrics.}
I evaluate on AlpacaEval (805 instruction-following prompts)~\cite{alpaca_eval, dubois2024length}.
For each method I report: (i) \emph{average latency}---the mean wall-clock time per response; and (ii) \emph{quality}---length-controlled win rate (LCWR) with an LLM-as-judge.
I use the recommended default configuration from \textcite{dubois2024length} due to its high correlation with human preference.
I set the LCWR reference to the best autoregressive baseline.
I identify this reference by evaluating the autoregressive variants (2, 4, 8, and 16 epochs) against the 2-epoch variant and choosing the model with the highest quality.\footnote{They tie on length-controlled win rate, so I break ties using raw win rates.}
The 16-epoch model wins and serves as the fixed reference for all LCWR scores.

\begin{figure}[t]
  \centering
  \begin{subfigure}[t]{0.32\linewidth}
    \centering
    \includegraphics[width=\linewidth]{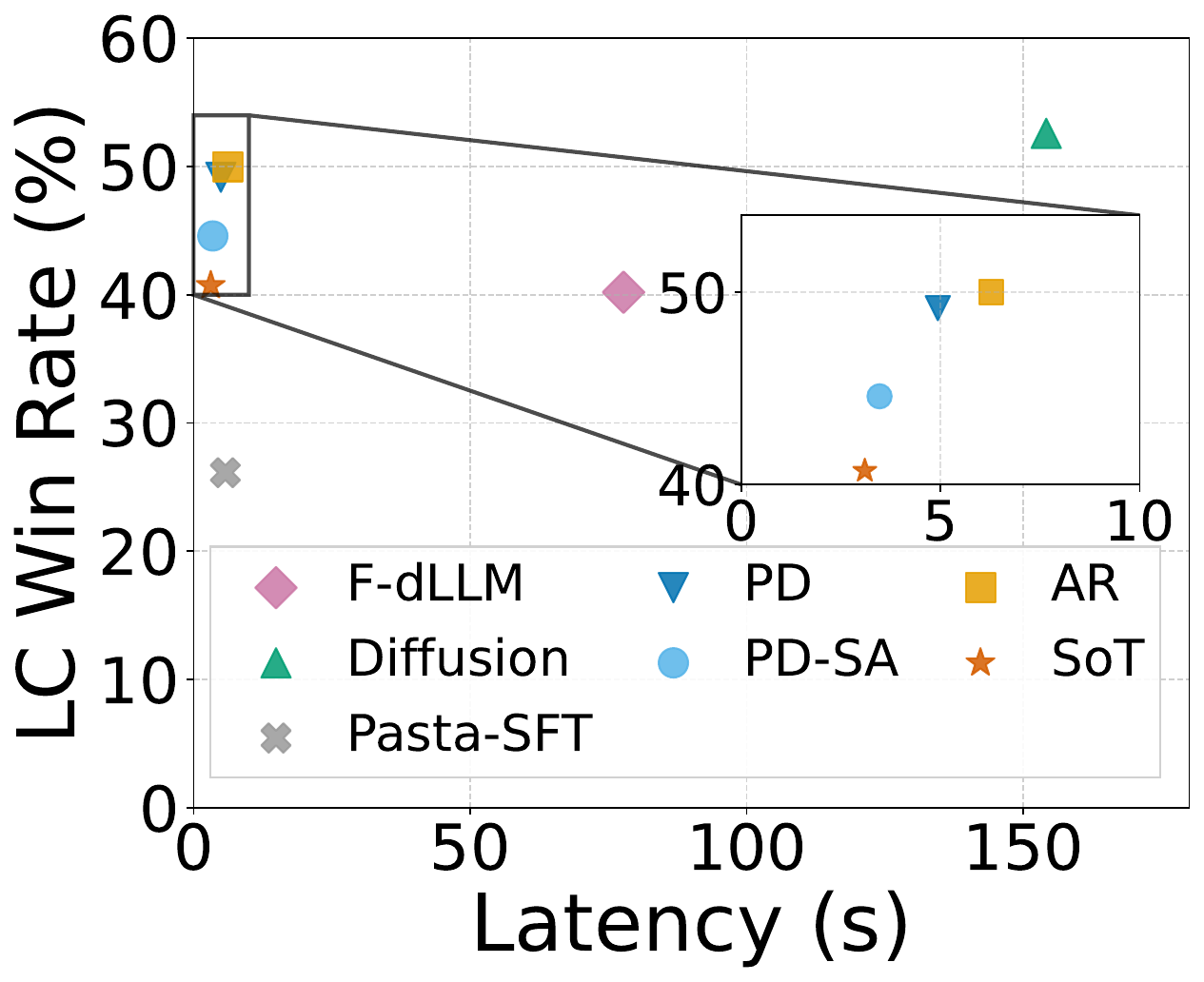}
  \end{subfigure}\hfill
  \begin{subfigure}[t]{0.32\linewidth}
    \centering
    \includegraphics[width=\linewidth]{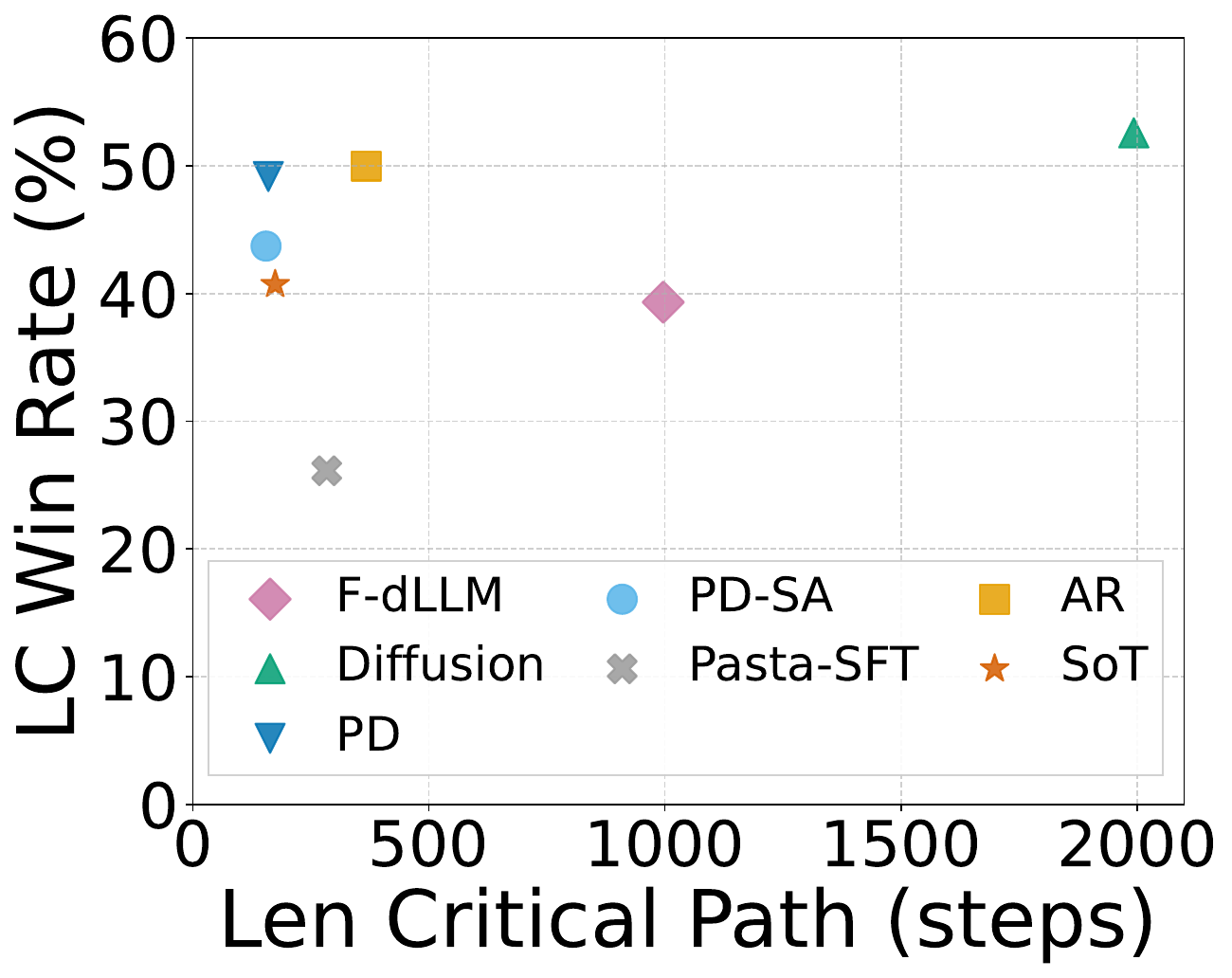}
  \end{subfigure}\hfill
  \begin{subfigure}[t]{0.32\linewidth}
    \centering
    \includegraphics[width=\linewidth]{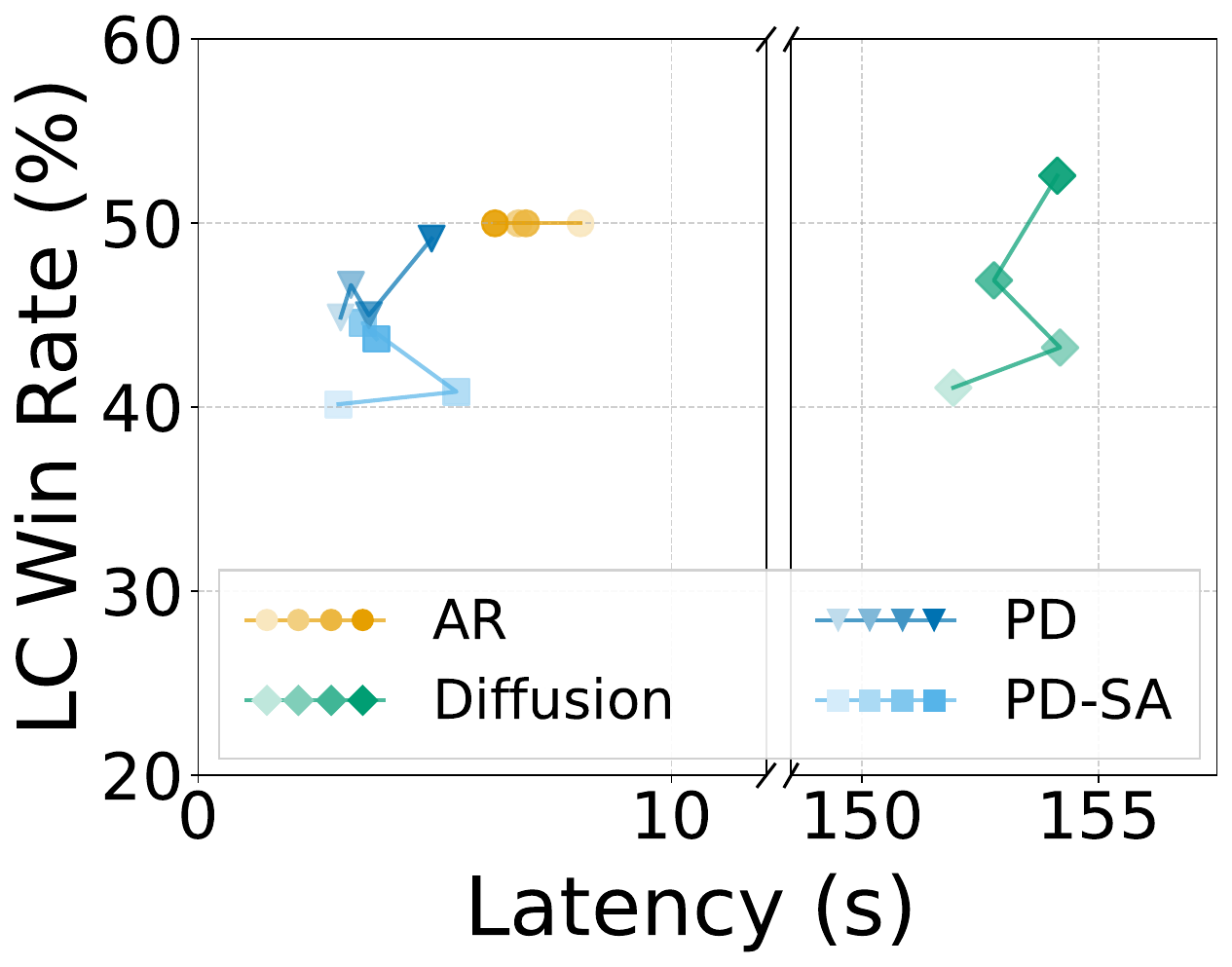}
  \end{subfigure}
  \caption{Evaluation of \pd{} on the AlpacaEval benchmark. Left: A comparison of latency versus length-controlled win rate shows \pd{} establishing a new Pareto frontier, offering a better quality--latency trade-off. Middle: An analysis of the average critical path length reveals that \pd{} requires substantially fewer sequential forward passes than autoregressive decoding. Right: A scaling analysis shows that \pd{}'s win rate continues to improve with more training, while the autoregressive baseline's performance flatlines. Within each method, color brightness from lightest to darkest encodes 2, 4, 8, and 16 training epochs. AR: Autoregressive; F-dLLM: \fastdllm{}; SoT: Skeleton-of-Thought; PD: \pdsft{}; PD-SA: \pdsa{}.}
  \label{fig:pd:main-results}
\end{figure}

\paragraph{Quality and Latency.}
I plot the latency--quality trade-off in the left panel of \Cref{fig:pd:main-results}.
\pdsft{} achieves a 49.2\% length-controlled win rate at 1.27$\times$ speedup relative to autoregressive decoding.
The sparse attention variant \pdsa{} trades quality for speed, achieving a 1.81$\times$ speedup while attaining a 44.6\% length-controlled win rate against the autoregressive reference (50.0\%).
Both variants outperform \fastdllm{}: \pdsa{} achieves a 22.4$\times$ speedup over \fastdllm{} and higher quality (44.6\% vs.\ 40.2\% length-controlled win rate).
Diffusion attains the highest quality (52.6\%) but requires more inference time, using 25$\times$ the latency of autoregressive decoding. Among semantic parallelism methods, \pd{} is firmly on the Pareto frontier. Skeleton-of-Thought offers marginally faster generations but incurs a significant quality decrease. Without additional RL training, \pastasft{} does not strategically use the annotation language to deliver a quality--latency advantage.

\paragraph{Speedup Analysis.}
I attribute much of \pd{}'s speedup over autoregressive decoding to its shorter \emph{critical path} of generation.
I define critical path length as the number of forward passes required to produce the final response. In a diffusion LM, the number of denoising steps $s$ determines the critical path, so the total complexity is the cost of attention multiplied by the number of iterations: $O(n^2 \cdot s)$, where $n$ is the sequence length. In the absence of \pdsync{} tokens, \pd{} reduces the number of steps from $n$ to $l_{\max}$, where $l_{\max}$ is the length of the longest chunk within a plan.
The middle panel of \Cref{fig:pd:main-results} shows that, on AlpacaEval, the average critical path of autoregressive decoding is 2.3$\times$ as long as that of \pdsft{} (367.3 vs.\ 160.0 steps) and 2.8$\times$ as long as the sparse attention variant \pdsa{} (367.3 vs.\ 155.2 steps).
\looseness=-1
Expectedly, \pd{} enables multiple chunks to denoise simultaneously.
The realized speedup (1.85$\times$) is smaller than the critical path reduction (2.8$\times$ for \pdsa{}, 2.3$\times$ for \pdsft{}) because each \pd{} step does more work: KV cache reuse is lower and per-step compute is heavier than an autoregressive token step. I also observe a difference in the total number of tokens generated (including annotation tokens and pad tokens) between \pd{} and autoregressive decoding. \pdsft{} and \pdsa{} produce approximately 9.1\%\footnote{The gap in output length is in part due to difference in optimal training epochs for \pd{} (8 epochs) and autoregressive (16 epochs). The output length difference between 16-epoch \pd{} and autoregressive reduces to 5.6\%.} and 3.4\% fewer tokens, respectively.

\paragraph{Scaling.}
The right panel of \Cref{fig:pd:main-results} shows how the latency-quality trade-off evolves with training epochs for the four training configurations I examine.
Autoregressive training shows no benefit from additional training compute: the length-controlled win rate remains flat at 50.0\% across 2, 4, 8, and 16 epochs. Both variants of \pd{} benefit moderately from additional training compute. \pdsa{} improves from 40.2\% (2 epochs) to 43.7\% (16 epochs), a gain of 3.5 percentage points, while \pdsft{} improves from 44.9\% (2 epochs) to 49.2\% (16 epochs), a gain of 4.3 percentage points.
Diffusion benefits significantly from additional training compute, rising from 41.1\% (2 epochs) to 52.6\% (16 epochs), a gain of 11.5 percentage points that ultimately surpasses the autoregressive baseline.

\paragraph{Takeaway.}
\pd{} sets a new latency--quality Pareto frontier and continuously improves with more training, while the autoregressive baseline plateaus.

\section{Additional Analysis}
\label{sec:pd:analysis}

I present additional analysis on several design decisions key to \pd{}.

\begin{figure}[t]
    \centering
    \begin{subfigure}[t]{0.31\linewidth}
        \includegraphics[width=\linewidth]{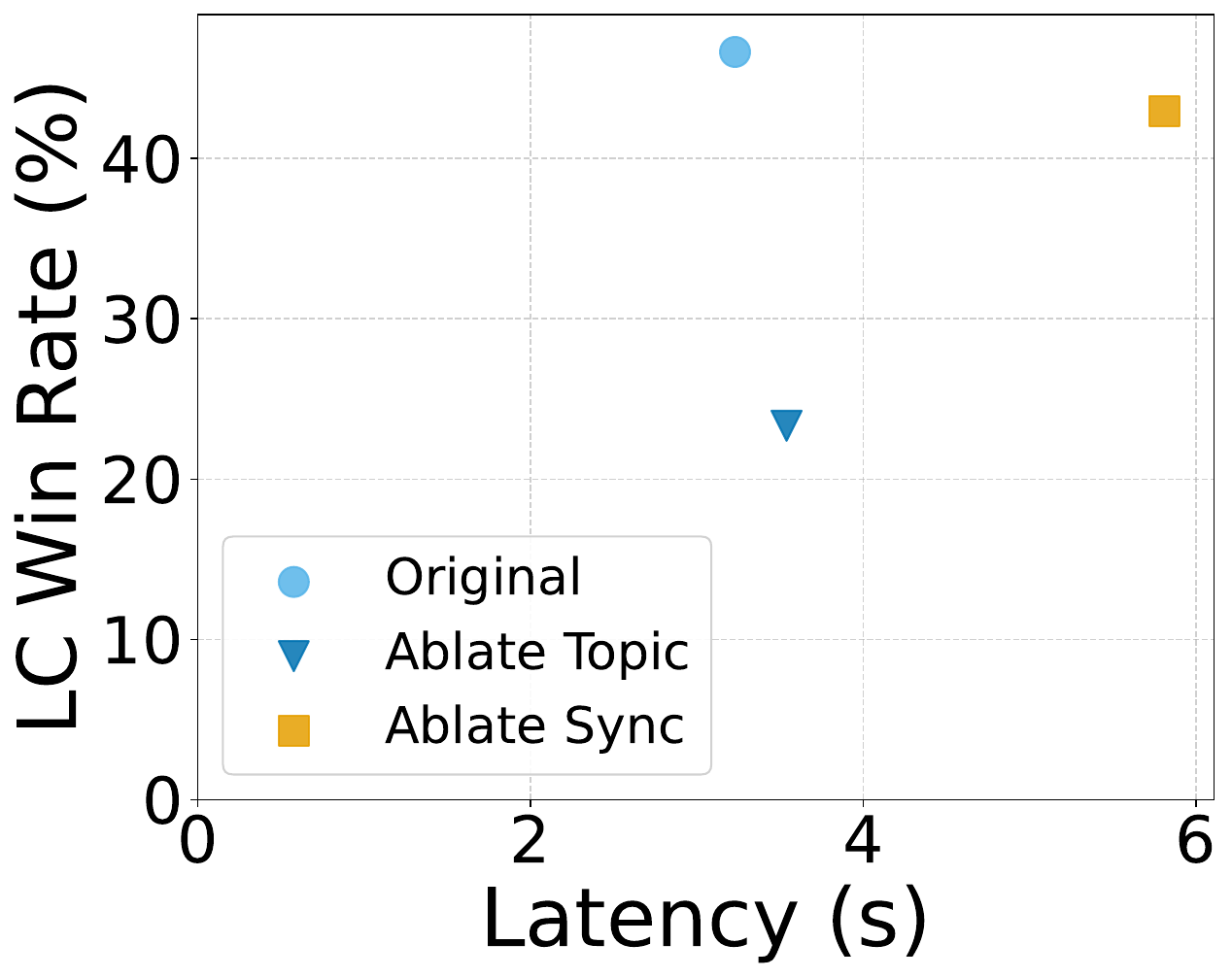}
        \label{fig:pd:plan-ablation}
    \end{subfigure}\hfill
    \begin{subfigure}[t]{0.32\linewidth}
        \includegraphics[width=\linewidth]{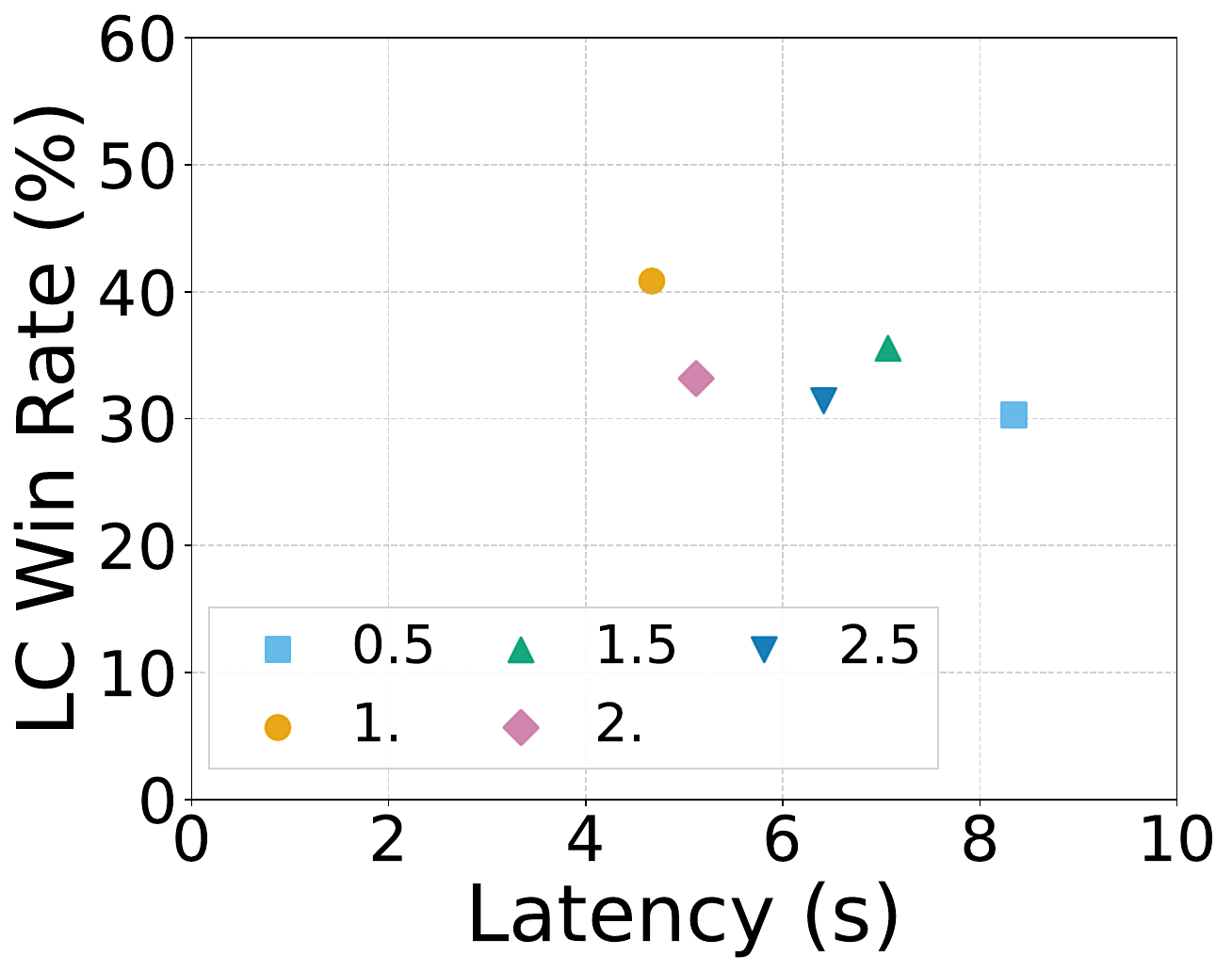}
        \label{fig:pd:length-scaling}
    \end{subfigure}\hfill
    \begin{subfigure}[t]{0.31\linewidth}
        \includegraphics[width=\linewidth]{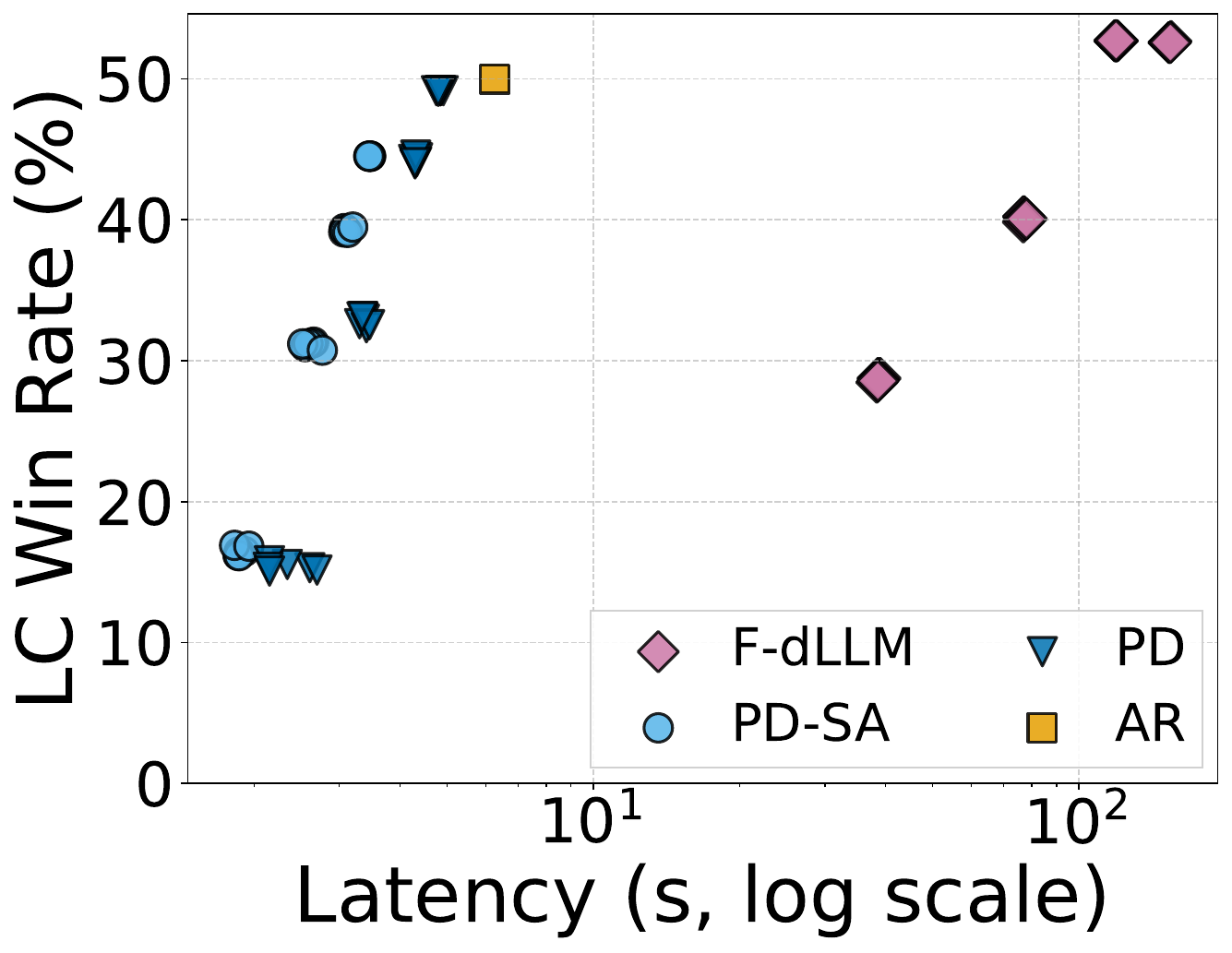}
        \label{fig:pd:fast-dllm-sweep}
    \end{subfigure}
   \caption{Additional analysis. Left (Plan ablation): Removing \texttt{topic} from the autoregressive plan harms quality, yet removing \pdsync{} tokens boosts speedup without significantly hurting quality.
   Middle (Chunk lengths): LCWR peaks at a length-scaling factor of 1.0; length prediction in the autoregressive plan does not exhibit systematic error.
   Right (Quality versus Latency Sweep): Varying the step ratio ($r=\{0.25, 0.5, 0.75, 1\}$) and the confidence threshold ($\tau=\{0.4, 0.5, 0.6, 0.7, 0.8, 0.9\}$) hyperparameters produces a smooth quality--latency trade-off.}
    \label{fig:pd:analysis}
\end{figure}

\paragraph{Plan Ablation.}
I remove two components of the autoregressive plan and evaluate their impact on inference quality and latency.
I train all model variants for 4 epochs in this ablation study.
The original 4-epoch \pdsft{} achieves a length-controlled win rate (LCWR) of 46.65\% at 3.23\,s latency.
Removing the \texttt{topic} attribute from the training data and re-training the \pdsft{} model significantly harms inference quality.
Specifically, it reduces LCWR from 46.65\% to 23.33\% and increases latency from 3.23 seconds to 3.54 seconds.
I conclude that \texttt{topic} attributes are critical for maintaining inference quality.

Recall that \pdsync{} marks a synchronization barrier: decoding beyond it begins only after the model finalizes all prior content.
Deleting all \pdsync{} tokens from the training data and re-training the \pdsft{} model modestly reduces quality.
Specifically, the LCWR falls from 46.65\% to 42.96\% while latency increases from 3.23\,s to 5.81\,s.
I conclude that omitting \pdsync{} hurts both quality and latency.

\paragraph{Chunk Lengths.}
I test whether the model predicts chunk lengths accurately.
Accurate length prediction is key to achieving good generation quality for \pd{}, as the diffusion denoising phase of generation cannot alter the chunk length.
Systematic over-prediction wastes time by adding masks and denoising steps, while systematic under-prediction harms quality by forcing content truncation.

To test for potential systematic deviation from the optimal generation length,
I multiply the model's predicted chunk length by a length-scaling factor to set the number of masks, sweeping the factor over $\{0.5, 1.0, 1.5, 2.0, 2.5\}$. I then measure LCWR and latency under identical inference settings.
Latency rises with factors above $1.0$ as expected because larger chunks require more mask tokens and denoising steps.
Quality peaks at the model's originally predicted length (i.e., scaling factor of 1.0).
Deviating from the predicted length in either direction reduces LCWR.
Interestingly, additional denoising steps do not improve quality.
The model's length predictions are accurate; I do not observe systematic over/under-prediction.

\paragraph{Quality versus Latency Sweep.}
By default, to maximize generation quality, I set the number of denoising steps for a chunk equal to that chunk's length, and I unmask a position only when the model is highly confident of its prediction.
In this analysis, I measure the quality--latency trade-off by varying the number of denoising steps as well as the confidence threshold for decoding.

In my experimental setup, all diffusion-based variants use the step ratio hyperparameter $r$ (\Cref{sec:pd:inference}), which sets the number of denoising steps to $s = r \cdot \max_k l_k$, where $l_k$ is the length of chunk $k$.
Because all chunks denoise in parallel, the total number of steps depends only on the longest chunk.
When $r = 1$ the model takes one denoising step per token in the longest chunk; smaller $r$ uses fewer steps at the cost of quality.
The confidence threshold $\tau$ used by \textcite{wu2025fastdllm} selects which positions to unmask at each denoising step.
For each masked position, if the model's top-token probability at that position is at least $\tau$, I decode that token and unmask the position; otherwise I keep it masked for later decoding.

\looseness=-1
Sweeping $r$ over $\{0.25, 0.5, 0.75, 1\}$ and $\tau$ over $\{0.4, 0.5, 0.6, 0.7, 0.8, 0.9\}$ yields a smooth trade-off between generation quality and inference latency. Across the step ratio and the confidence threshold, \pdsft{} and \pdsa{} yield higher quality at equal or lower latency across most operating points.

\paragraph{Takeaway.}
\looseness=-1
The analysis validates that the planning mechanism is minimal and reliable, and that the step ratio and confidence threshold provide tunable control of the quality--latency trade-off at inference time.

\paragraph{Table Summary of Evaluations.}
\label{sec:pd:table-summary}
For completeness, I provide a table summary of the results in \Cref{fig:pd:main-results}.
\begin{table}[h]
    \centering
    \caption{Comparison of methods based on latency and length-controlled win rate.}
    \label{tab:pd:method-comparison}
    \begin{tabular}{lcc}
        \toprule
        Method & Latency (s) & LC Win Rate (\%) \\
        \midrule
        Diffusion & 154.124 & 52.59\% \\
        AR & 6.272 & 50.00\% \\
        \fastdllm{}~\cite{wu2025fastdllm} & 77.723 & 40.20\% \\
        \pastasft{}~\cite{jin2025pasta} & 5.740 & 26.13\% \\
        Skeleton-of-Thought~\cite{ning2023skeleton} & 3.101 & 40.72\% \\
        \pdsa{} & 3.471 & 44.59\% \\
        \pdsft{} & 4.934 & 49.17\% \\
        \bottomrule
    \end{tabular}
\end{table}

\begin{figure}[h!]
    \centering
    \includegraphics[width=0.8\linewidth]{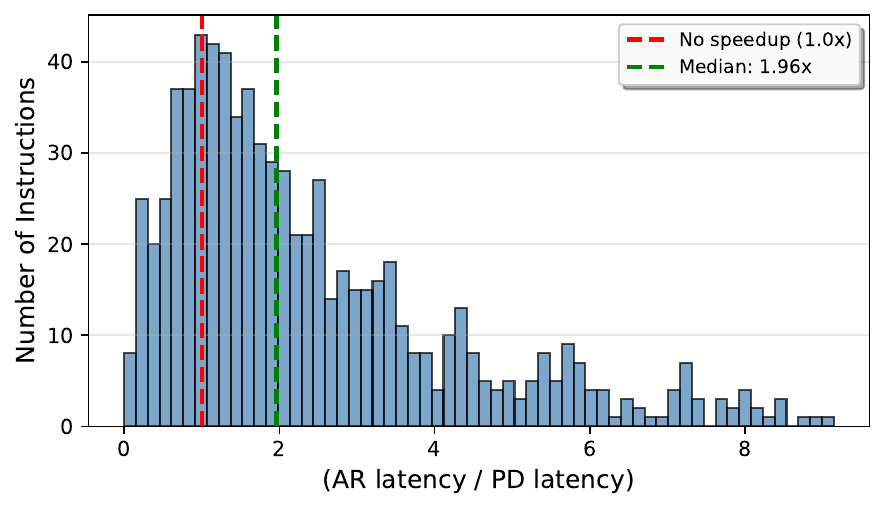}
    \caption{Distribution of speedup ratios.}
    \label{fig:pd:speedup-histogram}
\end{figure}

\paragraph{Full Speedup Distribution.}
\label{sec:pd:speedup}
This section details the performance variance observed across the evaluation dataset. \Cref{fig:pd:speedup-histogram} illustrates the distribution of speedup ratios obtained by comparing \pd{} against the autoregressive baseline. The dataset consists of $N = 803$ test samples. The analysis yields a mean speedup ratio of $3.08$ and the distribution exhibits a standard deviation of $4.10$. This significant variance relative to the mean suggests a right-skewed distribution, indicating that while the average performance gain is substantial, there exists a subset of cases where the proposed method achieves exceptionally high speedups.

\paragraph{KV Caching in Planned Diffusion.}
\label{sec:pd:kv-caching}
KV caching plays a substantial role in the efficiency of \pd{}. The model's hybrid attention mask determines the KV cache of \pd{}. The general principle is that the runtime can cache a token if future tokens do not affect its key and value embeddings: whether a token attends to future positions in the sequence determines its cacheability. The autoregressive planning stage uses causal attention and a conventional application of KV caching~\cite{pope2022efficiently}. In contrast, the diffusion stage employs a bidirectional mask, in which tokens inside an \pdasync{} chunk attend to each other. Because bidirectional attention does not support KV caching~\cite{israel2025enabling}, the runtime cannot cache tokens inside an \pdasync{} chunk until their respective denoising process completes. However, subsequent tokens, such as those in a new planning stage following a \pdsync{} tag, can efficiently attend to the KV cache of the preceding planning and diffusion stages. This caching mechanism is essential for combining the speed of diffusion with the computational savings of autoregressive KV caching.

\paragraph{Sparse Attention Mask.}
The attention mask for \pd{} combines causal and bidirectional attention. Causal attention is used for sequential planning stages. Bidirectional attention is used within \pdasync{} chunks for parallel denoising. In \pd{}, I enable concurrent chunks to cross-attend, unlike the sparse attention variant. After a \pdsync{} token, subsequent tokens can attend to all prior tokens. For illustrative purposes, this example shortens the diffusion chunks. The dense attention mask appears in \Cref{fig:pd:attention-mask}.

In \pdsa{}, the asynchronous blocks cannot attend to one another. Sparsity enables more compute-efficient training and inference.

\paragraph{Dataset Performance Breakdown.}
\label{sec:pd:dataset-breakdown}
The AlpacaEval dataset~\cite{alpaca_eval} comprises 5 datasets: self-instruct~\cite{wang2023self}, open-assistant~\cite{kopf2023openassistant}, vicuna~\cite{vicuna2023}, koala~\cite{koala_blogpost_2023}, and hh-rlhf~\cite{bai2022training}. For \pd{}, I report the length-controlled win rate for each and each standard error respectively. I observe that win rate is not especially sensitive to the dataset source of the instruction.

\begin{table}[h]
    \centering
    \caption{\pd{} LC Win Rates across different datasets.}
    \label{tab:pd:lc-win-rates}
    \begin{tabular}{lcc}
        \toprule
        Dataset & Number of Samples & LC Win Rate (\% $\pm$ SE) \\
        \midrule
        self-instruct & 249 & 45.7 $\pm$ 0.65 \\
        open-assistant & 187 & 44.2 $\pm$ 0.91 \\
        vicuna & 80 & 44.2 $\pm$ 1.66 \\
        koala & 156 & 48.0 $\pm$ 0.57 \\
        hh-rlhf & 129 & 46.3 $\pm$ 1.04 \\
        \bottomrule
    \end{tabular}
\end{table}

\paragraph{Qualitative Analysis.}
\label{sec:pd:qualitative}

To better understand the behavior of \pd{}, I analyze specific examples of generated outputs, focusing on the interaction between the autoregressive planning stage and the parallel diffusion stage. The examples below display the instruction and the raw output, which reveals the internal control tags generated by the planner.

In the raw output:
\begin{itemize}[leftmargin=*, topsep=0pt, itemsep=0pt]
    \item \texttt{<promise>-<topic>... length</topic>} indicates the autoregressive plan for a specific chunk.
    \item \pdasync{}\ldots\pdcasync{} indicates the text generated in parallel via discrete diffusion.
    \item \texttt{[PAD]} tokens indicate the unused capacity of the allocated chunk length.
\end{itemize}

The following example demonstrates a successful decomposition of a list-based query. The instruction asks for popular careers in development studies. The autoregressive planner successfully identifies distinct semantic entities before generation begins.

The planner generates specific, discriminatory topics for each chunk: \texttt{UN orgs}, \texttt{WB orgs}, \texttt{NGOs}, \texttt{media orgs}, etc. Because the planning stage explicitly disambiguates the content of each chunk, the subsequent diffusion execution stage, which generates these chunks in parallel with independent attention masks, does not suffer from mode collapse. Each diffusion process conditions on a unique semantic target, resulting in a comprehensive, non-repetitive list that correctly follows the instruction.

\begin{instructionbox}{Instruction}
Please answer the following question based on the information provided in the article. \\

Development studies is an interdisciplinary branch of social science. Development studies is offered as a specialized master's degree in a number of reputed universities around the world. It has grown in popularity as a subject of study since the early 1990s, and has been most widely taught and researched in developing countries and countries with a colonial history, such as the UK, where the discipline originated.  Students of development studies often choose careers in international organisations such as the United Nations, World Bank, non-governmental organisations (NGOs), media and journalism houses, private sector development consultancy firms, corporate social responsibility (CSR) bodies and research centers.\\\\
What are some popular careers for students of development studies
\end{instructionbox}

\begin{rawoutputbox}{Raw Output}
<promise>-<topic>introduction 2</topic>
<async>Some popular careers for students of development studies include:[PAD][PAD][PAD][PAD][PAD][PAD][PAD][PAD][PAD][PAD]</async><sync/>
<promise>-<topic>UN orgs 4</topic>
<promise>-<topic>WB orgs 4</topic>
<promise>-<topic>NGOs 4</topic>
<promise>-<topic>media orgs 4</topic>
<promise>-<topic>development consultancy 4</topic>
<promise>-<topic>CSR bodies 4</topic>
<promise>-<topic>research centers 4</topic>
<async>1. International Organizations: Students often work for organizations such as the United Nations (UN) to address global issues, including poverty, education, healthcare, and education.[PAD][PAD][PAD][PAD][PAD][PAD][PAD]</async>
<async>2. World Bank (WB): Similar to the UN, the World Bank offers opportunities to work on projects related to development, finance, and infrastructure in various countries.[PAD][PAD][PAD][PAD][PAD][PAD][PAD]</async>
<async>3. Non-governmental Organizations (NGOs): Many students of development studies pursue careers in NGOs, as these organizations are often involved in development projects and initiatives.[PAD][PAD][PAD][PAD][PAD][PAD][PAD][PAD]</async>
<async>4. Media and Journalism Houses: Students may become reporters and journalists, covering topics related to global issues, development, and the work of various organizations.[PAD][PAD][PAD][PAD][PAD][PAD][PAD][PAD][PAD][PAD]</async>
<async>5. Private sector development consultancy firms: Companies that provide advice and support to local businesses, helping them improve their operations, strategies, and adapt to the market.[PAD][PAD][PAD][PAD][PAD][PAD][PAD][PAD]</async>
<async>6. Corporate Social Responsibility (CSR) bodies: Students can work in these bodies to help businesses and organizations understand and meet their social responsibilities and promote sustainable practices.[PAD][PAD][PAD][PAD][PAD][PAD][PAD][PAD]</async>
<async>7. Research Centers: Students may also work in research centers, where they can contribute to the advancement of development studies research, analysis, and presentation.[PAD][PAD][PAD][PAD][PAD][PAD][PAD][PAD][PAD][PAD]</async><sync/>
<promise>-<topic>summary 4</topic>
<async>These are just a few examples of the many careers available to students of development studies. The field is diverse and offers opportunities to contribute to the global advancement of development and social change.[PAD][PAD][PAD][PAD]</async>
\end{rawoutputbox}

\begin{instructionbox}{Clean Output}
Some popular careers for students of development studies include: \\
1. International Organizations: Students often work for organizations such as the United Nations (UN) to address global issues, including poverty, education, healthcare, and education. \\
2. World Bank (WB): Similar to the UN, the World Bank offers opportunities to work on projects related to development, finance, and infrastructure in various countries. \\
3. Non-governmental Organizations (NGOs): Many students of development studies pursue careers in NGOs, as these organizations are often involved in development projects and initiatives. \\
4. Media and Journalism Houses: Students may become reporters and journalists, covering topics related to global issues, development, and the work of various organizations. \\
5. Private sector development consultancy firms: Companies that provide advice and support to local businesses, helping them improve their operations, strategies, and adapt to the market. \\
6. Corporate Social Responsibility (CSR) bodies: Students can work in these bodies to help businesses and organizations understand and meet their social responsibilities and promote sustainable practices. \\
7. Research Centers: Students may also work in research centers, where they can contribute to the advancement of development studies research, analysis, and presentation. These are just a few examples of the many careers available to students of development studies. The field is diverse and offers opportunities to contribute to the global advancement of development and social change.
\end{instructionbox}

The following example illustrates a failure mode that can result from enforcing independence between generated chunks. The instruction asks for a list of places to visit in November.

In this case, the autoregressive planner fails to decompose the semantic content effectively. Instead of generating specific locations (e.g., ``NYC'', ``Denver'', ``LA''), the planner generates generic structural tags: \texttt{list part 1}, \texttt{list part 2}, \texttt{list part 3}.

Because \pdsa{} uses block-sparse masking, where chunks cannot attend to each other during generation to maximize parallelism, the diffusion process for each chunk conditions only on the generic topic ``list part X'' and the instruction. Lacking specific constraints, the model defaults to the most probable completion for the prompt in every single chunk. This results in severe redundancy (New York City appears 4 times; Denver appears 4 times).

This failure mode highlights the importance of using dense attention (\pdsft{}), as discussed in \Cref{sec:pd:experiments}, or the necessity of high-quality synthetic training data that encourages the planner to be semantically explicit.

\begin{instructionbox}{Instruction}
Make a list of the top places in the U.S. to visit at the given time. \\ \\ November
\end{instructionbox}

\begin{rawoutputbox}{Raw Output}
<promise>-<topic>introduction 2</topic>
<promise>-<topic>list prompt 1</topic>
<async>Sure, here's a list of the top places in the U.S. to visit in November:</async>
<async>Here are the top places to visit:[PAD][PAD]</async><sync/>
<promise>-<topic>list part 1 4</topic>
<promise>-<topic>list part 2 3</topic>
<promise>-<topic>list part 3 3</topic>
<promise>-<topic>list part 4 3</topic>
<promise>-<topic>list part 5 3</topic>
<promise>-<topic>list part 6 3</topic>
<promise>-<topic>list part 7 3</topic>
<promise>-<topic>list part 8 3</topic>
<promise>-<topic>list part 9 3</topic>
<promise>-<topic>list part 10 3</topic>
<async>1. New York City[PAD][PAD][PAD][PAD][PAD]</async>
<async>2. New York City - Lots of shopping, museums, and entertainment.[PAD][PAD][PAD][PAD][PAD]</async>
<async>2. New York City - It's a big city with lots of fun things, shops, and places to eat.[PAD][PAD][PAD][PAD][PAD][PAD]</async><async>3. New York City - New York City is a big city with lots of things to see, like the Statue of Liberty, the park, and the buildings.[PAD][PAD][PAD][PAD][PAD][PAD][PAD]</async>
<async>4. Denver, Colorado - A great place for football, hiking, and skiing. It has the big city of Denver, the state of Colorado, and the beautiful mountains around it. It's also famous for the Denver Zoo.[PAD][PAD][PAD][PAD]</async>
<async>5. Los Angeles, California - Known for its beautiful coastline, Hollywood, and shopping, Los Angeles is a top place to visit in November. Visit places like Santa Monica, Santa Barbara, and Santa Cruz, and enjoy the local food, beaches, and culture.[PAD][PAD][PAD][PAD][PAD][PAD][PAD]</async>
<async>8. Sacramento, California - Known for its rich history and beautiful coastline, Sacramento is a major city in the state of California. It is the capital city of the United States and is home to many major events and festivals, such as the Sacramento Street Festival, the Sacramento Zoo, and the Sacramento Riverwalk.[PAD][PAD][PAD][PAD][PAD][PAD][PAD][PAD]</async>
<async>10. Denver, Colorado
- Denver is known for its beautiful landscapes, history, and art scene. It's one of the most popular cities in the U.S. and has a lot to do and see. Visit the U.S. Space Center, the Denver Zoo, and the University of Colorado Boulder to get a taste of the city.[PAD][PAD][PAD][PAD][PAD][PAD][PAD][PAD][PAD][PAD]</async>
<async>10. Denver, Colorado
- Denver is a major city in the U.S. with diverse history, culture, and landscapes. It is home to numerous attractions, such as the Denver Zoo, Science Center, andUSS, as well as the University of Colorado Boulder and the Denver Zoo. The city is also known for its mountains, forests, and parks, making it a great place for outdoor activities.[PAD][PAD][PAD][PAD][PAD][PAD][PAD]</async>
<async>10. Denver, Colorado
- Denver is a major city in the U.S. and offers a mix of history, culture, and beauty. Some of the top places to visit in Denver include the U.S. Open Center, Central Park, the Denver Zoo, and the Denver Nature Center. The University of Colorado, also located in Denver, is a great place to explore, and the Denver Art Museum offers a unique glimpse into the city.[PAD][PAD][PAD][PAD][PAD][PAD][PAD][PAD][PAD][PAD]</async>
\end{rawoutputbox}

\begin{instructionbox}{Clean Output}
Sure, here's a list of the top places in the U.S. to visit in November: Here are the top places to visit: \\1. New York City \\2. New York City - Lots of shopping, museums, and entertainment. \\2. New York City - It's a big city with lots of fun things, shops, and places to eat. \\3. New York City - New York City is a big city with lots of things to see, like the Statue of Liberty, the park, and the buildings. \\4. Denver, Colorado - A great place for football, hiking, and skiing. It has the big city of Denver, the state of Colorado, and the beautiful mountains around it. It's also famous for the Denver Zoo. \\5. Los Angeles, California - Known for its beautiful coastline, Hollywood, and shopping, Los Angeles is a top place to visit in November. Visit places like Santa Monica, Santa Barbara, and Santa Cruz, and enjoy the local food, beaches, and culture. \\8. Sacramento, California - Known for its rich history and beautiful coastline, Sacramento is a major city in the state of California. It is the capital city of the United States and is home to many major events and festivals, such as the Sacramento Street Festival, the Sacramento Zoo, and the Sacramento Riverwalk. \\10. Denver, Colorado\\- Denver is known for its beautiful landscapes, history, and art scene. It's one of the most popular cities in the U.S. and has a lot to do and see. Visit the U.S. Space Center, the Denver Zoo, and the University of Colorado Boulder to get a taste of the city. \\10. Denver, Colorado\\- Denver is a major city in the U.S. with diverse history, culture, and landscapes. It is home to numerous attractions, such as the Denver Zoo, Science Center, andUSS, as well as the University of Colorado Boulder and the Denver Zoo. The city is also known for its mountains, forests, and parks, making it a great place for outdoor activities. \\10. Denver, Colorado\\- Denver is a major city in the U.S. and offers a mix of history, culture, and beauty. Some of the top places to visit in Denver include the U.S. Open Center, Central Park, the Denver Zoo, and the Denver Nature Center. The University of Colorado, also located in Denver, is a great place to explore, and the Denver Art Museum offers a unique glimpse into the city.
\end{instructionbox}

\section{Related Work}
\label{sec:pd:related}

The work in this chapter builds upon recent developments in diffusion-based language models and parallel text generation. I position the contributions in three primary research areas: diffusion language models, semantic parallelism, and other parallel generation techniques.

\paragraph{Diffusion Language Models.}
Diffusion models have recently emerged as a new paradigm for generative language tasks~\cite{austin2021structured, sahoo2024simple, lou2024discretediffusionmodelingestimating, shi2025simplifiedgeneralizedmaskeddiffusion, nie2025large, ye2025dream, liu2025discretecopuladiffusion}. A significant body of research focuses on accelerating the inference process, which traditionally involves many iterative denoising steps. These acceleration techniques include KV caching for diffusion models~\cite{ma2025dkv, liu2025dllm}, the use of autoregressive verification~\cite{hu2025accelerating, israel2025accelerating}, and the development of fast sampling strategies that reduce the number of required steps~\cite{wu2025fastdllm, li2025diffusion, hong2025wide}. \pd{} complements these acceleration techniques and can integrate any diffusion sampling strategy, enabling further latency reductions. Other related work includes block diffusion~\cite{arriola2025block}, which enforces autoregressive structure over blocks, and planned denoising~\cite{liu2025thinkgeneratediscretediffusion}, which learns an adaptive denoising schedule; neither targets semantic parallelism.

\paragraph{Semantic Parallelism.}
Diffusion models parallelize token-level denoising but do not learn to exploit semantic independence across larger chunks of text. I define \emph{semantic parallelism} as a broad class of techniques that produce models capable of parallelizing over semantically independent chunks of tokens.
Many recent works explore semantic parallelism~\cite{ning2023skeleton,liu2024apar,jin2025pasta,rodionov2025hogwildinferenceparallelllm,pan2025learningadaptiveparallelreasoning,wen2025parathinkernativeparallelthinking,yang2025multiverselanguagemodelssecretly}.
These works all rely on autoregressive decoding within each chunk; \pd{} is, to the best of my knowledge, the first to instead denoise each chunk via diffusion.
When referring to hybrid models between autoregression and discrete diffusion, I define a hybrid model as a single architecture capable of pure left-to-right autoregression and pure discrete diffusion. This definition precludes existing works that interpolate between the autoregressive and discrete diffusion objective, such as Block Diffusion~\cite{arriola2025block}, Eso-LM~\cite{sahoo2025esoteric}, or ARDMs~\cite{hoogeboom2022autoregressivediffusionmodels}.
Another necessary clarification is that readers should not confuse discrete diffusion with vanilla continuous diffusion. For continuous diffusion, multimodal hybrids over autoregression and diffusion exist, such as HybridVLA~\cite{liu2025hybridvla} and Monoformer~\cite{zhao2024monoformer}.

\paragraph{Other Parallel Generation Techniques.}
Insertion-based models generate text by predicting where and what tokens to insert in a sequence. Multiple insertions can occur simultaneously, reducing the total number of decoding steps~\cite{stern2019insertion}. Insertion-based models parallelize at the token level; \pd{} instead learns to identify independent chunks and denoises them concurrently.
Speculative decoding accelerates autoregressive models by drafting multiple tokens and verifying them in parallel~\cite{leviathan2023fast,chen2023accelerating,Zhang_2024}.
\pd{} requires no separate draft model---a single model handles both autoregressive planning and parallel generation.

\section{Conclusion}
\label{sec:pd:conclusion}

This chapter introduces \pd{}, a novel hybrid architecture that combines sequential autoregressive planning with parallel diffusion denoising to improve the quality--latency trade-off in text generation.
The model exploits opportunities for parallelism within the semantic structure of text.
Experimental evaluation shows that \pd{} expands the quality--latency Pareto frontier, achieving significant speedup over pure autoregressive and diffusion baselines with a minimal drop in quality.
A shorter critical path and the approach's scalability with respect to training drive this efficiency gain.
I also show that \pd{} offers fine-grained control over the quality--latency trade-off at inference time.

The key insight that a single model can learn both autoregressive and diffusion objectives to annotate semantic dependence for parallel execution reinforces the self-orchestrating paradigm introduced in \Cref{ch:pasta} and extended to memory management in \Cref{ch:tip}.

\chapter{Unified Abstraction}
\label{ch:unified}

The three systems of this dissertation---PASTA (\Cref{ch:pasta}), TIP (\Cref{ch:tip}), and Planned Diffusion (\Cref{ch:pd})---reduce to a single abstraction: the \emph{task}---a named unit of generation that the model delimits during its output.
PASTA's tasks are response chunks that the runtime decodes in parallel; TIP's tasks are single deductive steps whose KV cache the runtime frees once redundant; Planned Diffusion's tasks are plan topics that the runtime denoises in parallel.
Each system trains the model to decompose generation into tasks and declare relationships between them, and a co-designed runtime interprets those annotations to orchestrate execution.
The key insight is that parallelism and context eviction are not separate capabilities; both follow from dependency relationships between tasks.
\looseness=-1
I introduce two flavors of unified self-orchestration building on tasks, each pairing an annotation language with a co-designed runtime: a declarative flavor in which the model annotates dependency structure and the runtime derives orchestration decisions, and an imperative flavor in which the model issues orchestration commands and the runtime executes them directly.

\section{Declarative Self-Orchestration}
\label{sec:unified:declarative}

The declarative flavor organizes tasks into the \emph{task DAG}: a directed acyclic graph where nodes are tasks and edges encode semantic dependence.
A task can be flat---a single chunk of text---or nested: a nested task contains an inner DAG of sub-steps and designates one sub-step as its \emph{output}; once the inner DAG produces its output, the runtime frees the remaining inner steps that derived it.
The model constructs the task DAG incrementally during generation by annotating two things: decomposition (where one task ends and another begins) and dependency (which prior tasks each new task depends on).
A minimal annotation language expresses both: the model emits a \emph{task header} at the start of each task that names it and lists its dependencies; Figure~\ref{fig:unified-annotation} illustrates on the geometry example from Figure~\ref{fig:semdep-dag}.

\begin{figure}[t]
\centering
\begin{tcolorbox}[colback=gray!5, colframe=gray!50, width=0.85\textwidth, boxrule=0.4pt, arc=2pt, fontupper=\small\ttfamily]
\textcolor{blue!70}{<task name=coordinates deps=\{\}>}\\
The two points are (2, -2) and (10, 4).\\
\textcolor{blue!70}{</task>}\\[4pt]
\textcolor{green!50!black}{<task name=formula deps=\{\}>}\\
The distance formula is $d = \sqrt{(x_2-x_1)^2 + (y_2-y_1)^2}$.\\
\textcolor{green!50!black}{</task>}\\[4pt]
\textcolor{orange!80!black}{<task name=compute deps=\{coordinates,formula\}>}\\
$d = \sqrt{(10-2)^2 + (4-(-2))^2} = \sqrt{64+36} = 10$.\\
\textcolor{orange!80!black}{</task>}
\end{tcolorbox}
\caption{The geometry example from Figure~\ref{fig:semdep-dag} in the declarative annotation language. Each task header declares a name and a dependency set. The \texttt{coordinates} and \texttt{formula} tasks declare no dependencies on each other; \texttt{compute} depends on both.}
\label{fig:unified-annotation}
\end{figure}

PASTA and Planned Diffusion each use a restricted form of this language.
Both express the same dependency pattern: a set of tasks with empty dependency sets (the independent chunks) followed by a task that depends on all of them (the synchronization point)---they differ only in execution mechanism (autoregressive threads versus diffusion denoising).
From these annotations, the runtime derives orchestration strategy: tasks with no dependency edge between them can execute concurrently.
In the geometry example of Figure~\ref{fig:unified-annotation}, the runtime generates \texttt{coordinates} and \texttt{formula} in parallel (no edge between them) and schedules \texttt{compute} after both complete.

TIP maps to the nested task structure because long chain-of-thought reasoning naturally decomposes into subgoal pursuits.
Once the model completes a subgoal, the thought process that derived it becomes irrelevant to future generation---as Figure~\ref{fig:tip:registers} illustrates.
Under the declarative abstraction, TIP models each subgoal pursuit as a nested task and frees the inner reasoning steps upon completion, retaining only the subgoal's output.

Table~\ref{tab:dag-view} summarizes how each system's annotations map to the task DAG.

\begin{table}[t]
\centering
\caption{Each system's annotations as a restricted view of the task DAG.}
\label{tab:dag-view}
\footnotesize
\setlength{\tabcolsep}{4pt}
\begin{tabular}{@{}llll@{}}
\toprule
System & Task structure & Dependency information & Runtime action \\
\midrule
PASTA & Flat: response chunks & Independence within a wave & Parallel decoding \\
Planned Diffusion & Flat: topics in a plan & Independence between topics & Parallel denoising \\
TIP & Nested: subgoals & Nesting with designated output & Free inner steps \\
\bottomrule
\end{tabular}
\end{table}

\noindent The flat and nested views yield the two orchestration strategies from a single abstraction: parallelism from the absence of edges between contemporaneous nodes, eviction from the completion of a nested subgoal whose output no future node depends on.

\section{Imperative Self-Orchestration}
\label{sec:unified:imperative}

An alternative to describing task-DAG structure within generation is issuing optimization commands directly: not declaring independence but forking threads; not defining subgoal boundaries but freeing inner steps.
I call this \emph{imperative self-orchestration}.
The model emits orchestration commands---\texttt{fork}, \texttt{wait}, \texttt{free}---and the runtime executes them directly rather than deriving a strategy from a dependency graph.
Figure~\ref{fig:imperative-annotation} illustrates this language on the same geometry example.

\begin{figure}[t]
\centering
\begin{tcolorbox}[colback=gray!5, colframe=gray!50, width=0.85\textwidth, boxrule=0.4pt, arc=2pt, fontupper=\small\ttfamily]
\textcolor{blue!70}{<fork name=coordinates>}\\
The two points are (2, -2) and (10, 4).\\
\textcolor{blue!70}{</fork>}\\[4pt]
\textcolor{green!50!black}{<fork name=formula>}\\
The distance formula is $d = \sqrt{(x_2-x_1)^2 + (y_2-y_1)^2}$.\\
\textcolor{green!50!black}{</fork>}\\[4pt]
\textcolor{orange!80!black}{<wait name=coordinates/> <wait name=formula/>}\\
$d = \sqrt{(10-2)^2 + (4-(-2))^2}$\\
\textcolor{orange!80!black}{<free name=coordinates/> <free name=formula/>}\\
$\phantom{d} = \sqrt{64+36} = 10$.
\end{tcolorbox}
\caption{The geometry example from Figure~\ref{fig:semdep-dag} expressed in an imperative annotation language. The model issues \texttt{fork} to spawn parallel threads, \texttt{wait} to synchronize, and \texttt{free} to release KV cache pages---commanding the runtime rather than declaring dependencies.}
\label{fig:imperative-annotation}
\end{figure}

\looseness=-1
Each system's annotation language is a restricted form of this command set.
PASTA and Planned Diffusion express the same orchestration pattern: a sequence of \texttt{fork} commands for each independent chunk followed by an implicit \texttt{wait} on all active threads at the synchronization point.
TIP's thought register reuse corresponds to a \texttt{free} command that releases the reasoning steps leading up to a completed subgoal, preserving only the subgoal's output in the KV cache.
Table~\ref{tab:imperative-view} summarizes how each system's annotations map to orchestration commands.

\begin{table}[t]
\centering
\caption{Each system's annotations as a restricted form of orchestration commands.}
\label{tab:imperative-view}
\footnotesize
\setlength{\tabcolsep}{4pt}
\begin{tabular}{@{}lll@{}}
\toprule
System & Commands used & Runtime action \\
\midrule
PASTA & \texttt{fork}, \texttt{wait} (implicit, all) & Spawn and join parallel threads \\
Planned Diffusion & \texttt{fork}, \texttt{wait} (implicit, all) & Spawn parallel denoising \\
TIP & \texttt{free} (implicit, by thought register) & Drop subgoal's inner KV cache pages \\
\bottomrule
\end{tabular}
\end{table}

The runtime executes each orchestration command as it arrives.
A \texttt{fork} spawns a parallel generation thread; a \texttt{wait} blocks until the named thread completes; a \texttt{free} drops a completed subgoal's inner KV cache pages while preserving its output.
In the geometry example of Figure~\ref{fig:imperative-annotation}, the runtime spawns \texttt{coordinates} and \texttt{formula} as parallel threads, blocks on both \texttt{wait} commands, frees their KV cache pages, and then generates the final computation.

The imperative approach also enables orchestration decisions that dependency structure cannot express.
The task DAG encodes which tasks depend on which---but some orchestration decisions are not about dependency:

\begin{itemize}[leftmargin=*, topsep=0pt, itemsep=2pt]
    \item \textit{Priority.} The model may know that one task lies on the critical path and should receive more compute or earlier scheduling. A \texttt{prioritize(name)} command communicates urgency, which dependency edges cannot encode---they constrain ordering but say nothing about how to allocate resources among independent tasks.
    \item \textit{Prefetching.} The model may predict that it will need certain context later and instruct the runtime to keep it warm in cache. A \texttt{pin(name)} command expresses a prediction about future access patterns, not a statement about current dependency.
\end{itemize}

\noindent These decisions require reasoning about execution strategy, not just semantic structure.

\looseness=-1
\section{Conclusion} The language model under a declarative abstraction identifies dependency structure and delegates optimization decisions to the runtime; the language model under an imperative abstraction must fulfill both responsibilities. Imperative self-orchestration therefore compounds two objectives: dependency analysis and execution strategy optimization. A declarative unified abstraction can thus be easier to implement than an imperative one.
This practical difference aside, the two flavors share what matters most: a task abstraction that turns a previously monolithic process of computation, LLM output generation, into a structured set of task execution and gives increasingly capable language models a mechanism to optimize their own inference.
This mirrors classical programming system design, importing a familiar vocabulary---tasks, dependencies, scheduling---and proven engineering principles that can guide future work.
Together with the empirical results of \Cref{ch:pasta,ch:tip,ch:pd}, this dissertation presents self-orchestrating language models as an answer to what the introduction identifies as one of the most pressing open problems in computing: improving LLM inference efficiency.

\chapter{Conclusion}
\label{ch:conclusion}

This dissertation demonstrates that the language model itself can annotate semantic dependence, and that a co-designed runtime can exploit these annotations to optimize inference. I show this across three systems---PASTA for parallel generation, TIP for context eviction, and Planned Diffusion for decomposed diffusion sampling. In this concluding chapter, I present the impact of this work, unify the three systems under a shared design methodology, trace its inspiration to data dependence analysis in the compiler literature, and close with an aspiration for self-orchestrating inference beyond this dissertation.

\section{Impact}
\label{sec:conclusion:impact}

Learning to Keep a Promise~\cite{jin2025learning} was the first work to demonstrate parallel thinking in language models, leading the way for a substantial line of research on this subject---including Hogwild!\ Inference~\cite{rodionov2025hogwildinferenceparallelllm}, Adaptive Parallel Reasoning~\cite{pan2025learningadaptiveparallelreasoning}, Group Think~\cite{hsu2025groupthink}, SPRINT~\cite{biju2025sprint}, Multiverse~\cite{yang2025multiverselanguagemodelssecretly}, ASPD~\cite{chen2025aspd}, ParaThinker~\cite{wen2025parathinkernativeparallelthinking}, Parallel-R1~\cite{zheng2025parallel}, ThreadWeaver~\cite{lian2025threadweaver}, and HybridFlow~\cite{dong2025hybridflow}, with A Survey on Parallel Reasoning~\cite{wang2025parallelreasoning} formally organizing the subfield.
Across this wave, the shared premise is the one PASTA made concrete: that the model itself should annotate which parts of its output can run in parallel, and that an execution engine should act on those annotations.

Several systems directly build on PASTA's design.
ASPD~\cite{chen2025aspd} inherits PASTA's three-layer framing---a language for expressing parallelism, an execution engine that acts on it, and KV-cache-aware decoding---and replaces PASTA's ``pre-allocated position ranges''~with branch-invisible attention masks and shared position IDs, resolving position-encoding mismatches when actual generation lengths deviate from predictions.
SPRINT~\cite{biju2025sprint} extends PASTA's ``decompose a task into parallel subtasks and subsequently merge their full context back into a single main thread'' pattern with planner/executor rounds for long-horizon reasoning.
FastDriveCoT~\cite{gu2026fastdrivecot} carries the same decompose-and-dispatch pattern into autonomous driving, decomposing a structured chain-of-thought into a dependency graph and generating independent sub-tasks concurrently.

\section{Unifying Design Methodology}
\label{sec:conclusion:methodology}

The three systems of this thesis---PASTA, TIP, and Planned Diffusion---share a uniform design methodology. Each project defines an annotation language that the model emits alongside its response, a co-designed runtime that interprets those annotations to orchestrate execution, and a training procedure that teaches the model to strategically use the annotation language for efficiency gain. Each component closes a distinct gap: the annotation language enables the model to surface semantic dependence, the runtime enables the system to act on that dependence during execution, and training enables the model to produce correct annotations. Table~\ref{tab:unifying-methodology} lays this parallel out across the three systems: PASTA annotates parallel response chunks and runs them as concurrent decoding threads, TIP annotates reasoning steps with registers and evicts their KV-cache entries when the model overwrites a register, and Planned Diffusion annotates parallel topics and schedules their denoising.
\Cref{ch:unified} takes this further, showing that all three annotation languages reduce to a single primitive, the task, and that both parallelism and context eviction follow from dependency relationships between tasks.

\begin{table}[t]
\centering
\caption[Unifying design methodology across the three systems of this thesis]{Unifying design methodology across the three systems of this thesis. Each project instantiates the same three components---annotation language, runtime, and training procedure---to address a different inference bottleneck.}
\label{tab:unifying-methodology}
\footnotesize
\setlength{\tabcolsep}{4pt}
\begin{tabular}{@{}llll@{}}
\toprule
Project & Annotation Language & Runtime & Training \\
\midrule
PASTA & \texttt{\scriptsize <promise/>, <async>, <sync/>} & Parallel decoding threads & SFT + preference opt.\ \\
TIP & \texttt{\scriptsize <reg\_0>}\ldots\texttt{\scriptsize <reg\_\{N-1\}>} & KV cache eviction & SFT + RL with GRPO \\
Planned Diffusion & \texttt{\scriptsize <topic>, <async>, <sync/>} & Diffusion denoising order & SFT \\
\bottomrule
\end{tabular}
\end{table}

\paragraph{Expanding the quality-speedup frontier.}
Each system must balance three factors: the annotation overhead the model incurs (\promise{} and \sync{} tokens in PASTA, register tokens in TIP, plan tokens in Planned Diffusion), the speedup the runtime delivers, and the quality delta from altered execution. A Pareto improvement requires that speedup outweighs annotation overhead at equivalent or better quality. This imposes a minimum problem size: TIP needs reasoning traces long enough that intermediate steps stop contributing to later ones; PASTA and Planned Diffusion need responses with multiple parallelizable chunks. Below that size, annotations cost more than they buy. The horizon of model workloads keeps growing, though: long-reasoning chains, multi-step coding problems, and agent trajectories now routinely span minutes of token generation---precisely the regime where annotation cost becomes negligible relative to runtime savings.

\paragraph{Where the methodology applies.}
The recipe generalizes to inference bottlenecks whose optimization requires semantic dependence: dependence that follows from the meaning of the generated content rather than from the structure of the decoding algorithm. No static analysis on the prompt recovers it; no external scheduler observes it; the model alone can supply it. Parallel decoding, denoising scheduling, and KV-cache eviction all fit this shape.

\section{Vision}
\label{sec:conclusion:vision}

Language models have arguably made the largest jump in AI capabilities in recent years, from mathematical reasoning to competitive programming. Yet serving infrastructure has not made commensurate progress: it still operates under the assumption that models cannot reason about their own execution. Concretely, state-of-the-art serving systems~\cite{kwon2023efficient, zheng2023efficiently} manage model execution---paged attention, speculative decoding---but never let the model manage the serving system.

But the model can manage the serving system. As this dissertation has shown, these models can produce semantic dependence annotations over their own future generation---information no external scheduler can access. When I surface that information through semantic dependence, I obtain Pareto improvements in parallel decoding, in denoising scheduling, and in memory management---three very different systems problems, one methodology. This dissertation points at a future where the boundary between ``the model'' and ``the infrastructure'' dissolves, because the model is uniquely positioned to manage its own execution in the most resource-efficient way.

I do not pretend this is easy. When designing self-orchestrating systems, the model must decide, in a matter of split seconds, which tokens to generate in parallel, how to schedule denoising, or which context to evict. The co-design between model training and the interfaces for self-orchestration must be exactly right to avoid creating new efficiency bottlenecks. But this is precisely what makes self-orchestrating language model systems the most exciting direction I can imagine.

\appendix
\chapter{PASTA Supplementary Material}
\label{app:pasta}

\section{Dataset Annotation Prompt}
\label{app:pasta:dataset-prompts}
Below is the prompt I use to annotate the \lang{} seed dataset. I omit additional in-context examples for presentation.

\begin{lstlisting}
As a highly-paid expert annotator, you will be given a chatbot response, your job is to decide whether and how this response may be generated asynchronously in parallel by a large language model. Parallel text generation allows the model to generate segments of text simultaneously rather than sequentially, significantly improving efficiency and response times.

You will first identify whether the given chatbot response may be generated in parallel. You are to then annotate the chatbot response using specific tags that highlight segments suitable for parallel generation.

Use <async> tags to denote segments of text that may be generated asynchronously in parallel with respect to the text that follows. Thus apply <async> tags only to sentences that do not serve as necessary context for subsequent sentences. Sentences that are crucial for understanding or generating following text are not suitable for parallel asynchronous generation. For each <async> tag, include a very concise topic description of the text surrounded within the <async> tags. The topic description will be accessible to text generation after the closing async tag to ensure continuity and coherence.

Use the singleton <sync/> tag for synchronization. All content generated before <sync/>, including text marked by <async> is accessible to subsequent text generation after the <sync/> tag, ensuring continuity and coherence.

Detailed Instructions:
- Tagging Rules:
 - Use <async> tag in pairs.
 - Ensure that each <async> tag encompasses at least five words.
 - Refrain from altering the content of the response during annotation.
 - Use a maximum of 3 words in the topic description.
 - Use <sync/> sparingly as it introduces significant slowdown.

Input conversation:
```json
[
  {
    "from": "human",
    "value": "Article: In March 1968, CBS and Sony formed CBS/Sony Records, a Japanese business joint venture. With Sony being one of the developers behind the compact disc digital music media, a compact disc production plant was constructed in Japan under the joint venture, allowing CBS to begin supplying some of the first compact disc releases for the American market in 1983.\n\nNow answer this question: In what year did CDs come to the American market?"
  },
  {
    "from": "chatbot",
    "value": "CDs (Compact Discs) came to the American market in 1983. This was made possible through a joint venture between CBS and Sony, named CBS/Sony Records, which was formed in March 1968. As Sony was one of the developers behind the compact disc digital music media, they built a compact disc production plant in Japan under this joint collaboration. This plant enabled CBS to supply some of the first compact disc releases for the American market in 1983. The introduction of CDs revolutionized the music industry as they provided a new and improved way to store and listen to music, providing higher sound quality and durability compared to the earlier vinyl records and cassette tapes."
  }
]
```
Annotated chatbot response:
```json
{
  "from": "chatbot",
  "value": "CDs (Compact Discs) came to the American market in 1983. <async topic='joint venture'>This was made possible through a joint venture between CBS and Sony, named CBS/Sony Records, which was formed in March 1968.</async> <async topic='japan production plant'>As Sony was one of the developers behind the compact disc digital music media, they built a compact disc production plant in Japan under this joint collaboration. This plant enabled CBS to supply some of the first compact disc releases for the American market in 1983.</async> <async topic='impact'> The introduction of CDs revolutionized the music industry as they provided a new and improved way to store and listen to music, providing higher sound quality and durability compared to the earlier vinyl records and cassette tapes.</async>"
}
```

...
\end{lstlisting}

\chapter{TIP Supplementary Material}
\label{app:tip}

\section{Step Dependence Annotation Prompts}
\label{app:tip:annotation-prompts}

I annotate step dependence in two passes. The first pass segments a reasoning trace into discrete steps; the second pass annotates the dependence structure among those steps.

\subsection{Step Segmentation Prompt}
\label{app:tip:prompt-steps}

\begin{lstlisting}
Given this mathematical problem and solution:

Problem: {question}
Solution part to annotate:
{thought}

Your task is to:

1. Add XML step tags around the EXISTING text in the solution. DO NOT modify or rephrase the original text.
Each step should be wrapped like this:
<step id="X" task="Y">[EXACT ORIGINAL TEXT]</step>

When identifying steps, consider:
- A step must be self-contained: it states one complete idea, calculation, or deduction.
- Each answer/substep verification is a separate logical step
- Each exploration of a different solution strategy is a separate step
- Each self-contained sequential series of mathematical operations or transformations is a step
- Each step has a short, unique and distinct task name describing what the step is doing.

**CRITICAL UNIQUENESS REQUIREMENT:**
- EVERY task name MUST be completely unique within the entire solution
- NEVER reuse the same task name, even for similar operations
- If you perform similar operations multiple times, number them or add distinguishing details
- If you revisit or reconsider something, use specific descriptors like "second attempt", "alternative approach", "verify using different method"

**Examples of how to handle repeated operations:**
- Instead of: "Apply Stirling" twice, use: "Apply Stirling approximation" and "Apply Stirling formula to verify"
- Instead of: "Consider cos(36)" multiple times, use: "Recall cos(36) value", "Verify cos(36) using geometry", "Double-check cos(36) calculation"
- Instead of: "Apply Vieta" repeatedly, use: "Apply Vieta's formulas", "Use Vieta for second polynomial", "Verify using Vieta's relations"

Examples of logical steps:
<step id="1" task="Isolate y">Rearrange 3x + 2y = 12 to obtain y = -1.5x + 6.</step>
<step id="2" task="Compute derivative">Differentiate f(x) = x^3 - 3x to get f'(x) = 3x^2 - 3.</step>
<step id="3" task="Check discriminant">Calculate D = b^2 - 4ac = 25 - 24 = 1 to confirm two real roots.</step>
<step id="4" task="Verify root">Substitute x = 2 into f(x) and note f(2) = 0.</step>
<step id="5" task="Apply Pythagoras">Find the hypotenuse: c = sqrt(a^2 + b^2) = sqrt(9 + 16) = 5.</step>

Now write the annotated solution:
\end{lstlisting}

\subsection{Dependence Annotation Prompt}
\label{app:tip:prompt-deps}

\begin{lstlisting}
Given this mathematical problem and solution:

Problem: {question}
Annotated solution part to update:
{thought_with_steps}

Your task is to:

1. Add the "depends_on" XML attribute for the EXISTING XML step tags. DO NOT modify the existing tags. DO NOT rephrase the text. DO NOT modify "id". DO NOT modify "task". Only add the "depends_on" attribute. The "depends_on" attribute specifies which step tags the given tag logically depends on.
Given an existing step tags:
<step id="1" task="X">some content</step>
<step id="2" task="Y">some other text that depends on step 1</step>
The result would be:
<step id="1" task="X">some content</step>
<step id="2" task="Y" depends_on="1">some other text that depends on step 1</step>

Specify multiple dependent tasks delimited by semicolon:
<step id="1" task="X">some content</step>
<step id="2" task="Y">some content</step>
<step id="3" task="Y" depends_on="1;2">some other text that depends on step 1 and step 2</step>

When identifying dependencies, consider:
- A step must be self-contained: it states one complete idea, calculation, or deduction.
- Each answer/substep verification is a separate logical step
- Each exploration of a different solution strategy is a separate step
- Each self-contained sequential series of mathematical operations or transformations is a step
- Each step has a short, unique and distinct task name describing what the step is doing.
- A step depends on another step if it utilizes the information from the step either positively or negatively.
- Dependencies are transitive

General rules of thumbs:
- Making on progress on a task/subtask generally depends on the step that identifies the task/subtask
- Steps in parallel solutions paths do not depend on each other, until the merge point that considers the results from the paths
- The final answer should depend on all prior steps (since dependencies are transitive, just specify the final steps of the parallel threads that are not joined yet), including answer checking steps

Examples of logical steps:
<step id="1" task="Isolate y">Rearrange 3x + 2y = 12 to obtain y = -1.5x + 6.</step>
<step id="2" task="Compute derivative">Differentiate f(x) = x^3 - 3x to get f'(x) = 3x^2 - 3.</step>
<step id="3" task="Check discriminant">Calculate D = b^2 - 4ac = 25 - 24 = 1 to confirm two real roots.</step>
<step id="4" task="Verify root">Substitute x = 2 into f(x) and note f(2) = 0.</step>
<step id="5" task="Apply Pythagoras">Find the hypotenuse: c = sqrt(a^2 + b^2) = sqrt(9 + 16) = 5.</step>

Do not write ``` or ```xml.
DO NOT modify how the step tags annotation the text. DO NOT rephrase the text. DO NOT modify "id". DO NOT modify "task".

Now write the updated annotations:
\end{lstlisting}

\chapter{Planned Diffusion Supplementary Material}
\label{app:pd}

\section{Data Annotation Prompt}
\label{app:pd:annotation-prompt}
I present a shortened version of the data-annotation prompt below. I use it to instruct Gemini Flash 2.0 (temperature = 1.0, top-p = 0.95) to annotate the training data.

\begin{verbatim}
You will first identify whether the given chatbot response may be generated in
parallel. You are to then annotate the chatbot response using specific tags
that highlight segments suitable for parallel generation.

Use <async> tags to denote segments of text that may be generated asynchronously
in parallel with respect to the text that follows. Thus apply <async> tags only
to sentences that do not serve as necessary context for subsequent sentences.
Sentences that are crucial for understanding or generating following text are
not suitable for parallel asynchronous generation. For each <async> tag, include
a very concise topic description of the text surrounded within the <async> tags.
The topic description will be accessible to text generation after the closing
async tag to ensure continuity and coherence.

Use the singleton <sync/> tag for synchronization. All content generated before
<sync/>, including text marked by <async> is accessible to subsequent text
generation after the <sync/> tag, ensuring continuity and coherence.

Detailed Instructions:
- Tagging Rules:
 - Use <async> tag in pairs.
 - Ensure all text content is encapsulated within <async> tags.
 - Ensure that each <async> tag encompasses at least five words.
 - Refrain from altering the content of the response during annotation.
 - Use a maximum of 3 words in the topic description.
 - Use <sync/> sparingly as it introduces significant slowdown.
\end{verbatim}

\defbibheading{bibintoc}{\chapter*{#1}\addcontentsline{toc}{backmatter}{\refname}} 

\printbibliography[title={\refname},heading=bibintoc]

\end{document}